%% file: neurips_2025.tex
\documentclass{article}

\PassOptionsToPackage{numbers, compress, sort}{natbib}

\usepackage[preprint]{neurips_2025}

\usepackage[utf8]{inputenc} 
\usepackage[T1]{fontenc}    
\usepackage{hyperref}       
\usepackage{url}            
\usepackage{booktabs}       
\usepackage{amsfonts}       
\usepackage{nicefrac}       
\usepackage{microtype}      
\usepackage{xcolor}         

\usepackage{graphicx}
\usepackage[font=footnotesize]{caption} 

\usepackage{multirow} 

\newcommand\blfootnote[1]{%
  \begingroup
  \renewcommand\thefootnote{}\footnotetext{#1}%
  \addtocounter{footnote}{-1}%
  \endgroup
}

\usepackage{xspace}

\input{preamble}
\newcommand{\paragrapht}[1]{\vspace{-10pt}\paragraph{#1}}

\title{Emergent Temporal Correspondences\\ from Video Diffusion Transformers}

\author{
  Jisu Nam$^{*1}$ \qquad Soowon Son$^{*1}$ \qquad Dahyun Chung$^{2}$ \qquad Jiyoung Kim$^{1}$ \vspace{0.3em}\\
  \textbf{Siyoon Jin}$^{1}$ \qquad \textbf{Junhwa Hur}$^{\dagger3}$ \qquad \textbf{Seungryong Kim}$^{\dagger1}$  \vspace{0.3em}\\
  $^{1}$KAIST \qquad $^{2}$Korea University \qquad $^{3}$Google DeepMind  \vspace{0.3em}\\
   {\tt\small\ \href{https://cvlab-kaist.github.io/DiffTrack}{https://cvlab-kaist.github.io/DiffTrack}} \\
}

\begin{document}

\maketitle

\blfootnote{$^*$Equal contribution.}
\blfootnote{$^\dag$Co-corresponding author.}

\begin{figure}[htbp]
    \centering
   \vspace{-20pt}
    \includegraphics[width=\linewidth]{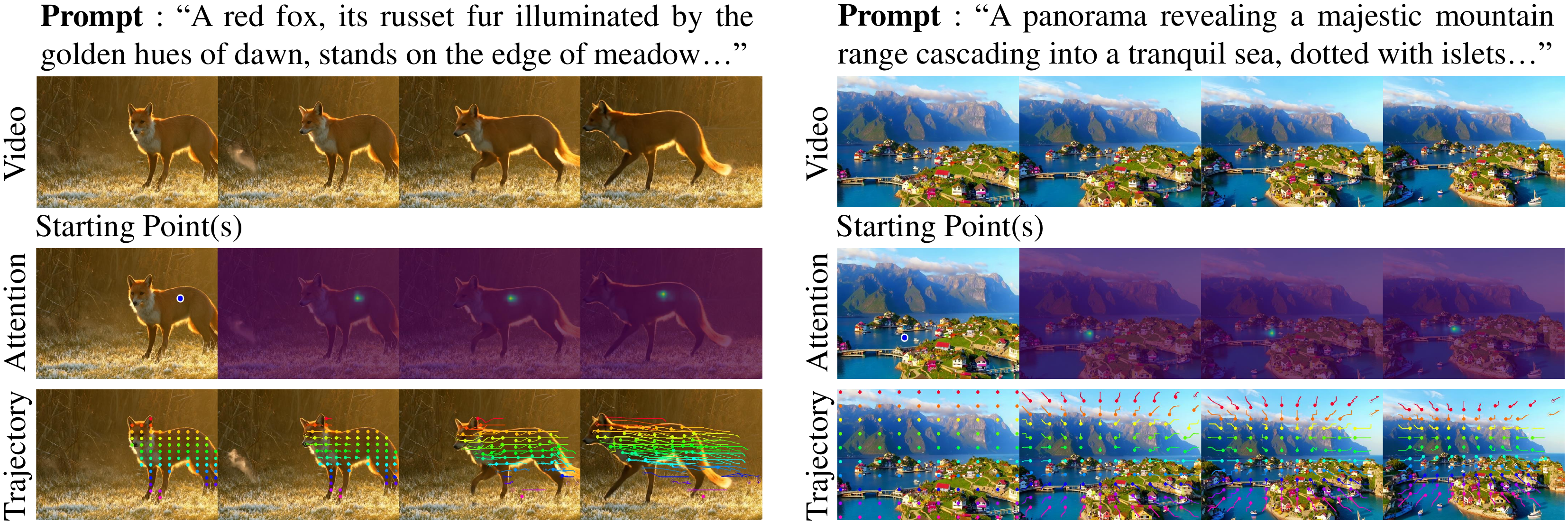}
    \caption{\textbf{Teaser:} DiffTrack reveals how video Diffusion Transformers (DiTs) establish temporal correspondences during video generation. Given a prompt and starting points, DiffTrack tracks how individual points align across subsequent frames via cross-frame attention in video DiTs (second row). This enables the extraction of coherent motion trajectories (third row) from both generated and real-world videos in a zero-shot manner.}
\label{qual:teaser}
\end{figure}

\input{sec/0_abstract}
\input{sec/1_intro}
\input{sec/3_pre}
\input{sec/4_method}

\input{sec/5_con}

\clearpage
\onecolumn

\input{sec/suppl}

{
    \small
    \bibliographystyle{plain} 
    \bibliography{egbib}
}

\end{document}

%% file: sec/0_abstract.tex
\begin{abstract}
\label{sec:abstract}
Recent advancements in video diffusion models based on Diffusion Transformers (DiTs) have achieved remarkable success in generating temporally coherent videos. 
Yet, a fundamental question persists: how do these models internally establish and represent temporal correspondences across frames?  We introduce \textit{DiffTrack}, the first quantitative analysis framework designed to answer this question. DiffTrack constructs a dataset of prompt-generated video with pseudo ground-truth tracking annotations and proposes novel evaluation metrics to systematically analyze how each component within the full 3D attention mechanism of DiTs (\eg,~representations, layers, and timesteps) contributes to establishing temporal correspondences. Our analysis reveals that query-key similarities in specific, but not all, layers play a critical role in temporal matching, and that this matching becomes increasingly prominent during the denoising process. We demonstrate practical applications of DiffTrack in zero-shot point tracking, where it achieves state-of-the-art performance compared to existing vision foundation and self-supervised video models. Further, we extend our findings to motion-enhanced video generation with a novel guidance method that improves temporal consistency of generated videos without additional training. We believe our work offers crucial insights into the inner workings of video DiTs and establishes a foundation for further research and applications leveraging their temporal understanding. 
\end{abstract}

%% file: sec/1_intro.tex
\section{Introduction}
\label{sec:intro}

Recent video diffusion models~\cite{zheng2024open, yang2024cogvideox, kong2024hunyuanvideo, polyak2025moviegencastmedia, liu2024sora, hacohen2024ltx, genmo2023mochi, kling2024, runway2024gen3}, powered by Diffusion Transformers (DiTs)~\cite{esser2024scaling, peebles2023scalable}, have achieved remarkable progress in generating realistic videos with high photometric fidelity, temporal coherence, and complex motion.
DiTs utilize full 3D attention to process all frames and text in a single sequence, enabling the effective propagation of spatial and temporal information, thus improving temporal coherence in generated videos.
Their strong temporal priors have also led to various downstream tasks, such as video depth estimation~\cite{zhang2024world}, pose estimation~\cite{cai2024can}, and 3D/4D reconstruction~\cite{sun2024dimensionx, liu2024reconx, wu2024cat4d, gao2024cat3d}.

Despite the success of these models, a fundamental question remains unanswered: \textit{How do video DiTs establish temporal correspondences between frames during generation, and how can we explicitly extract them?}
While existing works~\cite{hedlin2023unsupervised, tang2023emergent,zhang2023tale, meng2025not,nam2024dreammatcher,jin2025appearance} explore internal representations in U-Net-based image diffusion models, they mainly focus on two-frame correspondences.

In this paper, we introduce \textit{DiffTrack}, the first in-depth quantitative analysis framework designed to pinpoint temporal matching within the video DiT architecture.
DiffTrack provides insights into how and where video DiTs establish temporal correspondences during video generation, enabling the direct extraction of motion information from generated videos (\cf \cref{qual:teaser}).
Notably, our framework is adaptable to any DiT-based video generative models.

Our analysis investigates 3D attention, a core component of video DiTs, to determine which representations (\eg query-key similarities \vs intermediate features), layers, and denoising timesteps are most critical for establishing temporal correspondence among frames.

To systematically analyze their intricate interplays during video generation,
we construct a dataset of prompt-generated video using a video backbone under analysis and obtain pseudo ground-truth motion tracks from an off-the-shelf tracking method~\cite{karaev2024cotracker}. We then propose novel evaluation metrics that jointly assess temporal matching accuracy and confidence.
Given the dataset, we extract descriptors at desired layers and timesteps, estimate temporal correspondence, and evaluate these correspondences using our proposed metrics.

Through the analysis, we uncover several key findings.
1) Query-key matching in 3D attention blocks provides clear temporal matching information. 2) A few specific layers play a dominant role in establishing temporal matching.
3) Temporal matching strengthens throughout the denoising process.

We further demonstrate the practical value of our analysis through experiments in zero-shot point tracking.
By using the most significant feature descriptors for temporal matching, extracted at the optimal layer and timestep identified by DiffTrack, our approach achieves state-of-the-art results compared to existing vision foundation models~\cite{caron2021emerging, darcet2023vision, oquab2023dinov2, tang2023emergent, rombach2022high} and self-supervised video models~\cite{qian2023semantics, jabri2020space, xu2021rethinking, li2023spatial, blattmann2023stable, an2024cross}, demonstrating the effectiveness of our framework.

Additionally, we extend our analysis to motion-enhanced video generation with a novel guidance technique, Cross-Attention Guidance (CAG).
CAG works by perturbing cross-frame attention maps in the most dominant layers identified by DiffTrack, and guides the model away from degraded samples during sampling. Notably, CAG achieves significant gains in both human evaluations and automatic metrics, without requiring additional training, auxiliary networks, or supervision.

In summary, our contributions are:
\begin{itemize}
\item We identify the importance of understanding temporal correspondence in video DiTs and introduce {DiffTrack}, a novel framework that quantitatively analyzes and identifies temporal matching information within DiTs during video generation.
\item We provide a detailed analysis of the open-source video DiT models including CogVideoX~\cite{yang2024cogvideox}, HunyuanVideo~\cite{kong2024hunyuanvideo}, and CogVideoX-I2V~\cite{yang2024cogvideox}, revealing key insights into their internal mechanisms.
\item We demonstrate the effectiveness of {DiffTrack} in zero-shot point tracking, achieving state-of-the-art performance among existing vision foundation and self-supervised video models.
\item We present motion-enhanced video generation with CAG, a novel guidance method that improves the motion consistency of generated videos without auxiliary models or supervision.

\end{itemize}

%% file: sec/3_pre.tex
\section{Preliminaries}
\label{sec:pre}

\subsection{Video Diffusion Models}
\label{pre:VDM}
Video diffusion models~\cite{liu2024sora, zheng2024open, kong2024hunyuanvideo, yang2024cogvideox, polyak2025moviegencastmedia, hacohen2024ltx, genmo2023mochi, kling2024, runway2024gen3} generate videos from input text prompts through iterative denoising.
To enhance efficiency, latent video diffusion models \cite{he2022latent} perform this denoising in a latent space, typically using 3D Variational Autoencoders (VAE)~\cite{yang2024cogvideox} for spatial and temporal compression.
A 3D VAE encodes a video $\mathbf{X} \in \mathbb{R}^{(1+F)\times H \times W \times 3}$, with its frame count ($1{+}F$), height ($H$), width ($W$), into a compressed latent representation.
This latent is then patchified and unfolded into a sequence $\mathbf{z}_\mathrm{video}$ of length $(1{+}f)hw$, where $f$, $h$, and $w$ are derived from spatial compression ratio $p$ (\ie, $h{=}{H}/{p}$ and $w{=}{W}/{p}$) and a temporal compression ratio $q$ (\ie, $f{=}{F}/{q}$) with often skipping the first frame~\cite{yang2024cogvideox}.
To incorporate text input, a text encoder~\cite{raffel2020exploring} embeds the prompt into the text embedding $\mathbf{z}_\mathrm{text}$ with sequence length $S$.
Then the concatenated embeddings ($\mathbf{z}_\mathrm{video}$ and $\mathbf{z}_\mathrm{text}$), of length $(1+f)hw + S$, guide the denoising process, enabling text-aligned video generation. Given a predetermined noise schedule, forward process progressively adds Gaussian noise to $\mathbf{z}_{\mathrm{video}, t}$ for timestep $t$, producing noisier latents $\mathbf{z}_{\mathrm{video}, t+1}$.
In the denoising process, a neural network $\mathbf{\epsilon}_\theta(\mathbf{z}_{\mathrm{video}, t}, \mathbf{z}_{\mathrm{text}}, t)$ predicts and removes the noise to obtain $\mathbf{z}_{\mathrm{video}, t-1}$, aligning with the text prompt.
After $T$ denoising steps, the fully denoised latent $\mathbf{z}_{\mathrm{video}, 0}$ is decoded by the 3D VAE to generate the final video $\mathbf{X}'$.

\subsection{Video Diffusion Transformers}
\label{sec:dit}

Following the success of Sora~\cite{liu2024sora},
DiT~\cite{esser2024scaling, peebles2023scalable} has become a standard architecture for video generative models~\cite{yang2024cogvideox, hacohen2024ltx, genmo2023mochi, kong2024hunyuanvideo, kling2024}. DiT employs multiple layers of full 3D attention, enabling direct interaction between visual and textual information through attention and feed-forward processing.

At each timestep $t$, the concatenated sequences of $\mathbf{z}_{\mathrm{video}, t}$ and $\mathbf{z}_\mathrm{text}$ are augmented with 3D positional embeddings, \eg, RoPE~\cite{su2024roformer} or sinusoidal embeddings~\cite{vaswani2017attention}.
Then, at each layer $l$, full 3D attention transforms the sequences into queries $\mathbf{Q}_{t,l}$, keys $\mathbf{K}_{t,l}$, and values $\mathbf{V}_{t,l}$, all lying in $\mathbb{R}^{((1+f)hw + S) \times d}$ with the channel dimension $d$. 
The resulting output of full 3D attention is computed as:
\begin{align}
    \mathtt{Attn}(\mathbf{Q}_{t,l}, \mathbf{K}_{t,l}, \mathbf{V}_{t,l}) &= \mathbf{A}_{t,l}\mathbf{V}_{t,l} \\
   \text{with} \quad \mathbf{A}_{t,l} = \mathtt{Softmax}&({\mathbf{Q}_{t,l} \mathbf{K}_{t,l}^T}/{\sqrt{d}}).
\end{align}
To understand how the video latent $\textbf{z}_{\mathrm{video}, t}$ and text embedding $\textbf{z}_\mathrm{text}$ interact in full 3D attention, the notations of query $\mathbf{Q}_{t,l}$ and key $\mathbf{K}_{t,l}$ can be rewritten at the token sequence level:
\begin{align}
\mathbf{Q}_{t,l} &= \mathtt{Concat}({\mathbf{Q}_{ t, l}^1}, \dots, {\mathbf{Q}^{1+f}_{t, l}}, {\mathbf{Q}^\mathrm{text}_{t, l}}) \\
\mathbf{K}_{t,l} &= \mathtt{Concat}({\mathbf{K}_{ t, l}^1}, \dots, {\mathbf{K}^{1+f}_{t, l}}, {\mathbf{K}^\mathrm{text}_{t, l}}).
\end{align}

\input{figure/ill_attention}

Here, $\mathbf{Q}^i_{t, l}$ and $\mathbf{K}^i_{t, l}$ are projections from the $i$-th frame latent, with $i \in \{1, \dots, 1{+}f\}$, and $\mathbf{Q}^\mathrm{text}_{t, l}$ and $\mathbf{K}^\mathrm{text}_{t, l}$ are from the text embeddings. $\mathtt{Concat}(\cdot)$ concatenates these across the token sequences.

\cref{fig:ill_attention} illustrates full 3D attention in video DiTs.
It can be categorized into four key interactions: 1) self-frame attention $\mathbf{A}_{t,l}^{i}$, 2) cross-frame attention $\mathbf{A}_{t,l}^{i,j}$, 3) text-frame attention $\mathbf{A}_{t,l}^{\mathrm{text},j}$ or $\mathbf{A}_{t,l}^{i, \mathrm{text}}$, and 4) self-text attention $\mathbf{A}_{t,l}^{\mathrm{text}}$, where $i,j \in [1, 1{+}f]$ with $i \neq j$.
Self-frame attention attends to its own frame, focusing on nearby spatial patches at the pixel location.
Text-frame attention injects semantic information from text embeddings into the frame latents.
Self-text attention refines the textual context itself.

Of these, cross-frame attention $\mathbf{A}_{t,l}^{i,j} \in \mathbb{R}^{hw \times hw}$ is of particular interest to our study. Its ability to allow every pixel in one frame to interact with any pixel in another frame is what explicitly enables the model to build and understand temporal relationships across the video sequence.

%% file: figure/ill_attention.tex
\begin{wrapfigure}{r}{0.4\textwidth} 
    \vspace{-12pt}
    \centering
    \includegraphics[width=1\linewidth]{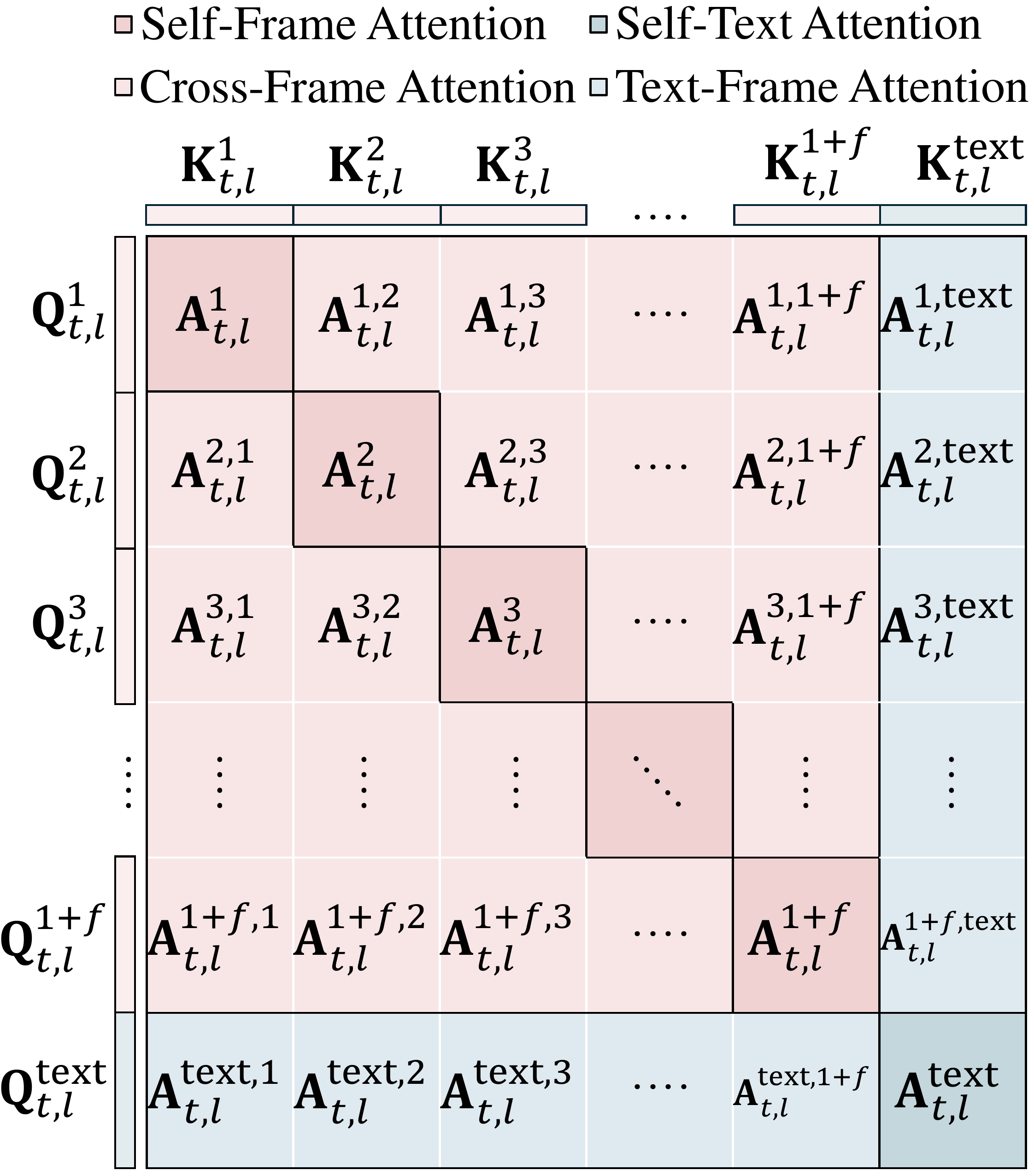}
    \captionsetup{width=1\linewidth}
    \caption{\textbf{Illustration of full 3D attention in video DiTs}, where video frame latents and text embeddings interact.}
    \label{fig:ill_attention}
    \vspace{-30pt}
\end{wrapfigure}

%% file: sec/4_method.tex
\section{DiffTrack}
\label{sec:method}

We introduce \textit{DiffTrack}, a framework to quantify how video DiTs capture temporal correspondence during video generation.
This is crucial yet nontrivial due to the inherent complexity of video DiTs, which involve multiple layers, denoising timesteps, and full 3D attention.
To systematically analyze them, DiffTrack provides an evaluation dataset and novel metrics for assessing the accuracy, confidence, and influence of temporal matching. While we present an in-depth analysis of the open-source video DiT model, CogVideoX-2B~\cite{yang2024cogvideox}, our framework is broadly applicable to other DiT architectures, as demonstrated by our analyses of CogVideoX-5B~\cite{yang2024cogvideox}, HunyuanVideo~\cite{kong2024hunyuanvideo} and CogVideoX-2B-I2V~\cite{yang2024cogvideox} in \cref{supp:other_backbones}.

\input{figure/dataset}

\subsection{Evaluation Dataset}  
To accurately investigate temporal correspondence during video generation, it's crucial to simulate the generation process faithfully. Reconstructing real-world videos often introduces challenges such as inversion errors~\cite{song2020denoising} due to imprecise prompts and distribution shifts from training data. 
To circumvent this, we curate specialized evaluation prompts and generate corresponding video datasets using the specific backbone under analysis.
This approach allows perfect video reconstruction during evaluation, as we use paired prompts with fixed seeds.
For an analysis using real-world videos from DAVIS~\cite{pont20172017}, please refer to \cref{supp:analysis}.

To analyze videos with object and camera motion, we curate two distinct datasets: (1) an object dataset for dynamic object-centric videos and (2) a scene dataset for static scenes with camera motion.
Each dataset includes 50 text prompts with corresponding 50 videos (\eg~$480 {\times} 720$ resolution, $49$ frames, generated by CogVideoX-2B~\cite{yang2024cogvideox}).

To assess temporal matching, the datasets also include pseudo ground truth track annotations.
We predefine a set of starting points $\mathbf{p}^1 \in \mathbb{R}^{N\times2}$ (in latent resolution) in the first frame of each video, where $N$ is the number of points. For the object dataset, we use SAM~\cite{kirillov2023segment} to segment foreground objects and sample grid points with a spacing of ${1}/{20}$ of the video's spatial resolution.
For the scene dataset, we sample a $10 \times 10$ regular grid of points across the entire frame. \cref{fig:dataset} shows examples of each dataset; further details are provided in \cref{sup:dataset}. Since ground-truth tracks are unavailable for generated videos, we use an off-the-shelf tracking method, CoTracker~\cite{karaev2024cotracker}, to obtain pseudo ground truth $\mathbf{T} \in \mathbb{R}^{F \times N \times 2}$ (in original resolution).

\subsection{Temporal Correspondence Estimation within Video DiTs}

To evaluate matching accuracy, we first extract temporal correspondences across video frames using feature descriptors from a video model. We independently establish pairwise correspondences between the first frame and each subsequent frame.

We first construct a matching cost between the descriptors from the first latent and each $j$-th latent, with $j \in \{2, \dots, 1{+}f\}$, at each timestep $t$ and layer $l$:
\begin{equation}
 \mathbf{C}^{1,j}_{t,l} = \mathtt{Softmax} ({\mathbf{D}_{t,l}^1 (\mathbf{D}_{t,l}^j)^T}/{\sqrt{d}}),
 \label{equ:matching_cost}
\end{equation}
where $\mathbf{D}_{t,l}^1$ and $\mathbf{D}_{t,l}^j$ are the feature descriptors corresponding to the first frame latent and the $j$-th latent, and $d$ is the channel dimension of both $\mathbf{D}_{t,l}^1$ and $\mathbf{D}_{t,l}^j$.
$\mathtt{Softmax}(\cdot)$ is applied over the keys for each pixel in the query.
Descriptors $\mathbf{D}_{t,l}$ are internal representation candidates from video DiTs, such as intermediate features after attention blocks or query-key matrices within attention blocks.

Matched correspondence points $\mathbf{p}^j_{t,l} \in \mathbb{R}^{N \times 2}$ at the $j$-th frame latent, timestep $t$, and layer $l$, with the number of points $N$, are obtained through $\mathtt{Argmax}$, which identifies the spatial location $\mathbf{x}$ of the highest value in $\mathbf{C}^{1,j}_{t,l}$ within the spatial domain of the $j$-th latent $\Omega$.
\begin{equation}
    \mathbf{p}^j_{t,l} = \underset{\mathbf{x} \in \Omega}{\mathtt{Argmax}} \ \mathbf{C}^{1,j}_{t,l}(\mathbf{p}^1, \mathbf{x}).
\label{equ:argmax}
\end{equation}

Motion tracks along video frames are obtained by concatenating ($\mathtt{Concat}$) the starting point $\mathbf{p}^1$ with the estimated matches $\mathbf{p}^j_{t,l}$ across the video latent space.
These tracks, in the latent spatial coordinates, are then spatio-temporally upscaled to the original RGB coordinates through linear interpolation ($\mathtt{Interp}$), yielding $\hat{\mathbf{T}}_{t,l} \in \mathbb{R}^{F \times N \times 2}$:
\begin{equation}
 \hat{\mathbf{T}}_{t,l} = \mathtt{Interp}(\mathtt{Concat}(\mathbf{p}^1, \mathbf{p}^2_{t,l}, \dots, \mathbf{p}^{1+f}_{t,l})).
\end{equation}

\subsection{Evaluation Metrics}
Given the estimated matched points, we propose three complementary metrics for evaluating temporal matching in video generation: \textit{matching accuracy}, \textit{confidence score}, and \textit{attention score}. Specifically, matching accuracy measures the precision of estimated tracks across frames. The confidence score quantifies the certainty with which the starting point attends to its estimated match. The attention score reflects the relative strength of cross-frame attention during generation, compared to self-frame and text-frame attention.

\paragrapht{Matching Accuracy.}  
We evaluate point accuracy using the percentage of correct keypoints (PCK) with a predefined error threshold between the estimated track $\hat{\mathbf{T}}_{t,l}$ and visible points in the ground truth $\mathbf{T}$. The \textit{matching accuracy} averages PCK over all visible points across cross-latents and videos.

\paragrapht{Confidence Score.} 
We use the maximum attention values between the first and $j$-th frame latent $\mathbf{A}_{t,l}^{1,j}$ (\cf \cref{sec:dit}) to measure how confidently the starting points $\mathbf{p}^1$ attend to their predicted matched points, formulated as:
\begin{equation}
    \mathbf{M}_{t,l}^{j} = \underset{\mathbf{x} \in \Omega}{\mathtt{Max}} \ \mathbf{A}_{t,l}^{1,j}(\mathbf{p}^1, \mathbf{x}),
\end{equation}
which quantifies the attention of $\mathbf{p}^1$ to the estimated point in the $j$-th latent at timestep $t$ and layer $l$.
$\mathtt{Max}$ takes the maximum attention value over all spatial locations $\mathbf{x} \in \Omega$.
The \textit{confidence score} is the average of these maximum values across all cross-latents with $j \in \{2, \dots, 1{+}f\}$, all visible points in ground truth, and videos.

\paragrapht{Attention Score.} 
We use the sum of cross-attention values across all cross-frame latents. This allows us to assess the influence of cross-frame interactions during video generation, in comparison to other types of attention, such as text-frame or self-frame attention. This is formulated as:
\begin{equation}
\mathbf{S}_{t,l} = \sum_{j \in \mathcal{F}} \sum_{\mathbf{x} \in \Omega} \mathbf{A}_{t,l}^{1,j}(\mathbf{p}^1, \mathbf{x}),
\end{equation}
which quantifies the sum of attention from $\mathbf{p}^1$ across cross-frames at timestep $t$ and layer $l$, over all spatial locations $\mathbf{x} \in \Omega$ and all cross-frame indices $j \in \mathcal{F}$, where $\mathcal{F} = [2, 1{+}f]$. The \textit{attention score} is the average of these summation values across all visible points in the ground truth and all videos.

Three metrics must be considered together (as detailed in \cref{sec:method:analysis}), as none alone is sufficient to ensure temporal matching. For instance, even with high matching accuracy across frames, it may not influence the generation process when attention scores from others (\eg, self-frame or text-frame) are higher. In another case, high confidence indicates the certainty of the matching score but does not ensure the correctness of the match. We thus compute the harmonic mean of the normalized matching accuracy, confidence score, and attention score to identify instances where all three metrics are high.

\input{figure/analysis}

\subsection{Analysis}
\label{sec:method:analysis}
With DiffTrack, we systematically analyze CogVideoX-2B~\cite{yang2024cogvideox} in the context of temporal correspondence.
For all analyses, we use full 3D attention in the model, which consists of 30 layers with 50 denoising timesteps.

Our analysis considers the following three perspectives.
\textit{Representation selection} compares intermediate features and query-key representations.
\textit{Layer-wise analysis} explores how well temporal matching is encoded at different depths within the attention blocks.
\textit{Noise-level analysis} examines how temporal relationships evolve throughout the denoising process.
Further in-depth analysis is provided in \cref{supp:analysis}. Additional analyses of CogVideoX-5B~\cite{yang2024cogvideox}, HunyuanVideo~\cite{kong2024hunyuanvideo} and CogVideoX-2B-I2V~\cite{yang2024cogvideox} are provided in \cref{supp:other_backbones}.

\paragrapht{Representation Selection.} 
\cref{fig:analysis}(a) compares the accuracy of intermediate feature matching, where features are extracted after each attention layer, and query-key matching, where queries and keys are obtained within each attention layer.
Our results indicate that \textit{query-key matching consistently outperforms intermediate feature matching}. This finding aligns with prior works~\cite{an2024cross, nam2024dreammatcher}, in which query-key matching captures geometric relationships for correspondence, while values contain visual appearance, potentially diluting geometric cues for accurate matching.

\paragrapht{Layer-wise Analysis.}
\cref{fig:analysis}(b) presents the harmonic mean of query-key matching across all timesteps and layers to identify which feature descriptors at which layer and timestep play a predominant role in temporal matching. We observe that the top-$20$ scores (red) originate from the same layer, indicating that \textit{a specific layer predominantly governs temporal correspondence}.
This behavior is further observed in the top-$50$ scores (green): a limited set of layers drives temporal matching.

\input{figure/graph_attention}

\paragrapht{Noise-level Analysis.} 

\cref{fig:analysis}(c) presents the harmonic mean of query-key matching across timesteps in the selected layers from \cref{fig:analysis}(b), which are identified as leading layers for temporal matching.
\textit{Temporal matching improves during the denoising process but slightly degrades toward the end.} This is because earlier timesteps (1) contain noisier latents, which hinder precise temporal matching, and (2) rely heavily on text embeddings and self-frame attention to establish the overall video semantics and layout. Toward the final timesteps, (3) self- and text-frame attention slightly increase again to refine the remaining appearance details, once the motion in the synthesized video has been established.

This observation is supported by \cref{fig:graph_attention}, which shows the attention scores for text-frame, self-frame, and cross-frame attention along timesteps.
After the early denoising steps, text-frame and self-frame attention remain low, reducing the influence of textual guidance.
Meanwhile, cross-frame attention remains the most influential, enhancing cross-frame coherence, with a slight drop near the end of the timesteps.

\input{figure/combined_metrics}

\paragrapht{Additional Analysis on Metrics.}
We further emphasize the importance of jointly considering three metrics, matching accuracy, confidence score, and attention score. \cref{fig:metric_1}(a) presents the top-$20$ timesteps and layers where the confidence score is high, but matching accuracy is low. We find that specific layers exhibit this discrepancy as they are overwhelmed by positional information induced by positional embeddings at each timestep. In \cref{fig:metric_1}(c), PCA visualization of queries and keys in these layers reveals a dominance of positional cues, while \cref{fig:metric_1}(b) shows that in these layers, matching cost visualizations indicate that points in the first frame tend to match exactly with their initial locations in other frames.
This suggests that these points are not correctly matched to their actual counterparts but instead strongly attend to the same spatial location across frames, reflecting the impact of positional bias.

Additionally, \cref{fig:metric_2}(a) presents the top-$20$ timesteps and layers where matching accuracy and confidence scores are high, but attention scores are low. We observe that this property is exhibited in certain layers. As shown in \cref{fig:graph_attention}, this occurs because text-frame attention remains highly active in these layers, maintaining a value around 0.5, unlike in other layers where text-frame attention drops below 0.2 (\cf \cref{fig:graph_attention}). This reduces the attention scores, which in turn limits the influence of cross-frame interactions during the generation process.

\section{DiffTrack for \textit{Zero-Shot} Point Tracking}
\label{sec:difftrack_zeroshot}
DiffTrack enables the joint extraction of motion trajectories and video generation, selecting the optimal layer and timestep based on matching accuracy. We demonstrate this in zero-shot point tracking~\cite{aydemir2024can} on real videos, without training specialized architectures~\cite{doersch2022tap} or fine-tuning video diffusion models~\cite{jeong2024track4gen}. To achieve this, we use the inverted noise of real videos obtained through DDIM inversion~\cite{song2020denoising} at the selected timestep and extract features from the chosen layer. Notably, the inversion error is negligible, as we use the final timestep $t=1$ based on our analysis of matching accuracy in \cref{fig:analysis}(a). However, this still faces challenges such as temporal context loss from 3D VAE compression and handling of long-term video sequences. We address these challenges below. The overall architecture and its details are provided in \cref{sup:zero-shot-tracking}.

\paragrapht{Temporal Compression in 3D VAE.}
As discussed in \cref{pre:VDM}, the 3D VAE temporally compresses video frames into a single-frame latent with a compression ratio $q$. While linear interpolation can recover motion trajectories from the latent to the RGB video space, it often fails to capture per-frame motion details, limiting accuracy. To mitigate this, we set $q{=}1$ to establish a direct one-to-one mapping between each video frame and its latent, thereby avoiding temporal compression and enabling precise tracking. In \cref{supp:analysis}, we demonstrate that the one-to-one mapped latents can still reconstruct the original videos using the 3D VAE decoder.

\paragrapht{Long-term Video Sequences.}
The fixed temporal resolution of pre-trained video models (\eg 49 frames in CogVideoX-2B~\cite{yang2024cogvideox}) limits their ability to model long-term contexts. Naively splitting and processing video chunks separately disrupts direct temporal correspondence with the global first frame. To address this, we construct each chunk to include the global first frame, maintaining a direct temporal connection while interleaving subsequent frames to minimize large motion changes.

\input{figure/main_qual}

\input{table/main_quan}

\subsection{Experimental Settings}
\label{sec:exp}
Implementation details and ablation studies are provided in \cref{sup:zero-shot-tracking-imp} and \cref{supp:abl_zeroshot}. We evaluate zero-shot tracking on two real-video datasets with precisely annotated tracks: TAP-Vid-DAVIS~\cite{doersch2022tap} and TAP-Vid-Kinetics~\cite{doersch2022tap}, following~\cite{kim2025exploring}. We measure the position accuracy of estimated tracks as the percentage of predicted points within thresholds from visible ground-truth points. We adopt five threshold levels~\cite{kim2025exploring, cho2024local, doersch2022tap, karaev2024cotracker}: $\delta^0, \delta^1, \delta^2, \delta^3, \delta^4$, corresponding to pixel distances of 1, 2, 4, 8, and 16, respectively, and report the average accuracy across all thresholds as $\delta^x_{\text{avg}}$. Starting points are sampled from the first frame as in~\cite{aydemir2024can}.

\subsection{Experimental Results}
We compare our method with existing vision foundation models trained on single images~\cite{caron2021emerging, darcet2023vision, oquab2023dinov2, tang2023emergent, rombach2022high} and self-supervised models trained on two-view images~\cite{an2024cross} or videos~\cite{qian2023semantics, jabri2020space, xu2021rethinking, li2023spatial, blattmann2023stable}. As shown in \cref{tab:query-first}, our approach achieves superior performance on both the Kinetics and DAVIS datasets, ultimately obtaining the highest average accuracy in $\delta^x_{\text{avg}}$.
The results highlight our in-depth analysis of temporal matching within the full 3D attention mechanism of video DiTs.

\cref{fig:main_qual} shows predicted motion trajectories on the DAVIS dataset, alongside qualitative comparisons between our method and prior approaches~\cite{oquab2023dinov2,xu2021rethinking}. Unlike previous methods, which struggle with temporal dynamics and often yield inconsistent tracks, DiffTrack on CogVideoX-2B~\cite{yang2024cogvideox} produces smoother and more accurate trajectories. Additional qualitative results are provided in \cref{supp:add_qual}.

\input{figure/figure_CAG}

\section{DiffTrack for \textit{Motion-Enhanced} Video Generation}
\label{sec:CAG}
We extend our findings to generate motion-enhanced videos by improving temporal correspondence within full 3D attention. As illustrated in \cref{fig:figure_CAG}, we propose \textit{Cross-Attention Guidance (CAG)}, a novel diffusion guidance technique applied at specific layers identified in \cref{fig:analysis}(b), steering video generation toward motion-enhanced samples. Unlike prior works~\cite{jeong2024track4gen, chefer2025videojam} that require large-scale video-trajectory training pairs, CAG requires no additional training, external conditions, or auxiliary modules, and operates entirely within the existing diffusion framework. 

\input{figure/main_qual_guidance}

Inspired by PAG~\cite{ahn2024self}, CAG simulates degraded motion by zeroing out selected cross-frame attention maps (\eg $l=13, 17, 21$ in CogVideoX-2B, based on the harmonic mean in \cref{fig:analysis}(b)), then guides the diffusion model away from these degraded samples, promoting temporally coherent video generation. This can be formulated as:
\begin{equation}
\tilde{\boldsymbol{\epsilon}}_\theta(\mathbf{z}_{\mathrm{video}, t}, \mathbf{z}_{\mathrm{text}}, t) 
= 
\boldsymbol{\epsilon}_\theta(\mathbf{z}_{\mathrm{video}, t}, \mathbf{z}_{\mathrm{text}}, t) 
+ s \cdot \left( \boldsymbol{\epsilon}_\theta(\mathbf{z}_{\mathrm{video}, t}, \mathbf{z}_{\mathrm{text}}, t) 
- \hat{\boldsymbol{\epsilon}}_\theta(\mathbf{z}_{\mathrm{video}, t}, \mathbf{z}_{\mathrm{text}}, t) \right),
\end{equation}
where $\boldsymbol{\epsilon}_\theta(\cdot)$ is the standard noise prediction at timestep $t$, conditioned on the text. $\hat{\boldsymbol{\epsilon}}_\theta(\cdot)$ denotes the noise prediction from a perturbed forward pass, where cross-frame attention maps $\mathbf{A}_{t,l}^{i,j}$ in selected layers are zeroed out to simulate motion degradation, producing $\hat{\mathbf{A}}_{t,l}$. $s$ is the guidance scale, and the final guided prediction $\tilde{\boldsymbol{\epsilon}}_\theta(\cdot)$ steers the model to denoise away from motion-degraded samples.

\subsection{Experimental Settings}
\label{sec:exp_guidance}
Further implementation and evaluation details are provided in \cref{supp:CAG}, and ablation studies are included in \cref{supp:abl_guidance}. We evaluate CAG against its baselines, CogVideoX-2B and CogVideoX-5B, using both automatic metrics and human evaluation.

For automatic metrics, we report temporal quality metrics: {Subject Consistency}, {Background Consistency}, and {Dynamic Degree} from Vbench~\cite{huang2024vbench}. Subject Consistency and Background Consistency measure the temporal coherence of subject and background appearance, respectively. While static scenes can achieve high scores on these metrics, we additionally calculate Dynamic Degree to quantify motion dynamics. We also report a frame-wise quality metric from Vbench, {Imaging Quality}, which detects frame-wise distortions.

For human evaluation, we follow the Two-Alternative Forced Choice (2AFC) protocol~\cite{chefer2025videojam, rombach2022high, blattmann2023stable}, where each rater compares a video from the baseline with a video from our method (baseline + CAG), and selects one based on overall video quality, motion, and text-video alignment. We collect 750 responses from a total of 25 participants for each baseline.

\input{table/guidance_quan}

\subsection{Experimental Results}
In \cref{tab:cag_comparison}, CAG outperforms the baseline across all human evaluations. CAG  achieves higher scores in Subject Consistency, Background Consistency, Dynamic Degree, and Imaging Quality on VBench, indicating that our guidance improves motion dynamics while enhancing motion consistency and overall video fidelity.

In \cref{fig:guidance_qual}, the baseline often fails to generate consistent and natural motion, resulting in physically implausible motion (\eg drinking beer in the first row) or blurry and disjointed appearances (\eg a blurred and fragmented human body in the second row). In contrast, CAG effectively corrects these artifacts by enhancing temporal matching and motion consistency, ensuring that corresponding physical points remain coherent across frames. More qualitative results are provided in \cref{supp:add_qual}.

%% file: figure/dataset.tex
\begin{figure*}
    \centering
\includegraphics[width=1.0\linewidth]{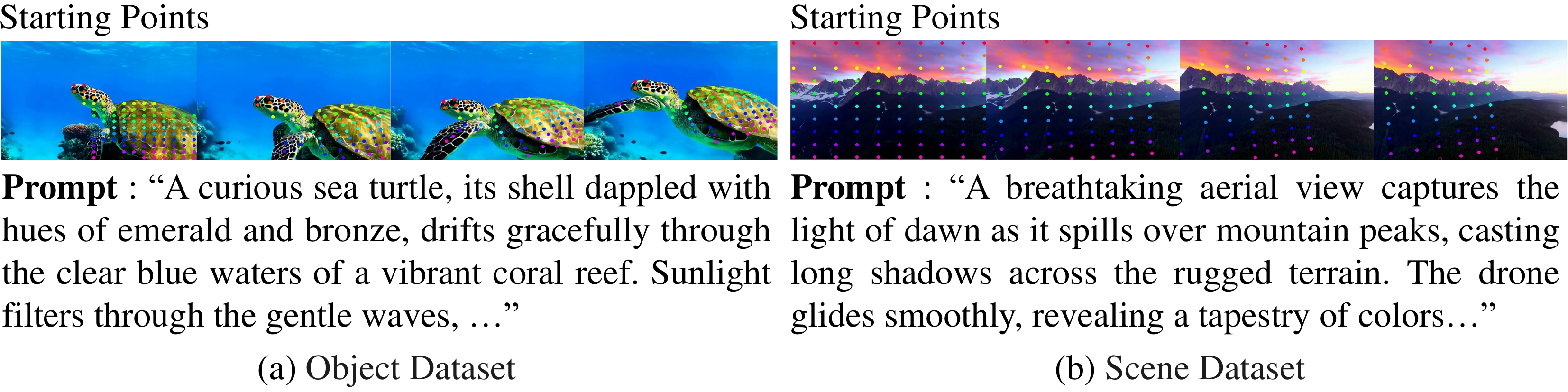}
    \vspace{-15pt}
    \caption{\textbf{Our curated evaluation dataset} includes: (a) an object dataset for dynamic object-centric videos and (b) a scene dataset for static scenes with camera motion. Each dataset comprises 50 prompt-generated video pairs per video generative model (\eg CogVideoX-2B~\cite{yang2024cogvideox}). In the benchmark, we predefine starting points in the first frame and obtain pseudo ground-truth trajectories using an off-the-shelf tracking method~\cite{karaev2024cotracker}.}
    \label{fig:dataset}
    \figureaftercaption
\end{figure*}

%% file: figure/analysis.tex
\begin{figure*}[t!]
    \centering
\includegraphics[width=0.9\linewidth]{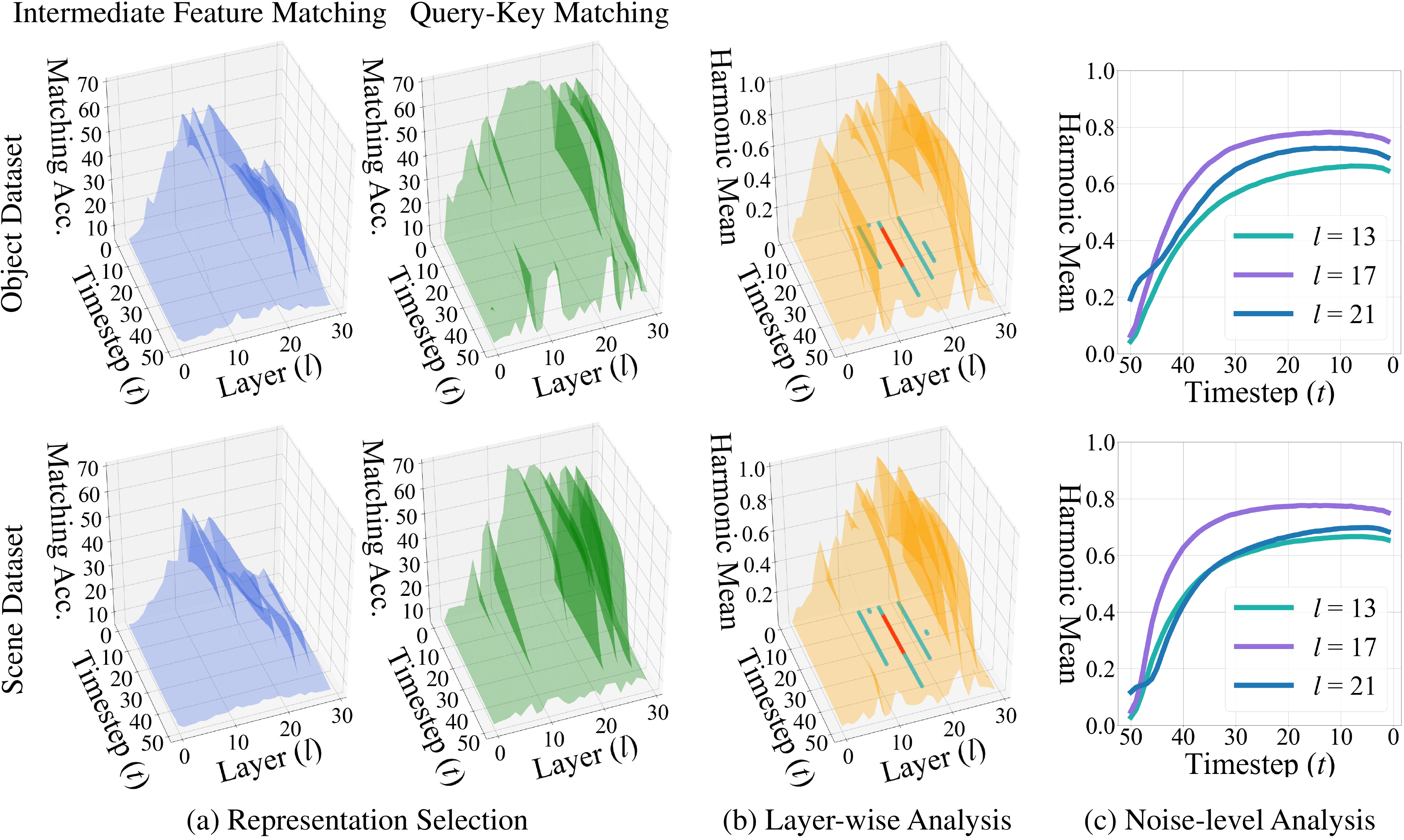}
    \vspace{-3pt}
    \caption{\textbf{Analysis of temporal matching in CogVideoX-2B~\cite{yang2024cogvideox}.} (a) Query-key matching outperforms intermediate feature matching, highlighting the effectiveness of cross-frame interactions in 3D attention. (b) The harmonic mean of query-key matching shows that temporal matching is primarily driven by a few specific layers.  (c) Temporal matching improves progressively during the denoising but slightly degrades near the final steps.}
    \label{fig:analysis}
    \figureaftercaption
\end{figure*}

%% file: figure/graph_attention.tex
\begin{wrapfigure}{r}{0.45\textwidth}
    \centering
    \vspace{-12pt}
    \includegraphics[width=1\linewidth]{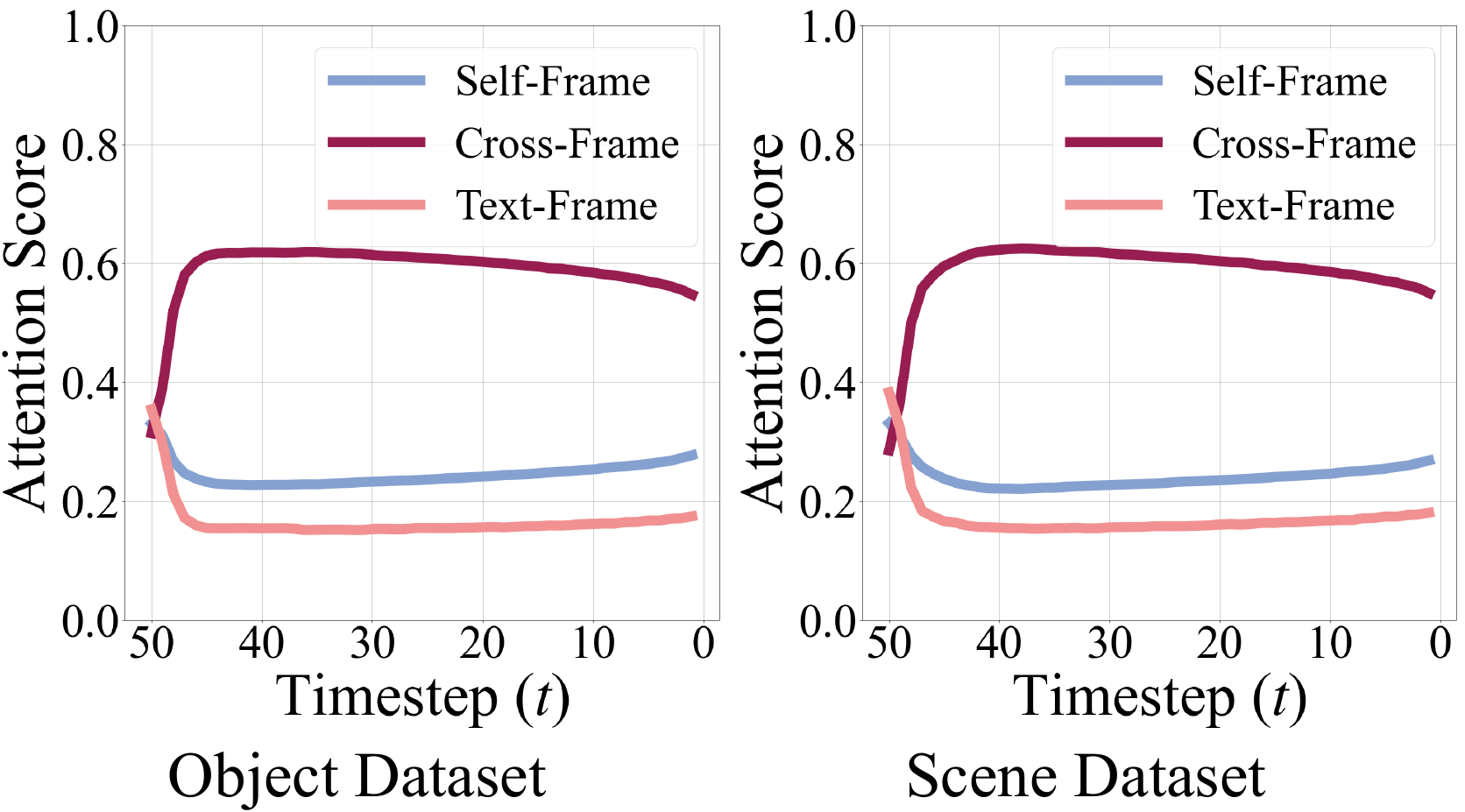} 
    \vspace{-16pt}
    \captionsetup{width=1\linewidth}
    \caption{\textbf{Evolution of attention scores across timesteps.}}
    \label{fig:graph_attention}
    \vspace{-10pt}
\end{wrapfigure}

%% file: figure/combined_metrics.tex
\begin{figure}[t] 
    \centering
    \begin{minipage}{0.55\textwidth} 
        \centering
        \includegraphics[width=\linewidth]{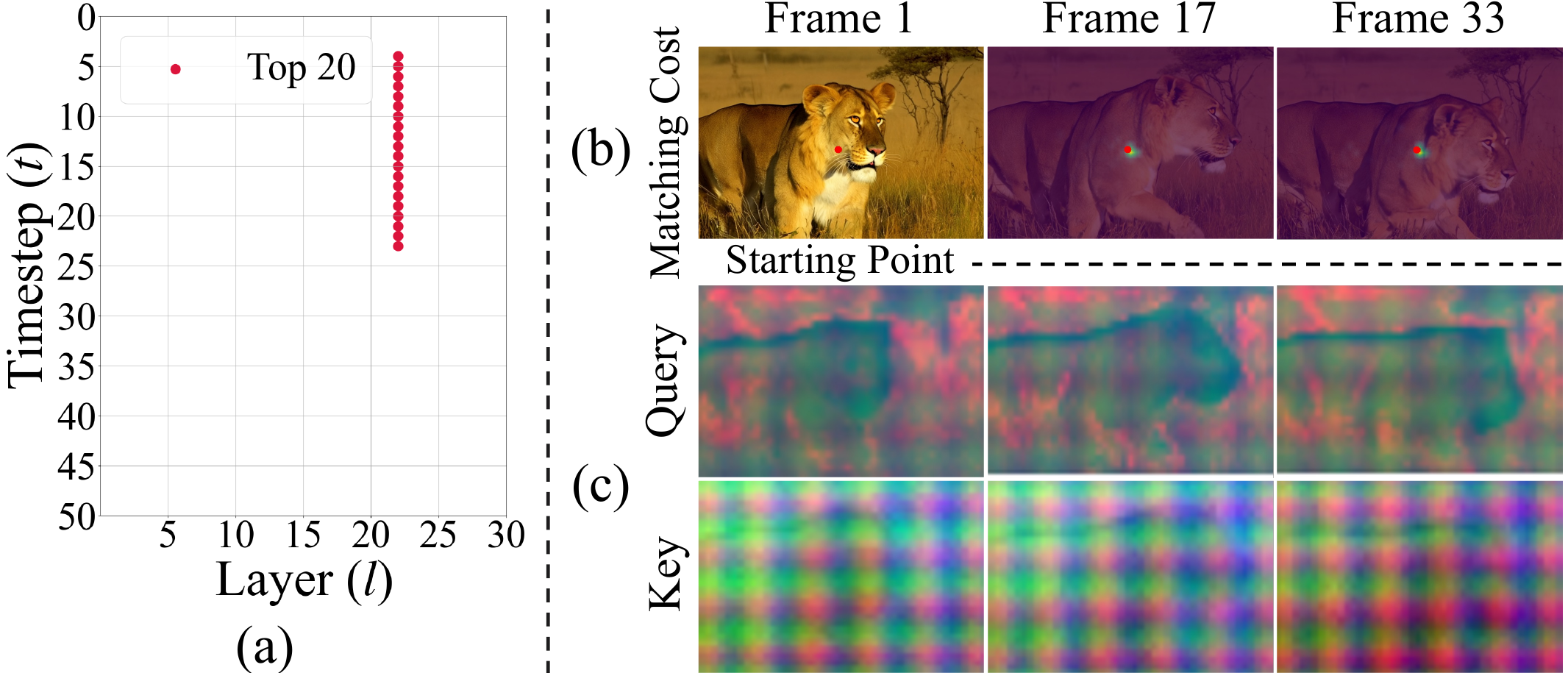}
        \caption{\textbf{Analysis of positional bias in temporal matching.}}
        \label{fig:metric_1}
    \end{minipage}
    \hspace{0.02\textwidth} 
    \begin{minipage}{0.41\textwidth} 
        \centering
        \includegraphics[width=\linewidth]{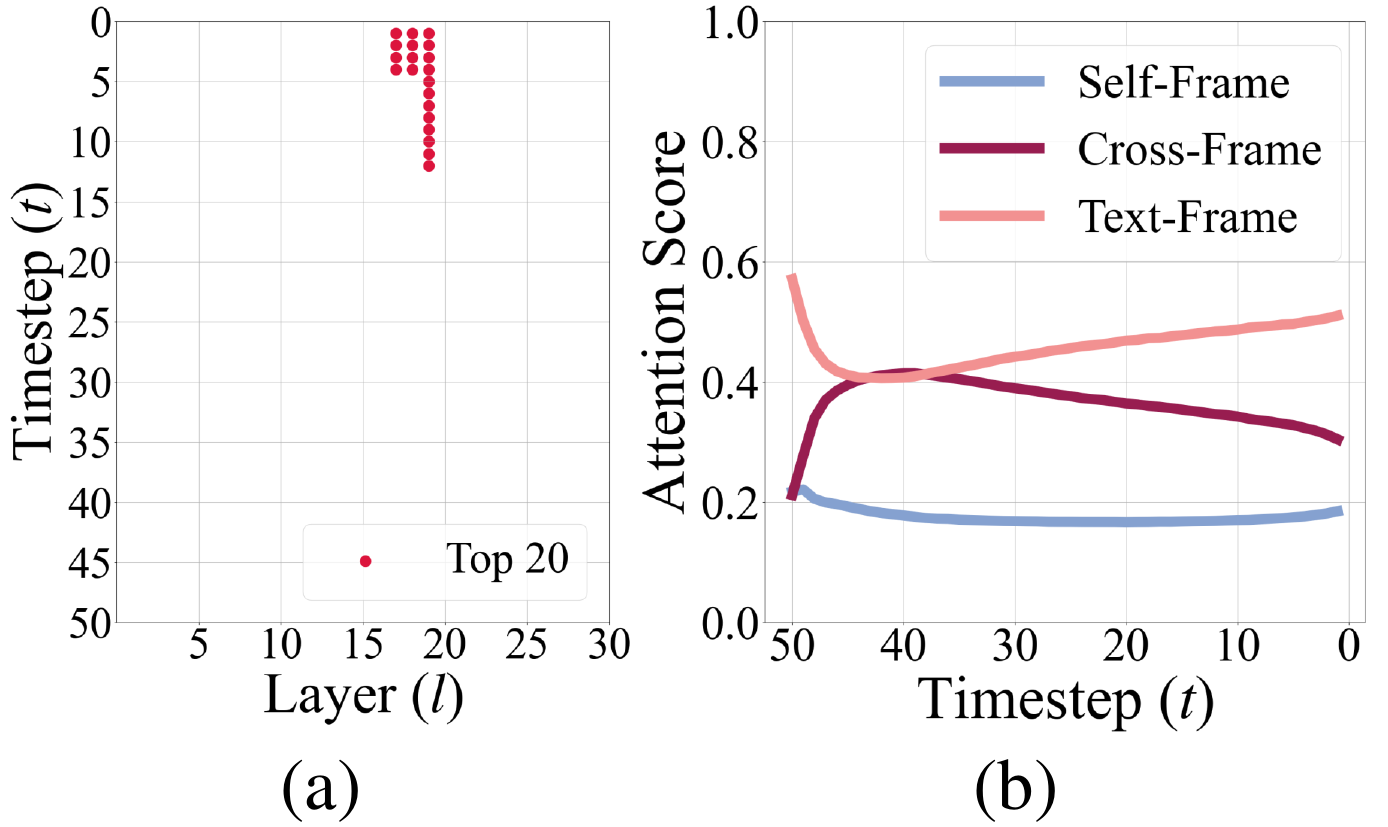}
        \caption{\textbf{Impact of persistent text-frame attention on attention score.}}
        \label{fig:metric_2}
    \end{minipage}
\end{figure}

%% file: figure/main_qual.tex
\begin{figure*}[t!]
    \centering
\includegraphics[width=1\linewidth]{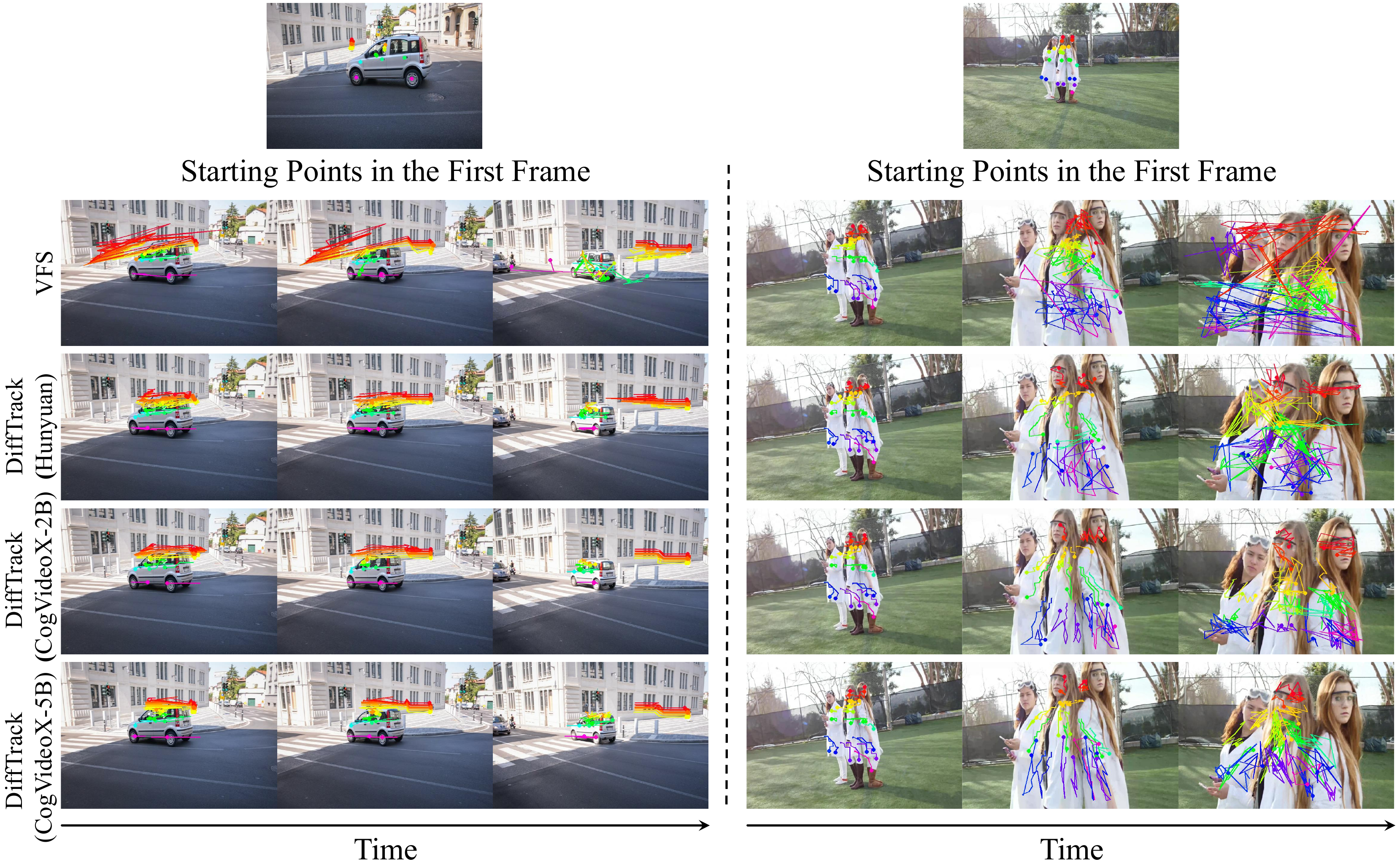}
    \vspace{-15pt}
    \caption{\textbf{Qualitative comparison.} CogVideoX-2B~\cite{yang2024cogvideox} combined with DiffTrack produces smoother and more accurate trajectories than DINOv2~\cite{oquab2023dinov2} and VFS~\cite{xu2021rethinking}, which struggle with temporal dynamics and often yield inconsistent tracks.}
    \label{fig:main_qual}
    \figureaftercaption
\end{figure*}

%% file: table/main_quan.tex
\begin{table*}
\centering
\resizebox{1\linewidth}{!}{

    \begin{tabular}{l|cccccc|cccccc}
    \toprule
    \multirow{2}{*}{Backbone} & \multicolumn{6}{c|}{Kinetics} & \multicolumn{6}{c}{DAVIS} \\
    & $<\delta^{0}$   & $<\delta^{1}$   & $<\delta^{2}$   & $<\delta^{3}$   & $<\delta^{4}$   & $<\delta^{x}_\text{avg}$   & $<\delta^{0}$   & $<\delta^{1}$   & $<\delta^{2}$   & $<\delta^{3}$   & $<\delta^{4}$    & $<\delta^{x}_\text{avg}$\\
    \midrule
    \midrule
    DINO (ViT-B/16)~\cite{caron2021emerging}& 2.8 & 10.9 & 33.9 & 58.2 & 74.8 & 36.1 & 2.9 & 10.7 & 34.3 & 59.8 & 77.0 & 37.3 \\ %
    DINOv2-Reg (ViT-B/14)~\cite{darcet2023vision}& 2.8 & 10.8 & 32.8 & 60.6 & 78.7 & 37.1 & 3.0 & 11.7 & 36.5 & 67.5 & 84.1 & 40.6 \\ %
    DINOv2 (ViT-B/14)~\cite{oquab2023dinov2}& 3.0 & 11.4 & 34.6 & 63.0 & \underline{80.3} & 38.4 & 3.1 & 12.1 & 38.7 & 70.7 & \textbf{85.8} & 42.1 \\ %
     DIFT (SD1.5)~\cite{tang2023emergent}& 3.7 & 14.6 & 44.6 & 69.0 & 77.5 & 41.9 & 3.5 & 13.0 & 39.3 & 63.1 & 72.2 & 38.2 \\ %
     DIFT (SD2.1)~\cite{tang2023emergent}& 3.7 & 14.9 & 45.4 & 70.9 & 79.6 & 42.9 & 3.6 & 13.3 & 40.1 & 65.8 & 75.7 & 39.7 \\ %
     \midrule
     SMTC (ViT-S/16)~\cite{qian2023semantics} & 4.1 & 15.5 & 34.2 & 54.4 & 72.1 & 36.1 & 2.6 & 12.1 & 29.4 & 52.5 & 73.0 & 33.9 \\
     CRW (ResNet-18)~\cite{jabri2020space}& 5.2 & 19.4 & 42.7 & 62.9 & 74.3 & 40.9 & 3.1 & 13.9 & 34.7 & 57.1 & 70.5 & 35.9 \\
     
     Spa-then-Temp (ResNet-50)~\cite{li2023spatial} & 5.3 & 19.4 & 41.6 & 58.9 & 69.7 & 39.0 & 3.2 & 13.8 & 33.1 & 53.4 & 67.5 & 34.2 \\

    VFS (ResNet-50)~\cite{xu2021rethinking} & 5.4 & 20.1 & 44.6 & 65.4 & 76.6 & 42.4 & 3.5 & 15.2 & 37.2 & 60.8 & 75.2 & 38.4 \\
     
     SVD~\cite{blattmann2023stable} & 4.3 & 16.0 & 37.9 & 56.3 & 69.8 & 36.6 & 3.6 & 14.6 & 34.1 & 55.7 & 71.4 & 35.9 \\
     ZeroCo (CroCo)~\cite{an2024cross} & \textbf{14.5} & 22.9 & 35.9 & 60.4 & 79.7 & 42.6 & 4.6 & 8.8 & 19.5 & 44.9 & 65.6 & 28.7 \\

    \midrule
    
    \textbf{DiffTrack (HunyuanVideo~\cite{kong2024hunyuanvideo})}
    & 5.9 & 22.0 & 49.1 & 70.4 & \underline{80.3} & 45.5 & 4.4 & 18.2 & 44.8 & 70.1 & 82.8 & 44.1 \\
    \textbf{DiffTrack (CogVideoX-2B~\cite{yang2024cogvideox}}) 
    & 6.2 & \underline{23.3} & \underline{51.2} & \underline{71.2} & 79.9 & \underline{46.3} & \underline{4.8} & \underline{19.4} & \underline{49.2} & \underline{73.6} & \underline{84.3} & \underline{46.3} \\
    \textbf{DiffTrack (CogVideoX-5B~\cite{yang2024cogvideox}}) 
    & \underline{6.8} & \textbf{25.9} & \textbf{55.4} & \textbf{74.9} & \textbf{82.7} & \textbf{49.2} & \textbf{5.2} & \textbf{20.5} & \textbf{50.7} & \textbf{73.9} & \underline{84.3} & \textbf{46.9} \\
    
    \bottomrule
    \end{tabular}
}
\tablebeforecaption
\caption{\textbf{Quantitative comparison on the TAP-Vid datasets~\cite{doersch2022tap}.} Video DiTs~\cite{yang2024cogvideox, kong2024hunyuanvideo} combined with DiffTrack outperform all vision foundation models trained on single images and self-supervised models trained on two-view images or videos for zero-shot tracking.}
\label{tab:query-first}
\tableaftercaption
\end{table*}

%% file: figure/figure_CAG.tex
\begin{wrapfigure}{r}{0.5\textwidth} 
    \centering
    \vspace{-12pt}
    \includegraphics[width=1\linewidth]{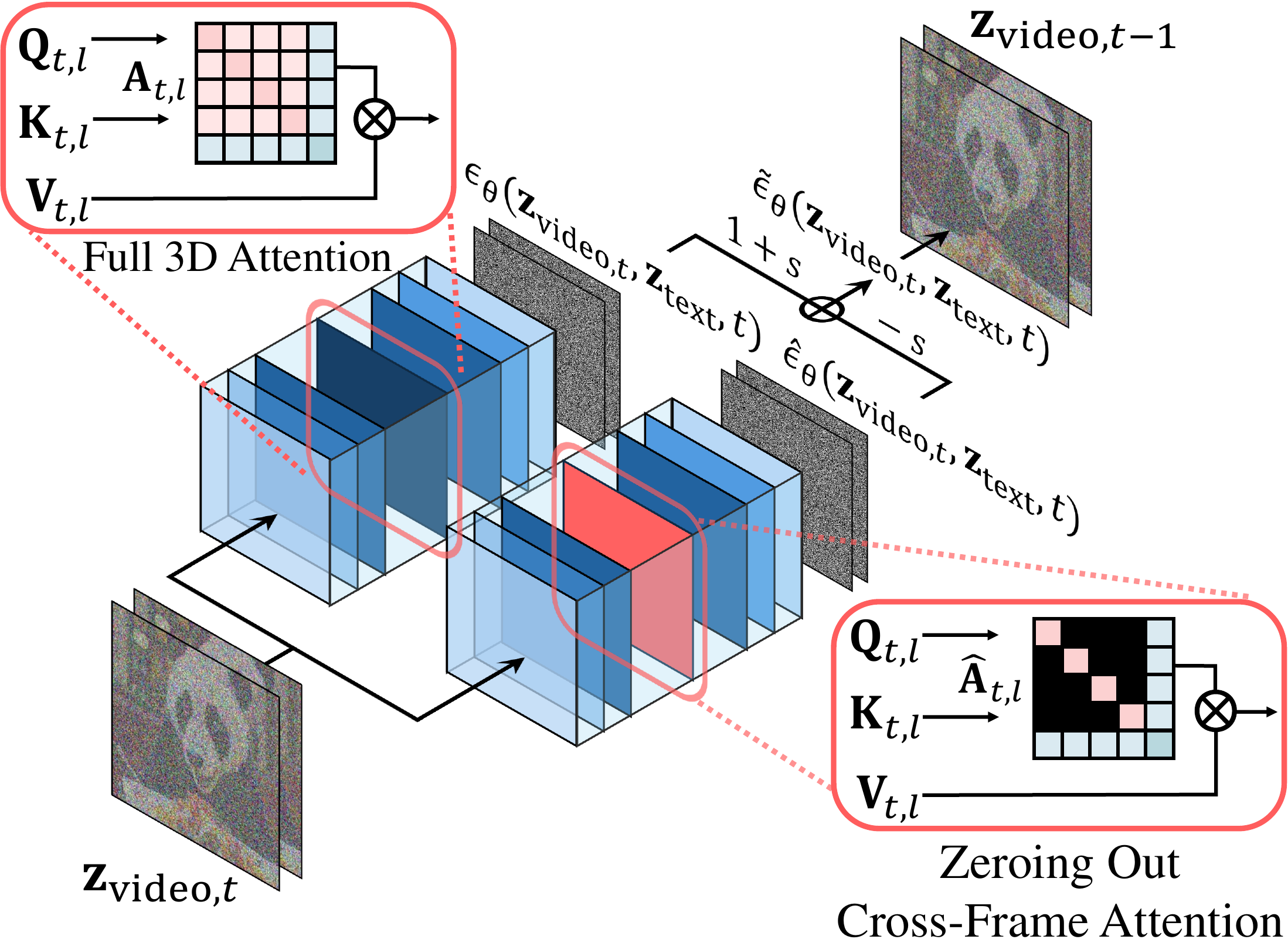} 
    \captionsetup{width=1\linewidth}
    \caption{\textbf{Overall architecture of CAG.}}
    \label{fig:figure_CAG}
   
    \vspace{-15pt}
\end{wrapfigure}

%% file: figure/main_qual_guidance.tex
\begin{figure*}[t!]
    \centering
\includegraphics[width=1\linewidth]{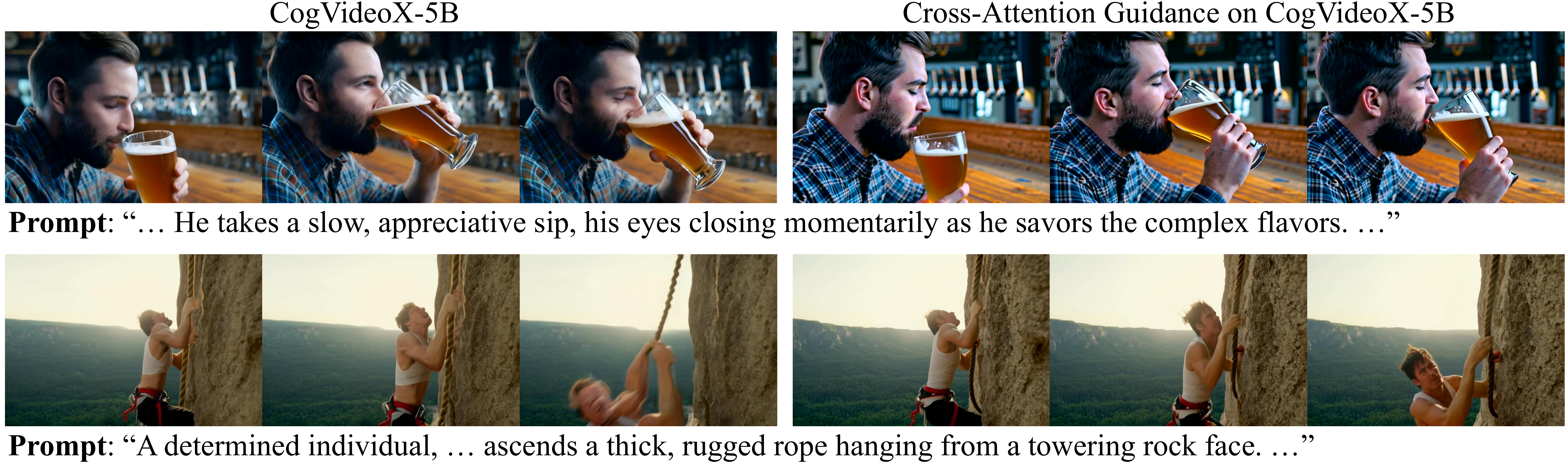}
    \vspace{-15pt}
    \caption{\textbf{Qualitative comparison with baseline~\cite{yang2024cogvideox} and CAG.} CAG enhances temporal matching and corrects motion inconsistencies in the synthesized videos.}
    \label{fig:guidance_qual}
    \figureaftercaption
\end{figure*}

%% file: table/guidance_quan.tex
\begin{wraptable}{r}{0.63\textwidth}  
\centering
\vspace{-12pt}  
\resizebox{1\linewidth}{!}{ 
\begin{tabular}{l|ccc|cccc}
    \toprule
   \multirow{3}{*}{\textbf{Method}} & \multicolumn{3}{c|}{\textbf{Human Eval}} & \multicolumn{4}{c}{\textbf{Auto. Metrics}} \\
     & \shortstack{Video\\Quality} & \shortstack{Motion\\Fidelity} & 
     \shortstack{Text\\Faithfulness} &
     \shortstack{Subject\\Consistency} & \shortstack{Background\\Consistency} & \shortstack{Dynamic\\Degree} & \shortstack{Imaging\\Quality} \\
    \midrule
   CogVideoX-2B & 12.85 & 14.99 & 18.56 & 0.9276 & 0.9490 & {0.7917} & 0.5657 \\
    CogVideoX-2B + \textbf{CAG} & \textbf{87.16} & \textbf{85.01} & \textbf{81.44} & \textbf{0.9313} & \textbf{0.9564} & \textbf{0.8235} & \textbf{0.6054} \\
    \midrule
    CogVideoX-5B & 39.10 & 40.91 & 30.00 &0.9158&0.9590&0.6667&0.5531\\
    CogVideoX-5B + \textbf{CAG} & \textbf{60.90} & \textbf{59.09}& \textbf{70.00} &\textbf{0.9283}&\textbf{0.9644}&\textbf{0.6863}& \textbf{0.6051}\\
    \bottomrule
\end{tabular}
}
\vspace{-5pt}  
\caption{{\textbf{Quantitative results of CAG.} Human evaluation reports the percentage of votes.}}
\vspace{-10pt}
\label{tab:cag_comparison}
\end{wraptable}

%% file: sec/5_con.tex
\section{Conclusion}
\label{sec:con}
We introduce DiffTrack, a framework for analyzing temporal correspondence in video DiTs, revealing how these models establish temporal correspondences during video generation. Our analysis identifies the crucial role of query-key similarities within specific layers of the full 3D attention mechanism and shows that temporal matching strengthens during denoising. We demonstrate the practicality of DiffTrack in zero-shot point tracking and motion-enhanced video generation, paving the way for leveraging temporal understanding in downstream tasks and improving video generation quality.

%% file: sec/suppl.tex
\clearpage
\appendix
\renewcommand{\thepage}{A.\arabic{page}}
\renewcommand{\thesection}{\Alph{section}}
\renewcommand{\thesubsection}{\Alph{section}.\arabic{subsection}}
\renewcommand{\thefigure}{A.\arabic{figure}}
\renewcommand{\thetable}{A.\arabic{table}}
\setcounter{page}{1}
\setcounter{figure}{0}
\setcounter{table}{0}

\section*{\Large Appendix}

\section{Dataset Curation Details}
\label{sup:dataset} 
In this section, we provide further details on the curation of our evaluation dataset. In \cref{fig:fg_prompt} and \cref{fig:bg_prompt}, we present 50 prompts per dataset for analysis on CogVideoX-2B. Additionally, we provide more dataset examples in \cref{fig:add_dataset_ex}.

\paragrapht{Curating Prompt-Video Pairs.} 
To analyze motion trajectories in depth, we require high-quality prompt-video pairs to ensure motion consistency and video fidelity, as suboptimal videos could introduce noise into our analysis.  
To achieve this, we begin with collecting GPT-enhanced prompts from VBench~\cite{huang2024vbench}, a benchmark designed for evaluating video generative models.  
For the object dataset, we gather prompts from animal categories, while for the scene dataset, we collect prompts from architecture and lifestyle categories.  
We further augment both datasets by generating additional prompts using GPT-4o~\cite{achiam2023gpt}, with VBench prompts as references, ultimately collecting 300 prompts for both the object and scene datasets.  
We then synthesize videos from each prompt using video generative models under analysis (\eg CogVideoX-2B~\cite{yang2024cogvideox}).
To simulate the generation process, we curate a distinct evaluation dataset for each analyzed model (CogVideoX-2B~\cite{yang2024cogvideox}, CogVideoX-5B~\cite{yang2024cogvideox}, HunyuanVideo~\cite{kong2024hunyuanvideo}, and CogVideoX-2B-I2V~\cite{yang2024cogvideox}).
For CogVideoX-2B~\cite{yang2024cogvideox}, each video has a resolution of $480 \times 720$ and consists of 49 frames.
Human annotators carefully select the final prompt-generated video pairs based on motion consistency and overall video fidelity, resulting in the top-$50$ pairs for each dataset.

\paragrapht{Generating Pseudo Ground-Truth.}  
To evaluate temporal matching, our dataset includes pseudo ground-truth trajectories, as no ground-truth trajectories exist for synthesized videos.  
We generate these using the off-the-shelf point tracking method CoTracker~\cite{karaev2024cotracker}. Specifically, we first define the starting points of the trajectories in the first frame.  
For the object dataset, which focuses on object dynamics, we segment the centered object using SAM~\cite{kirillov2023segment} and sample grid points at $1/20$ of the video's spatial resolution.  
For the scene dataset, where we aim to capture overall camera movement, we uniformly sample a $10\times10$ grid across the entire frame.  
Using CoTracker, we obtain per-point trajectories along with their visibility.  
For our analysis, we consider only points that CoTracker estimates as visible.

\clearpage
\section{DiffTrack on Other Video DiTs} 
\label{supp:other_backbones}
DiffTrack is compatible with any off-the-shelf video DiT architecture. We extend our analysis to additional DiT-based video models, including CogVideoX-5B~\cite{yang2024cogvideox}, HunyuanVideo~\cite{kong2024hunyuanvideo} and CogVideoX-2B-I2V~\cite{yang2024cogvideox}.

\paragrapht{DiffTrack on CogVideoX-5B.} In \cref{fig:analysis_5b}, we present an analysis of CogVideoX-5B using DiffTrack. The model consists of 42 transformer layers, and we use 50 sampling timesteps for this study. Our observations reveal that 1) query-key matching achieves higher matching accuracy compared to intermediate feature matching (\cref{fig:analysis_5b}(a)), 2) a small number of layers dominate temporal matching (\cref{fig:analysis_5b}(b)), and 3) temporal matching improves as noise levels decrease, with a slight drop at the final stages of denoising (\cref{fig:analysis_5b}(c)). These findings are consistent with the analysis presented in \cref{sec:method:analysis} for CogVideoX-2B.

\paragrapht{DiffTrack on HunyuanVideo.} \cref{fig:analysis_hy} presents additional analysis of HunyuanVideo using DiffTrack. The model consists of 60 layers, and we use 30 sampling timesteps for this experiment. We observe that: query-key matching outperforms intermediate feature matching (\cref{fig:analysis_hy}(a)), a few specific layers play dominant roles in temporal correspondence (\cref{fig:analysis_hy}(b)), and temporal matching improves as diffusion noise decreases (\cref{fig:analysis_hy}(c)).


\paragrapht{DiffTrack on CogVideoX-2B-I2V.} In \cref{fig:analysis_i2v}, we present further analysis of CogVideoX-2B-I2V~\cite{yang2024cogvideox}, a variant of CogVideoX-2B finetuned on 10 million videos for image-to-video generation.

As discussed in \cref{sec:method:analysis}, query-key matching outperforms intermediate feature matching (\cref{fig:analysis_i2v}(a)), and a small number of layers dominate temporal correspondence (\cref{fig:analysis_i2v}(b)).

\input{figure/i2v_layer_analysis}

Notably, \cref{fig:analysis_i2v}(c) shows that temporal matching declines sharply at later timesteps. This observation aligns with \cref{fig:i2v_layer_analysis}, where self-frame attention scores increase significantly while cross-frame attention scores drop.

As illustrated in \cref{fig:graph_attention}, the original CogVideoX-2B initially relies on text and self-frame attention to construct coarse structures and motion during early denoising steps, and gradually incorporates cross-frame attention in later steps to refine temporal coherence. In contrast, CogVideoX-2B-I2V reduces its reliance on text-frame attention while maintaining strong self-frame and cross-frame attention. In the context of image-to-video generation, this shift emphasizes propagating information from the initial frame rather than adhering closely to the input prompt. This behavioral difference reflects the distinct training objectives of the two models.

\input{table/i2v_davis}
We further evaluate CogVideoX-2B-I2V for point accuracy on the DAVIS dataset. The results summarized in \cref{quan:i2v_davis} indicate a significant performance drop compared to the baseline CogVideoX-2B. This occurs because image-to-video finetuning primarily preserves the first frame, exhibiting limited motion and often losing the temporal matching capability required for generating dynamic motion.

\clearpage
\input{figure/analysis_5b}

\input{figure/analysis_hy}

\input{figure/analysis_cogvideo_i2v}

\clearpage

\section{Zero-Shot Point Tracking Details}
\label{sup:zero-shot-tracking}
\input{figure/figure_zeroshot}
\subsection{Architectural Details} \Cref{fig:figure_zeroshot} illustrates the overall architecture of DiffTrack for zero-shot point tracking on real videos. In this example, a 25-frame video is divided into two chunks for visualization. We further provide a component analysis of this architecture in \cref{supp:abl_zeroshot}.

We find that temporal compression in the 3D VAE limits point accuracy due to the linear interpolation used to reconstruct motion trajectories from latent space. To mitigate this, we set the temporal compression ratio to $q=1$, establishing a one-to-one mapping between each video frame and its latent (\cref{fig:figure_zeroshot}(a)). This is achieved by passing each frame individually to the pre-trained VAE, which compresses $1+4f$ frames to $1+f$ latents, where we set $f=0$.

To enable direct temporal matching with the global first frame, we insert it into every chunk (\cref{fig:figure_zeroshot}(b)). To reduce the temporal gap between the global first frame and other frames within each chunk, we construct chunks using interleaved subsequent frames (\cref{fig:figure_zeroshot}(c)). Frames within each chunk are sampled at uniform intervals, determined by dividing the total video length by $f-1$. Chunks slide with a stride of 1, and the matching costs for overlapping frames across chunks are averaged.

In full 3D attention, each frame latent is projected into $\hat{\mathbf{Q}}^{i}_{t,l}$ and $\hat{\mathbf{K}}^{i}_{t,l}$, where $i$ denotes the frame index. With slight abuse of notation, we represent the projections of the inserted global first-frame latent in the second chunk as $\hat{\mathbf{Q}}^{1'}_{t,l}$ and $\hat{\mathbf{K}}^{1'}_{t,l}$. For each chunk, we compute the matching cost between the first-frame query and the key of the $j$-th frame (\cf \cref{equ:matching_cost}). To enhance performance, following~\cite{an2024cross,hong2021deep,hong2022cost,hong2024unifying,hong2022neural,cho2022cats++,cho2021cats,hong2022integrative}, we employ a bidirectional matching cost that additionally incorporates the transposed inverse cost between the query of the $j$-th frame and the key of the first frame. This results in $\hat{\mathbf{C}}^{1,j}_{t,l}$ for the first chunk and $\hat{\mathbf{C}}^{1',j}_{t,l}$ for the second chunk. Next, we apply an argmax operation (\cf \cref{equ:argmax}) to obtain the matched correspondence points: the starting point ${\mathbf{p}}^{1}_{t,l}$ and its match ${\mathbf{p}}^{j}_{t,l}$ in the $j$-th frame. By concatenating ${\mathbf{p}}^{j}_{t,l}$ across frames, we obtain the full motion trajectory.

\subsection{Implementation Details}
\label{sup:zero-shot-tracking-imp}
\paragraph{Zero-shot Point Tracking.} For zero-shot tracking evaluation~\cite{aydemir2024can}, we used the most significant layer and timestep, identified by matching accuracy in \cref{sec:method:analysis} and \cref{supp:other_backbones}: $l=17$, $t=1$ for CogVideoX-2B, $l=16$, $t=1$ for for CogVideoX-5B, and $l=16$, $t=1$ for HunyuanVideo. All experiments were conducted on an A6000 GPU.

We evaluate zero-shot tracking on two real-video datasets with precisely annotated tracks: TAP-Vid-DAVIS~\cite{doersch2022tap} and TAP-Vid-Kinetics~\cite{doersch2022tap}. DAVIS includes 30 videos with diverse object motions and appearance variations, while Kinetics comprises 1,189 in-the-wild videos featuring rapid scene transitions and motion blur. 

Following prior work~\cite{karaev2024cotracker, karaev2024cotracker3, cho2024local}, we evaluate point accuracy at $256 \times 256$ resolution. For CogVideoX-2B, we resize videos to $256 \times 256$ and then upsample to the training resolution of $480 \times 720$. The resulting feature descriptors have a spatial resolution of $30 \times 45$, as the 3D VAE decompresses spatial size to $1/16$.

\paragrapht{Human Evaluation.} We also provide an example of human evaluation in \cref{fig:user_study}.


\subsection{Comparison}
\label{supp:evaluation}

We demonstrate the effectiveness of DiffTrack in zero-shot point tracking~\cite{aydemir2024can} by comparing it with a diverse set of vision foundation models trained on single images and self-supervised models trained on videos or two-view images. Below, we detail the models included in our comparison. For fair evaluation, following~\cite{aydemir2024can}, we resize inputs to produce feature maps of size $30 \times 45$, except for ZeroCo~\cite{an2024cross}.

\paragrapht{Vision Foundation Models.} We evaluate DiffTrack against vision foundation models, including DINO, DINOv2, DINOv2-Reg, and DIFT (SD1.5, SD2.1).

{DINO}~\cite{caron2021emerging} is a self-supervised vision transformer that learns localized features of salient objects. {DINOv2}~\cite{oquab2023dinov2} improves upon DINO by leveraging a larger dataset and optimized training. {DINOv2-Reg}~\cite{darcet2023vision} introduces register tokens to further reduce attention artifacts. We use ViT-B/16 for DINO and ViT-B/14 for both DINOv2 and DINOv2-Reg. 

{Stable Diffusion 1.5 (SD1.5) and 2.1 (SD2.1)}~\cite{rombach2022high} are U-Net-based text-to-image diffusion models. Following {DIFT}~\cite{tang2023emergent}, we extract features from the third upsampling block to compute point correspondence.

\paragrapht{Self-Supervised Models.} We further evaluate DiffTrack using self-supervised video models, including SMTC, CRW, Spa-then-Temp, VFS, and SVD, as well as a self-supervised model trained on two-view images, ZeroCo. All of these models are trained solely on videos or two-view images without any labels.

{SMTC}~\cite{qian2023semantics} proposes a self-supervised video model that improves semantic and temporal consistency by training the architecture in a teacher-student manner. We use SMTC with ViT-S/16 for our comparison.

{Contrastive Random Walk (CRW)}~\cite{jabri2020space} trains ResNet-18 to learn temporally consistent feature representations through cycle-consistency, maximizing the likelihood of returning to the initial points when walking through palindromic video sequences. 

{Spa-then-Temp}~\cite{li2023spatial} combines spatial and temporal self-supervised learning by first leveraging contrastive learning for spatial features and then enhancing these features through hierarchical frame reconstruction and local correlation distillation. We use Spa-then-Temp based on ResNet-50 for comparison.

{Video Frame-level Similarity (VFS)}~\cite{xu2021rethinking} compares frame-level features from the same video as positive pairs, while frames from different videos serve as negative pairs. We use VFS with ResNet-50 for comparison.

{Stable Video Diffusion (SVD)}~\cite{blattmann2023stable} is a U-Net-based text-to-video generative model extended from SD 2.1 by incorporating additional temporal layers. Following~\cite{jeong2024track4gen}, we use features from the upsampler layer of the third decoder block to calculate point accuracy. We empirically observe that query-key matching achieves higher point accuracy than feature matching; therefore, we adopt query-key matching in the third decoder block of SVD.


{ZeroCo}~\cite{an2024cross} demonstrates query-key matching in the cross-attention map within the self-supervised cross-view completion model CroCo~\cite{an2024cross}, capturing geometric correspondence more effectively than other correlation maps from the encoder or decoder. For comparison, we use an input size of $224 \times 224$ with a feature size of $14 \times 14$, as we empirically found that this trained resolution yields the best point accuracy compared to other input resolutions with higher feature sizes.

\clearpage
\section{Cross-Attention Guidance Details}
\label{supp:CAG}

\paragraph{Architectural Details.} In \cref{fig:figure_CAG}, we present the overall architecture of Cross-Attention Guidance (CAG). Inspired by PAG~\cite{ahn2024self}, which enhances image fidelity by transforming selected self-attention maps in the diffusion U-Net into identity matrices, we extend this idea to the video DiT architecture.

In PAG, the identity matrices are created by multiplying a diagonal mask into the attention map before the softmax operation—where diagonal elements are set to 0 and off-diagonal elements to $-\infty$. After softmax, this yields an identity matrix (diagonal values of 1, others 0), allowing values to pass through unchanged.

A naive extension to video assigns $-\infty$ to cross-frame positions and 0 elsewhere before the softmax operation. However, this undesirably suppresses the scale of self-frame and text-frame attention values. To address this, we instead zero out only the cross-frame attention values after softmax in $\mathbf{A}_{t,l}$, producing modified attention maps $\hat{\mathbf{A}}_{t,l}$ that preserve all other interactions.

\paragrapht{Implementation Details.}
For CAG, we used the top-$3$ dominant layers $l = 13, 17, 21$ for CogVideoX-2B and $l = 15, 17, 18$ for CogVideoX-5B, as identified by harmonic mean in \cref{sec:method:analysis} and \cref{fig:analysis_5b}. Following~\cite{ahn2024self}, we applied the guidance at all sampling timesteps. 

\paragrapht{Evaluation Details.}
We evaluate CAG against its baselines, CogVideoX-2B and CogVideoX-5B, on VBench~\cite{huang2024vbench}. We used the prompt suite provided by VBench for each evaluation dimension.

To assess temporal quality, we report three metrics: {Subject Consistency}, {Background Consistency}, and {Dynamic Degree}. {Subject Consistency} measures whether the appearance of the main subject (\eg a person or an object) remains consistent across frames, computed using DINO feature similarity. {Background Consistency} assesses the temporal coherence of background scenes, measured by CLIP feature similarity across frames. While a completely static video can achieve high scores on the aforementioned temporal quality metrics, it is essential to also evaluate the presence and magnitude of motion. To this end, we calculate {Dynamic Degree}, which quantifies motion dynamics using optical flow computed by RAFT~\cite{teed2020raft}.

To evaluate frame-wise quality, we report {Imaging Quality}, which detects frame-wise distortions (\eg, blur, noise), computed using the MUSIQ~\cite{ke2021musiq} image quality prediction model trained on the SPAQ~\cite{fang2020perceptual} dataset.

\clearpage

\section{Additional Analysis}
\label{supp:analysis}  
In this section, we present additional analysis of CogVideoX-2B~\cite{yang2024cogvideox} using our framework, DiffTrack. To calculate point accuracy (PCK), we use a predefined error threshold $\delta^3$ for all analyses presented in the main paper and appendix.

\input{figure/analysis_davis}

\paragrapht{Analysis with Real Videos.} In \cref{fig:analysis_davis}, we analyze CogVideoX-2B using real videos from the DAVIS dataset~\cite{doersch2022tap}. As discussed in \cref{sec:method:analysis}, reconstructing real videos often introduces inversion errors~\cite{song2020denoising} due to challenges in finding accurate text prompts or discrepancies between real videos and the training distribution.

To address this, instead of applying diffusion inversion, we add Gaussian noise to the frame latents at each timestep $t$. Specifically, we truncate all DAVIS videos to 49 frames, excluding those shorter than 49 frames. At each timestep $t$, we add Gaussian noise corresponding to $t$ to the frame latent, pass it through the diffusion transformer, and extract feature descriptors for analysis. Notably, this approach is consistent with prior work~\cite{tang2023emergent, zhang2023tale}, which identifies two-frame correspondences in U-Net-based image diffusion models.

The analysis with DAVIS in \cref{fig:analysis_davis} shows a similar trend to that observed in \cref{fig:analysis} using a curated synthetic dataset. Specifically, query-key matching consistently outperforms intermediate feature matching. Additionally, a few layers predominantly contribute to temporal matching, and these dominant layers align with those identified in \cref{fig:analysis}. Temporal matching strengthens during the denoising process, with slight drops near the end of the timesteps.

\input{figure/supp_frame_attention}

\paragrapht{Feature Matching vs. Query-Key Matching.}
\cref{fig:frame_attn} compares attention maps across timesteps, derived from intermediate feature matching and query-key matching. We observe that query-key matching successfully tracks physically matched points, whereas intermediate feature matching often fails. This further supports the observation in \cref{fig:analysis}(a) that query-key matching yields better temporal correspondence.

This finding is consistent with prior works~\cite{an2024cross, nam2024dreammatcher}, which demonstrate that query-key similarities include geometric cues crucial for accurate matching, whereas value warps the visual appearance based on these similarities.

\input{figure/PCA}

\paragrapht{PCA Visualization.}
\cref{fig:pca} visualizes the Principal Component Analysis (PCA) of queries, keys, and values across video frames. Queries and keys exhibit stronger structural cues, with similar distributions for nearby pixels, enabling more effective geometric matching. In contrast, values contain high-frequency noisy appearance features, potentially diluting structural cues for correspondence.

This analysis aligns with~\cite{an2024cross,nam2024dreammatcher}, which show that query-key similarities capture structural information crucial for accurate geometric matching, while values encode semantic appearance information, further warped by the query-key similarities.

\paragrapht{Attention Visualization.} \cref{fig:i2i_attn_layers} presents cross-frame attention maps between the first frame and subsequent frames. In the top-ranked layers based on harmonic mean scores (\cref{fig:i2i_attn_layers}(a)), the attention is sharp and accurately localized at the matched point. In contrast, the bottom three layers with the lowest harmonic scores (\cref{fig:i2i_attn_layers}(b)) exhibit diffuse and scattered attention patterns. This supports our analysis in \cref{fig:analysis}(b), which shows that only a few layers contribute significantly to temporal matching.

\cref{fig:i2i_attn_timesteps} illustrates how cross-frame attention evolves throughout the denoising process, revealing that attention becomes progressively sharper at later timesteps. This observation supports our analysis in \cref{fig:analysis}(c), which shows that temporal matching improves as denoising progresses.

\input{figure/sup_abl_query_key}

\cref{fig:t2i_attn_noun} displays the text-to-frame attention maps for the query word “shark” across timesteps. As discussed in \cref{fig:graph_attention}, the attention evolves from coarse to fine as noise levels decrease, highlighting that text primarily determines the global semantic layout during the early stages of denoising.

\paragrapht{Query-Query vs. Key-Key vs. Query-Key Matching.}  
\cref{fig:matching_compare} shows the matching accuracy of query-query, key-key, and query-key interactions on the object dataset. Query-key matching achieves the highest accuracy, indicating that query-key interactions within full 3D attention inherently learn temporal matching.

\clearpage
\section{Ablation Study} 
\subsection{Zero-Shot Point Tracking} 
\label{supp:abl_zeroshot}

\input{table/abl}

\paragraph{Component Analysis.} \cref{quan:comp_analysis} ablates the effects of temporal compression and long-term sequence handling in zero-shot point tracking performance on DAVIS. We evaluate four configurations: (I) the baseline, which employs temporal compression in the 3D VAE and sequential chunking; (II) the baseline without temporal compression ($q=1$), ensuring a direct one-to-one mapping between frames and frame latents; (III) an approach that inserts the first frame into every chunk, enabling explicit interaction with the first frame in all chunks; and (IV) our final method, which further applies interleaved frame construction to better handle long-term context. Compared to (I), (II) improves performance by eliminating interpolation errors introduced by temporal compression. Compared to (II), (III) shows a significant improvement, demonstrating the importance of direct temporal interaction with the global first frame within each chunk. Additionally, (IV) further enhances performance by interleaving frames within each chunk, effectively reducing the frame interval between the first and other frames.

\input{table/average_abl}

\paragrapht{Impact of Feature Selection.} \cref{quan:average_abl} highlights the effectiveness of feature selection by comparing matching accuracy on the object dataset using three strategies: (I) averaging matching costs over all timesteps at the most dominant layer ($l = 17$), (II) averaging over all layers at the most dominant timestep based on matching accuracy ($t = 1$), and (III) using matching costs from the most dominant layer and timestep ($l = 17$, $t = 1$). Selecting optimal features significantly improves accuracy, emphasizing the value of our analysis and the importance of feature selection for reliable temporal matching.

\input{table/fusion_abl}

\paragrapht{Multi-Feature Fusion for Zero-Shot Point Tracking.}  
Prior works~\cite{hedlin2023unsupervised, tang2023emergent, zhang2023tale, meng2025not,nam2024dreammatcher,jin2025appearance} on intermediate diffusion features suggest that fusing multiple timesteps and layers improves semantic correspondence by leveraging the hierarchical structure of diffusion representations. Motivated by this, we investigate whether such fusion also benefits temporal matching.

\input{figure/multi_feature_pca}

\cref{quan:fusion_analysis} summarizes results on the DAVIS dataset~\cite{pont20172017}. In contrast to prior findings in semantic correspondence, we observe that fusing across timesteps and layers has a negligible impact on point accuracy. This discrepancy arises because temporal matching requires precise, pixel-level geometric alignment to track the same physical point across frames, whereas semantic correspondence benefits from multi-scale contextual cues to match similar regions. 

PCA visualizations in \cref{fig:multi_feature_pca} support this finding. In \cref{fig:multi_feature_pca}(a) and (b), query-key features from different timesteps exhibit high similarity, indicating limited benefit from temporal fusion. In contrast, \cref{fig:multi_feature_pca}(c) and (d) show greater variation across features from different layers, suggesting that the tracking performance is primarily driven by a single dominant layer and that layer fusion introduces noise rather than informative diversity.

\paragrapht{Temporal Compression.}  
In \cref{fig:temporal_compression}, we present decoded frames from the 3D VAE decoder and corresponding attention visualizations for different temporal compression ratios $q$. (a) For $q=4$, the reconstructed frame is relatively blurred due to the limited expressivity of the compressed representation, while $q=1$ allows each frame to be reconstructed almost perfectly without compression. (b) The attention visualization for $q=4$ shows more diffuse attention, while $q=1$ exhibits sharper, more focused attention. This supports our architectural design in \cref{fig:figure_zeroshot}(a), which contributes to improved point accuracy.

\input{figure/temporal_compression}

\input{figure/supp_chunk_len}

\paragrapht{Number of Frames per Chunk.}
\cref{fig:chunk_len_table} ablates matching accuracy on DAVIS using different numbers of frames per chunk. We observe that as the number of frames increases, matching accuracy consistently improves. This is further supported by \cref{fig:chunk_len}, which visualizes attention maps across different chunk sizes, showing that longer chunks lead to sharper attention. This suggests that multi-frame interaction enhances temporal correspondence, highlighting the role of cross-frame attention over multiple frames within video DiTs, in contrast to image-based models~\cite{blattmann2023stable, zhang2023tale}.

\subsection{Motion-Enhanced Video Generation} 
\label{supp:abl_guidance}
\input{table/supp_motion_guidance}

\cref{tab:supp_cag} compares the least top-$3$ and most top-$3$ dominant layers for temporal matching ($l=30, 31, 32$ vs. $l=13, 17, 18$) of CogVideoX-5B in motion-enhanced video generation. We observe that guiding with the most dominant layers ($l=13, 17, 18$) consistently improves all evaluation metrics compared to the baseline, whereas guiding with the least dominant layers ($l=30, 31, 32$) results in only marginal gains in Image Quality and even lower performance in Subject Consistency and Background Consistency, under the same guidance scale. These findings further support our analysis that specific layers play a critical role in enhancing temporal consistency during generation.

\clearpage

\section{Additional Qualitative Results}
\label{supp:add_qual}
{Additional qualitative results for zero-shot point tracking and motion-enhanced video generation are provided in \cref{fig:supp_qual} and \cref{fig:supp_qual_guidance_5b}.}

\section{Related Works}
\label{sec:related_work}
\paragraph{Video Diffusion Models.} Early approaches~\cite{guo2023animatediff, singer2022make, li2022efficient, an2023latent, blattmann2023stable, blattmann2023align, chen2023videocrafter1, xing2024dynamicrafter, kim2025moditalker} were primarily based on U-Net~\cite{ho2020denoising, rombach2022high, podell2023sdxl}, often achieved by inflating pre-trained image diffusion models~\cite{ho2020denoising, rombach2022high, podell2023sdxl}, typically with separate spatial and temporal attention mechanisms.
Although efficient, this separation restricts direct frame-to-frame spatial and temporal interactions, leading to temporal inconsistency or a lack of large motions in generated videos.
Sora~\cite{liu2024sora} has demonstrated the effectiveness of Diffusion Transformers (DiTs)~\cite{esser2024scaling} in increasing scalability and improving temporal coherence.
Subsequent works~\cite{zheng2024open, yang2024cogvideox, polyak2025moviegencastmedia, hacohen2024ltx, genmo2023mochi, kling2024, runway2024gen3}, such as CogVideoX~\cite{yang2024cogvideox}, MovieGen~\cite{polyak2025moviegencastmedia}, Mochi1~\cite{genmo2023mochi}, and LTX-Video~\cite{hacohen2024ltx}, have adopted the DiT architecture, achieving unprecedented performance.
Unlike U-Net-based methods, DiTs employ full 3D attention, enabling cross-frame information sharing between frame latents as well as with text embeddings.
This explicit cross-frame attention improves temporal coherence~\cite{yang2024cogvideox}. Despite these advances, how video DiTs capture temporal correspondence during video generation remains unexplored.

\paragrapht{Representation Analysis in Video Diffusion Models.}  
Recent works~\cite{xiao2024video, jeong2024track4gen} explore internal representations in video diffusion models~\cite{guo2023animatediff, guo2024sparsectrl, blattmann2023stable} for controlled video generation and improved motion consistency but do not analyze temporal correspondence and are based on U-Nets~\cite{ho2020denoising, rombach2022high, podell2023sdxl}, which are known to struggle with large motion. Another work~\cite{cai2024ditctrl} explores attention control in video DiT~\cite{yang2024cogvideox} for subject consistency in long video generation but focuses solely on text-to-video attention, overlooking temporal matching between frames.

\paragrapht{Exploring Correspondence in Diffusion Models.} Numerous studies~\cite{hedlin2023unsupervised, tang2023emergent, zhang2023tale, meng2025not, nam2024dreammatcher, jin2025appearance, nam2025visual} have explored intermediate features from pre-trained image diffusion models~\cite{ho2020denoising, rombach2022high, podell2023sdxl} for correspondence~\cite{truong2020glu, truong2021learning, cho2022cats++, nam2023diffusion}.
However, their analyses mainly focus on two-frame correspondence, as image diffusion models are not designed for temporal correspondence in video sequences.

\paragrapht{Temporal Correspondence.}
TAP-Vid~\cite{doersch2022tap} formulates temporal correspondence in video sequences as point tracking, aiming to estimate the motion of physical points across frames. PIPs~\cite{harley2022particle} iteratively refine estimated trajectories within temporal windows, while TAPIR~\cite{doersch2023tapir} incorporates depthwise convolutions and enhances initialization. CoTracker~\cite{karaev2024cotracker} jointly tracks near-dense trajectories using spatial correlations. These methods often rely on training with synthetic datasets~\cite{doersch2022tap}, as annotating real-world data is highly challenging. To address this, a recent study~\cite{aydemir2024can} explored zero-shot tracking with visual foundation models~\cite{radford2021learning, kirillov2023segment, rombach2022high, caron2021emerging, oquab2023dinov2, darcet2023vision}, showing promising results. However, these analyses are limited to single-image models such as Stable Diffusion (SD)~\cite{rombach2022high} and DINOv2~\cite{oquab2023dinov2}, which process a single frame and thus lack temporal awareness.

\clearpage

\input{figure/foreground_prompt}

\input{figure/background_prompt}

\input{figure/additional_dataset_ex}
\input{figure/user_study}
\input{figure/supp_i2i_attn_layers}
\input{figure/supp_i2i_attn_timesteps}
\input{figure/supp_t2i_attn_timesteps_noun}
\input{figure/sup_qual}

\input{figure/sup_qual_guidance_5b}

\clearpage
\section{Broader Impact}
DiffTrack advances the understanding of temporal correspondence in video diffusion transformers (DiTs), enabling applications such as zero-shot tracking and motion-enhanced video generation. These capabilities can benefit diverse downstream tasks, including point tracking~\cite{karaev2024cotracker, karaev2024cotracker3, cho2024local}, 4D point tracking~\cite{cho2025seurat}, and motion-manipulated video generation~\cite{geng2024motion}.

However, DiffTrack's ability to enhance video quality may raise ethical concerns if misused to create misleading or fake content. It is essential to ensure that advancements from DiffTrack are applied responsibly, especially in video synthesis and manipulation.

\section{Limitations}
DiffTrack relies on pre-trained video diffusion transformers (DiTs), meaning that advancements in video backbones could enhance its performance, resulting in more accurate tracking and coherent motion generation.

While DiffTrack effectively analyzes temporal correspondences and improves motion consistency, it does not currently support motion manipulation—direct control of video synthesis with user-defined motion trajectories. Extending DiffTrack for motion-conditioned video generation is a potential future direction.

Furthermore, DiffTrack performs zero-shot point tracking without fine-tuning on specific datasets. Incorporating fine-tuning could further improve real-world performance, which we plan to explore in future research.

\clearpage

%% file: figure/i2v_layer_analysis.tex
\begin{wrapfigure}{r}{0.45\textwidth} 
    \centering
    \vspace{-12pt}
    \includegraphics[width=1\linewidth]{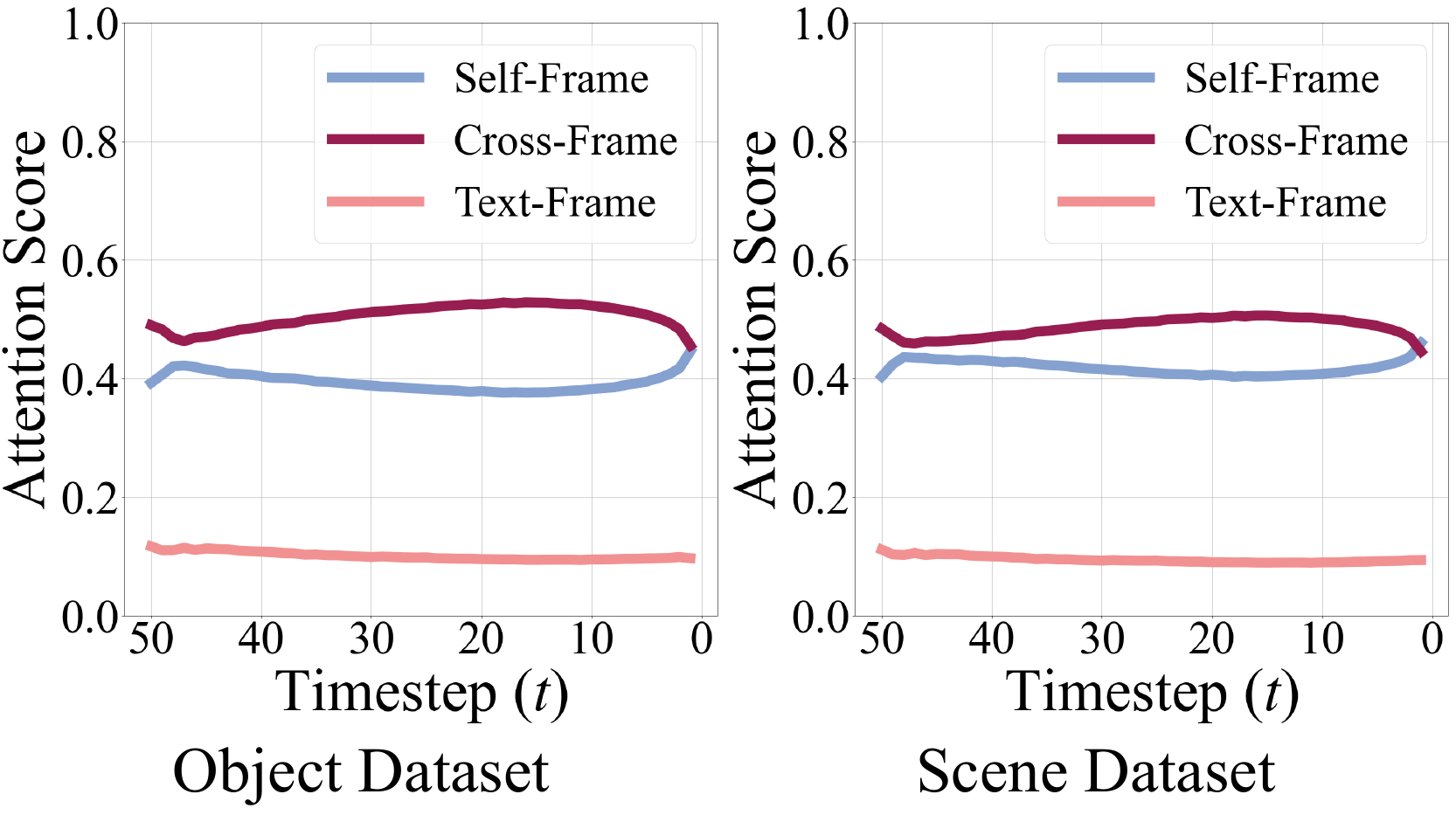} 
    \captionsetup{width=1\linewidth}
   \caption{\textbf{Evolution of attention scores across timesteps.} At later timesteps, cross-frame attention suddenly drops, while self-frame attention sharply increases.}
    \label{fig:i2v_layer_analysis}
    \vspace{-15pt}
\end{wrapfigure}

%% file: table/i2v_davis.tex
 \begin{wraptable}{r}{0.5\textwidth}  
\centering
\vspace{-10pt}
\resizebox{1\linewidth}{!}{ 
    \begin{tabular}{l|cccc}
        \toprule
         Backbone & $<\delta^{0}$ & $<\delta^{2}$ & $<\delta^{4}$ & $<\delta^{x}_\text{avg}$  \\
        \midrule
        DiffTrack (HunyuanVideo~\cite{kong2024hunyuanvideo}) & \underline{4.4} & \underline{44.8} & \underline{82.8} & \underline{44.1} \\
         DiffTrack (CogVideoX-2B~\cite{yang2024cogvideox}) & \textbf{4.8} & \textbf{49.2} & \textbf{84.3} & \textbf{46.3} \\
          DiffTrack (CogVideoX-2B-I2V~\cite{yang2024cogvideox}) & 0.6 & 11.5 & 41.2 & 16.4 \\
        \bottomrule
    \end{tabular}
}
\caption{\textbf{Quantitative comparison on the DAVIS dataset~\cite{doersch2022tap}.}}
\vspace{-10pt}
\label{quan:i2v_davis}
\end{wraptable}

%% file: figure/analysis_5b.tex
\begin{figure*}[t!]
    \centering
\includegraphics[width=\linewidth]{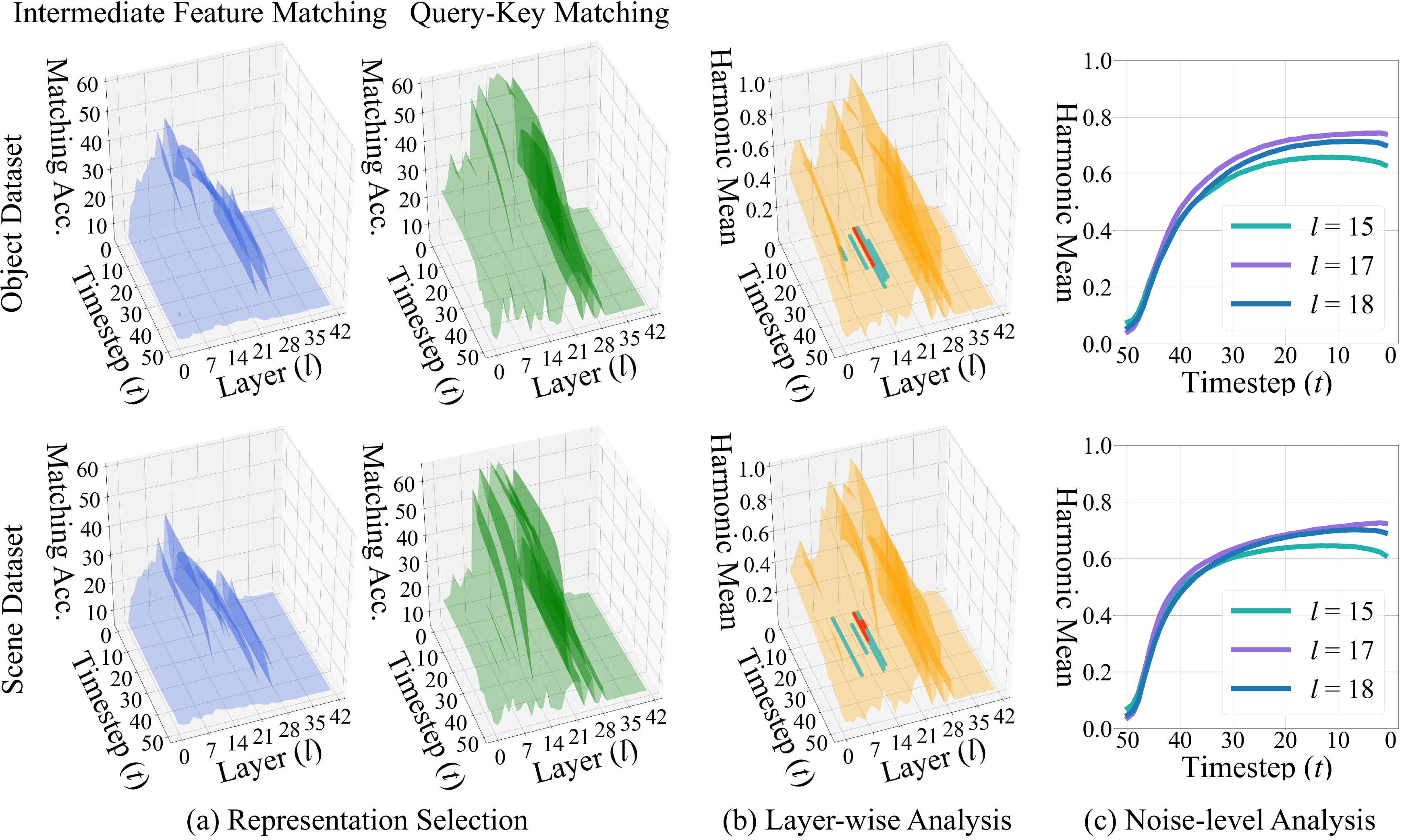}
    \vspace{-15pt}
    \caption{\textbf{Analysis of temporal matching in CogVideoX-5B~\cite{yang2024cogvideox}.} (a) \textit{Representation selection}: Query-key matching achieves higher accuracy than intermediate feature matching. (b) \textit{Layer-wise analysis}: Temporal correspondence is primarily governed by a limited set of layers. (c) \textit{Noise-level analysis}: Temporal matching improves as noise decreases but slightly degrades near the final steps.}
    \label{fig:analysis_5b}
    \figureaftercaption
\end{figure*}

%% file: figure/analysis_hy.tex
\begin{figure*}[t!]
    \centering
\includegraphics[width=\linewidth]{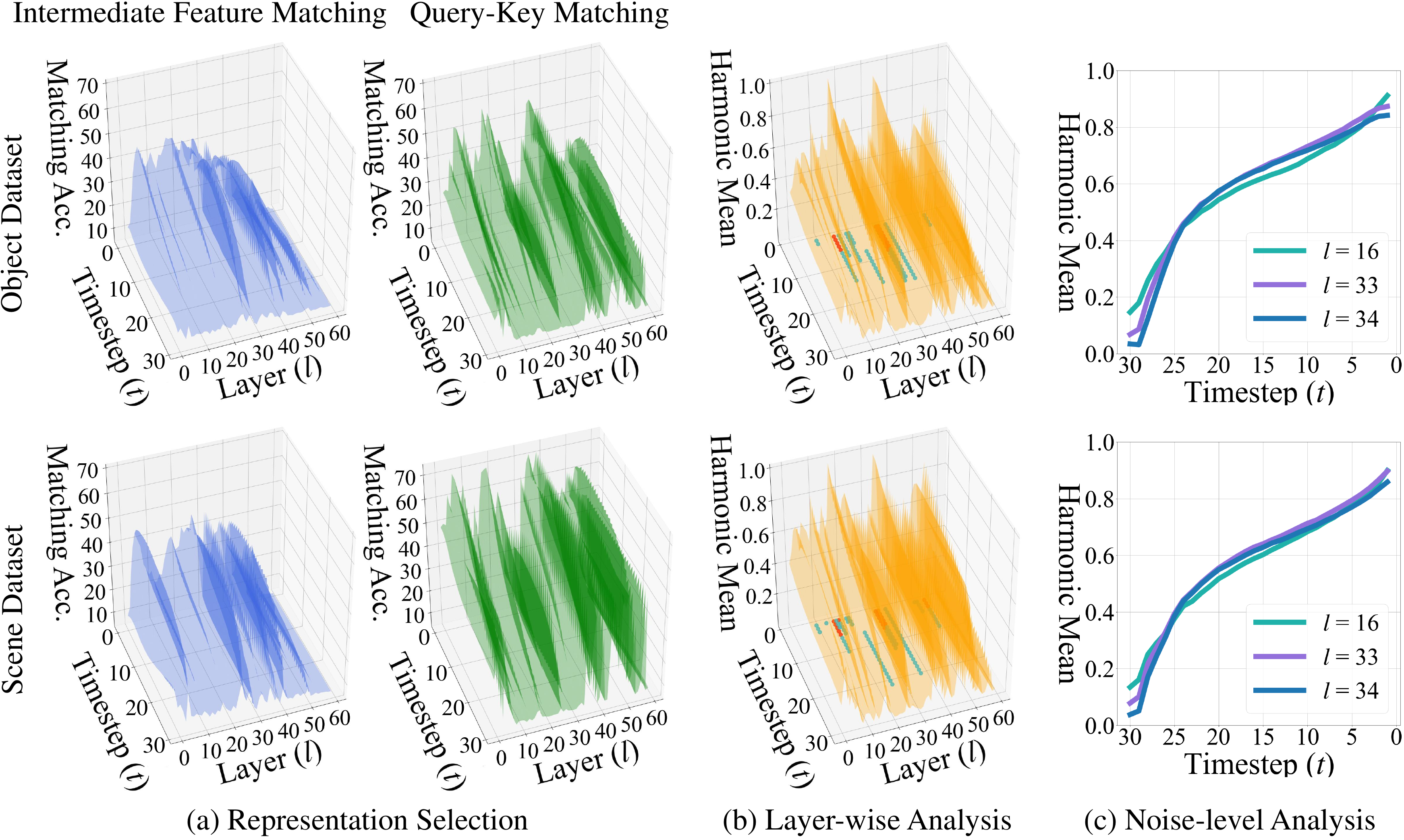}
    \vspace{-15pt}
    \caption{\textbf{Analysis of temporal matching in HunyuanVideo~\cite{kong2024hunyuanvideo}.} (a) \textit{Representation selection}: Query-key matching achieves higher accuracy than intermediate feature matching. (b) \textit{Layer-wise analysis}: Temporal correspondence is primarily governed by a limited set of layers. (c) \textit{Noise-level analysis}: Temporal matching improves as noise decreases.}
    \label{fig:analysis_hy}
    \figureaftercaption
\end{figure*}

%% file: figure/analysis_cogvideo_i2v.tex
\begin{figure*}[t!]
    \centering
\includegraphics[width=\linewidth]{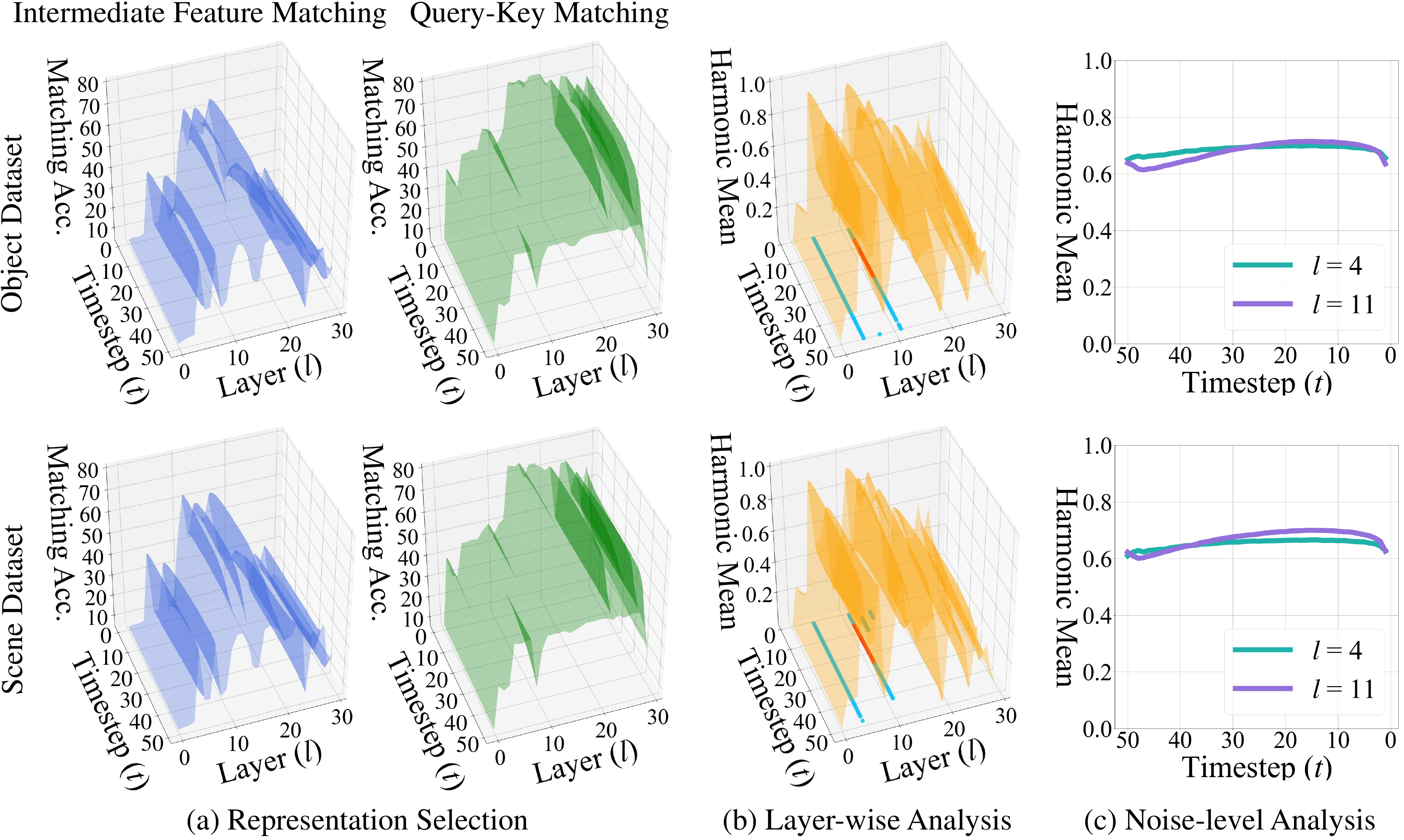}
    \caption{\textbf{Analysis of temporal matching in CogVideoX-2B-I2V~\cite{yang2024cogvideox}.} (a) \textit{Representation selection}: Query-key matching achieves higher accuracy than intermediate feature matching. (b) \textit{Layer-wise analysis}: Temporal correspondence is primarily governed by a limited set of layers. (c) \textit{Noise-level analysis}: Temporal matching improves during early denoising steps but sharply declines at later timesteps, coinciding with a notable decrease in cross-frame attention and a corresponding increase in self-frame attention.}
    \label{fig:analysis_i2v}
    \figureaftercaption
\end{figure*}

%% file: figure/figure_zeroshot.tex
\begin{figure}[h]
    \centering
    \includegraphics[width=0.8\linewidth]{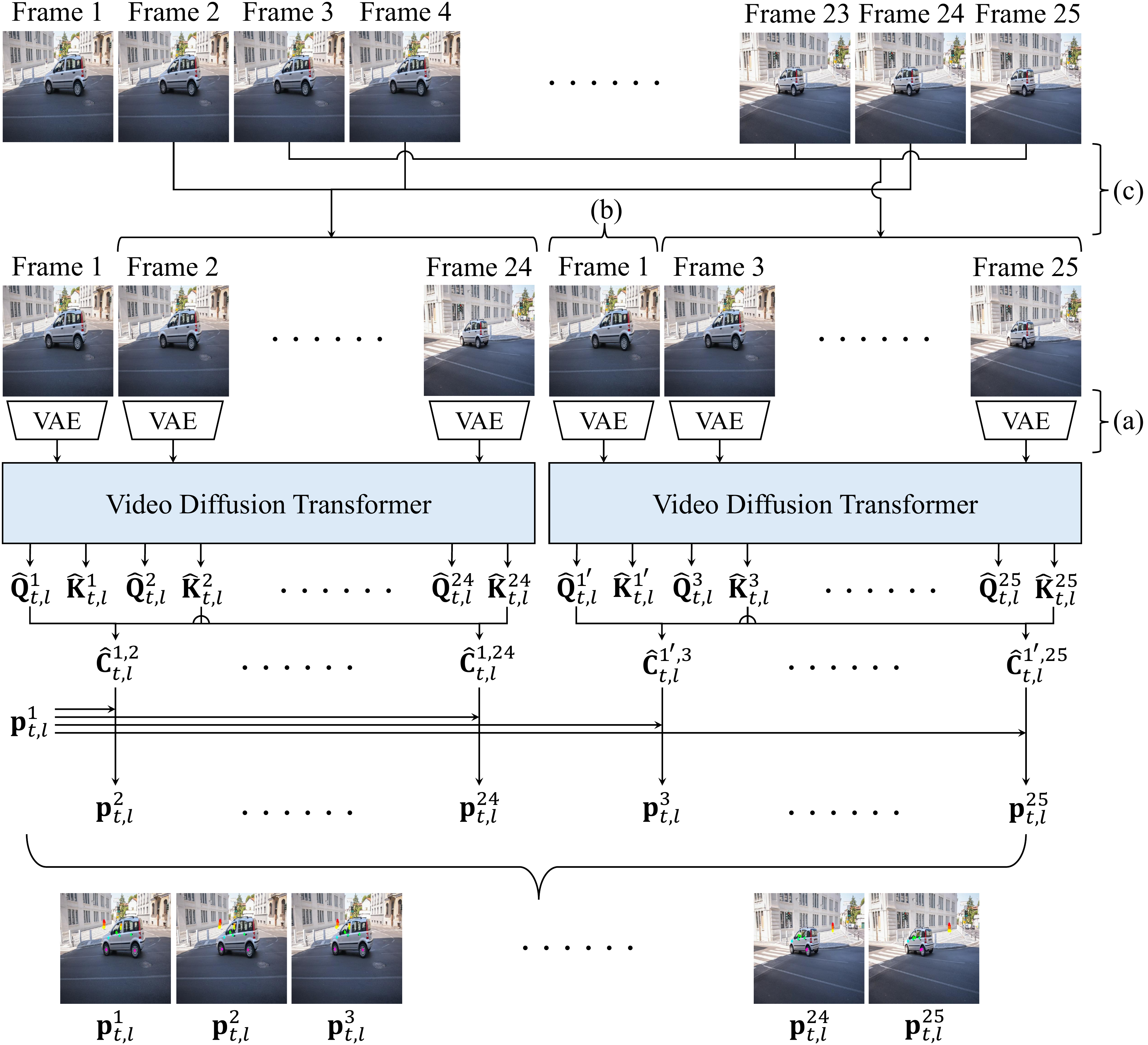} 
    \caption{\textbf{Overall architecture of DiffTrack for zero-shot point tracking.} 
(a) Removing temporal compression improves point accuracy by avoiding interpolation. 
(b) A global first-frame is inserted into each chunk for direct matching. 
(c) Interleaved chunk construction reduces the temporal gap to the first frame.}
    \label{fig:figure_zeroshot}
    \figureaftercaption
\end{figure}

%% file: figure/analysis_davis.tex
\begin{figure*}[h]
    \centering
\includegraphics[width=\linewidth]{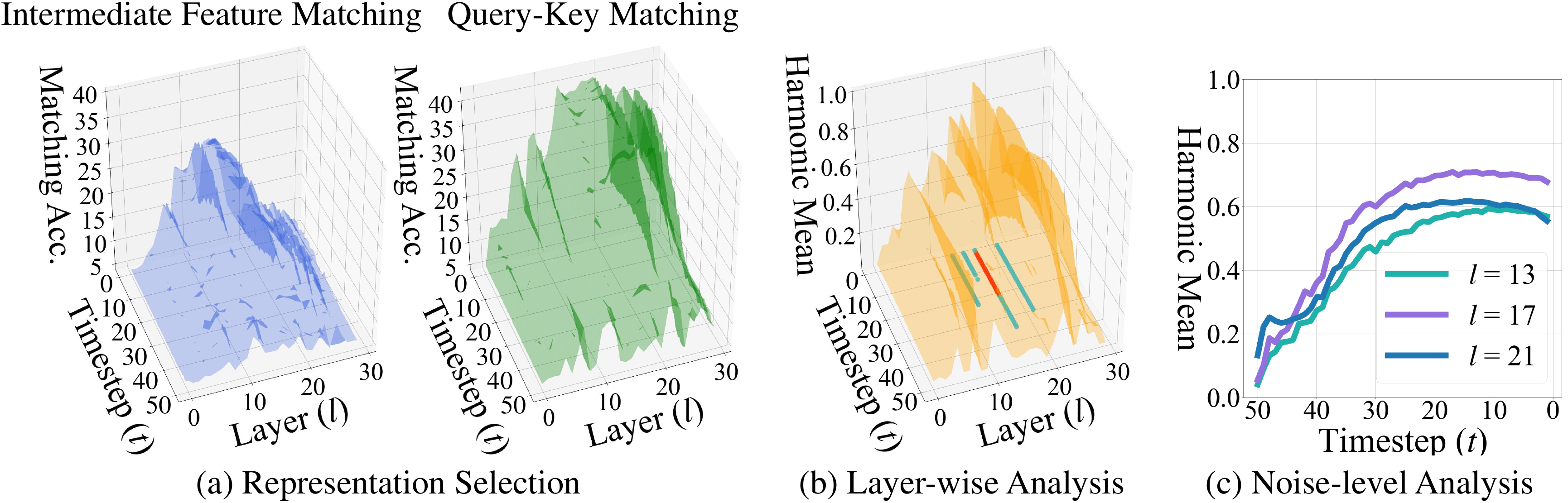}
    \vspace{-15pt}
    \caption{\textbf{Analysis of temporal matching in CogVideoX-2B~\cite{yang2024cogvideox} with DAVIS~\cite{doersch2022tap} dataset.} (a) \textit{Representation selection}: Query-key matching achieves higher accuracy than intermediate feature matching. (b) \textit{Layer-wise analysis}: Temporal correspondence is primarily governed by a limited set of layers. (c) \textit{Noise-level analysis}: Temporal matching improves throughout the denoising process and slightly degrades near the end.}
    \label{fig:analysis_davis}
    \figureaftercaption
\end{figure*}

%% file: figure/supp_frame_attention.tex
\begin{figure*}[h]
    \centering
\includegraphics[width=\linewidth]{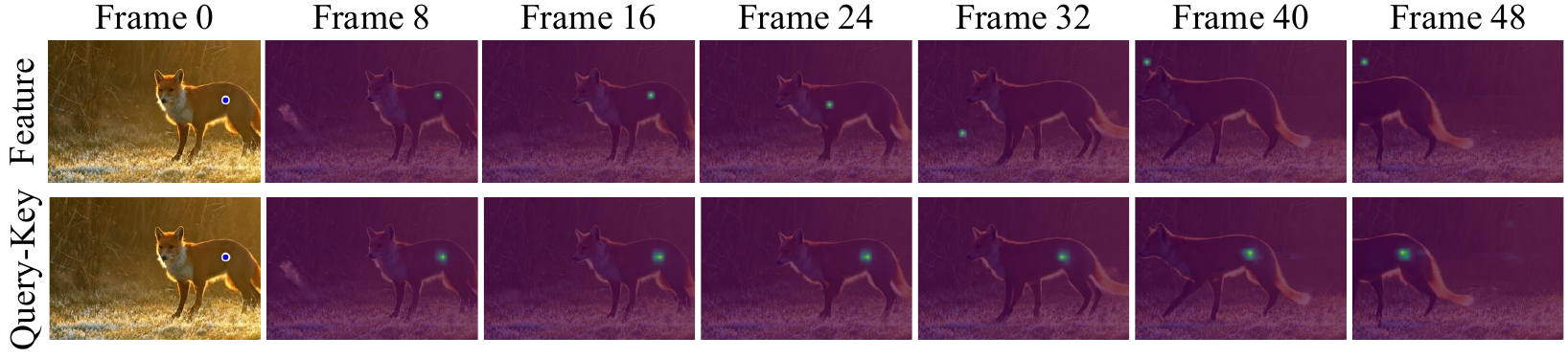}
    \vspace{-15pt}
  \caption{\textbf{Attention visualization comparison between intermediate feature matching and query-key matching.}}
    \label{fig:frame_attn}
    \figureaftercaption
\end{figure*}

%% file: figure/PCA.tex
\begin{wrapfigure}{r}{0.39\textwidth} 
    \centering
    \vspace{-5pt}
    \includegraphics[width=1\linewidth]
    {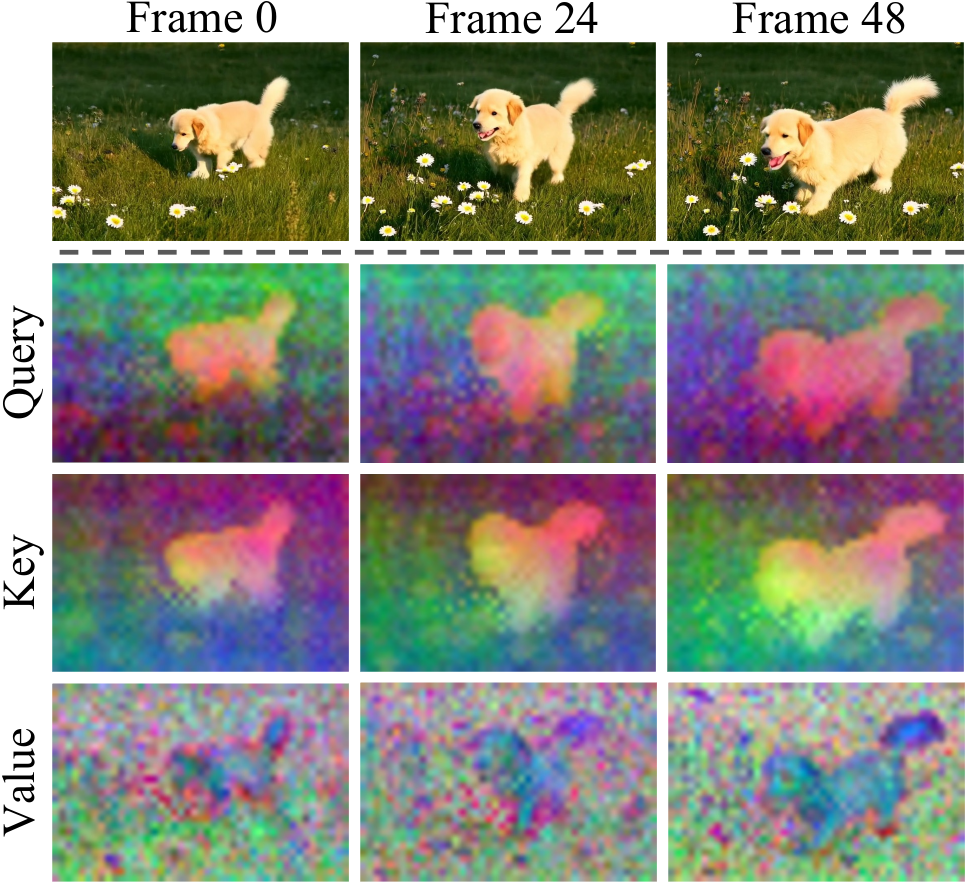} 
    \captionsetup{width=1\linewidth}
  \caption{\textbf{PCA visualization of queries, keys and values.} }
  \vspace{-15pt}
    \label{fig:pca}
\end{wrapfigure}

%% file: figure/sup_abl_query_key.tex
\begin{wrapfigure}{r}{0.5\textwidth} 
    \centering
    \vspace{-12pt}
    \includegraphics[width=1\linewidth]{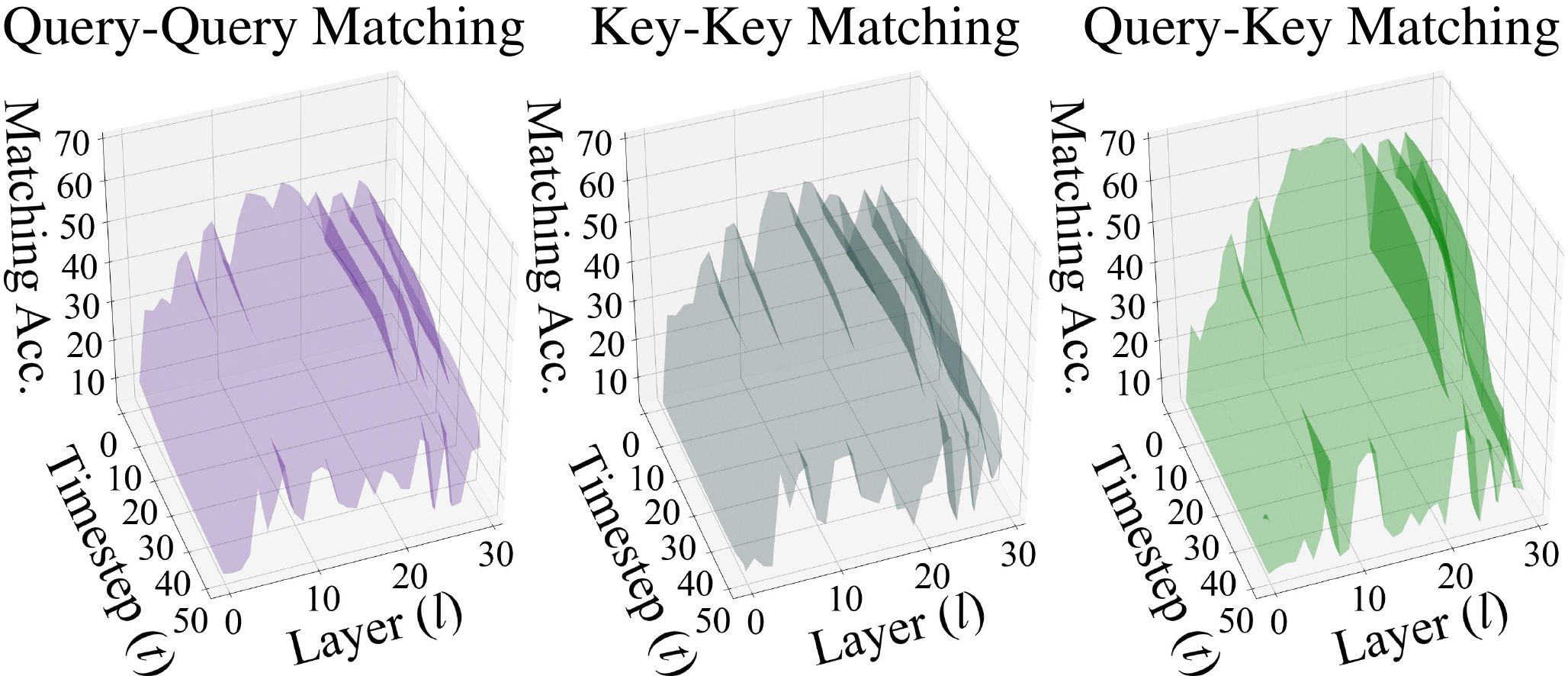}
    \captionsetup{width=1\linewidth}
  \caption{\textbf{Comparison of query-query, key-key, and query-key matching.}}
  \vspace{-15pt}
    \label{fig:matching_compare}
\end{wrapfigure}

%% file: table/abl.tex
 \begin{wraptable}{r}{0.6\textwidth}  
\centering
\vspace{-10pt}  
\resizebox{1\linewidth}{!}{  
    \begin{tabular}{c|l|cccc}
        \toprule
        & Component & $<\delta^{0}$   & $<\delta^{2}$   & $<\delta^{4}$   & $<\delta^{x}_\text{avg}$  \\
        \midrule
        (I) & Baseline & 2.6 & 21.9 & 60.2 & 26.9 \\
        (II) & (I) + w/o temporal compression & 2.7 & 31.6 & 64.5 & 32.3 \\
        (III) & (II) + w/ first-frame insertion & 4.7 & 47.6 & 70.5 & 44.6 \\
        (IV) & (III) + w/ interleaved frames & \textbf{4.8} & \textbf{49.2} & \textbf{84.3} & \textbf{46.3} \\
        \bottomrule
    \end{tabular}
}
\caption{\textbf{Ablation study:} analyzing the impact of temporal compression and long-term handling.}
\vspace{-10pt}
\label{quan:comp_analysis}
\end{wraptable}

%% file: table/average_abl.tex

\begin{wraptable}{r}{0.4\textwidth}  
\centering
\vspace{-10pt}  
\resizebox{1\linewidth}{!}{  
    \begin{tabular}{l|ll|c}
        \toprule
        & Layer ($l$) & Timestep ($t$) & $<\delta^{3}$ \\
        \midrule
        (I) & $l=17$ & $t\in[1,50]$  & 15.34 \\
        (II) & $l\in[0,29]$ & $t=1$ & 16.69 \\
        \midrule
        (III) & $l=17$ & $t=1$  & \textbf{63.50} \\
        \bottomrule
    \end{tabular}
}
\caption{\textbf{Ablation study,} analyzing the impact of feature selection.} 
\vspace{-15pt}
\label{quan:average_abl}
\end{wraptable}

%% file: table/fusion_abl.tex
 \begin{wraptable}{r}{0.5\textwidth}  %
\centering
\vspace{-10pt}  
\resizebox{1\linewidth}{!}{  
    \begin{tabular}{ll|cccc}
        \toprule
         Layer ($l$) & Timestep ($t$) & $<\delta^{0}$ & $<\delta^{2}$ & $<\delta^{4}$ & $<\delta^{x}_\text{avg}$  \\
        \midrule
        $l=17$ & $t=1$ & 4.8 & 49.2 & 84.3 & 46.3 \\
        $l=13, 17, 18$ & $t=1$ & 4.9 & 48.6 & 84.2 & 46.0 \\
        $l=17$ & $t=1,2,3$ & 4.9 & 49.3 & 85.0 & 46.6 \\
        \bottomrule
    \end{tabular}
}
\caption{\textbf{Ablation study,} analyzing fusing features across multiple timesteps and layers.}
\vspace{-15pt}
\label{quan:fusion_analysis}
\end{wraptable}

%% file: figure/multi_feature_pca.tex
\begin{wrapfigure}{r}{0.47\textwidth} 
    \centering
    \vspace{-15pt}
    \includegraphics[width=1\linewidth]
   {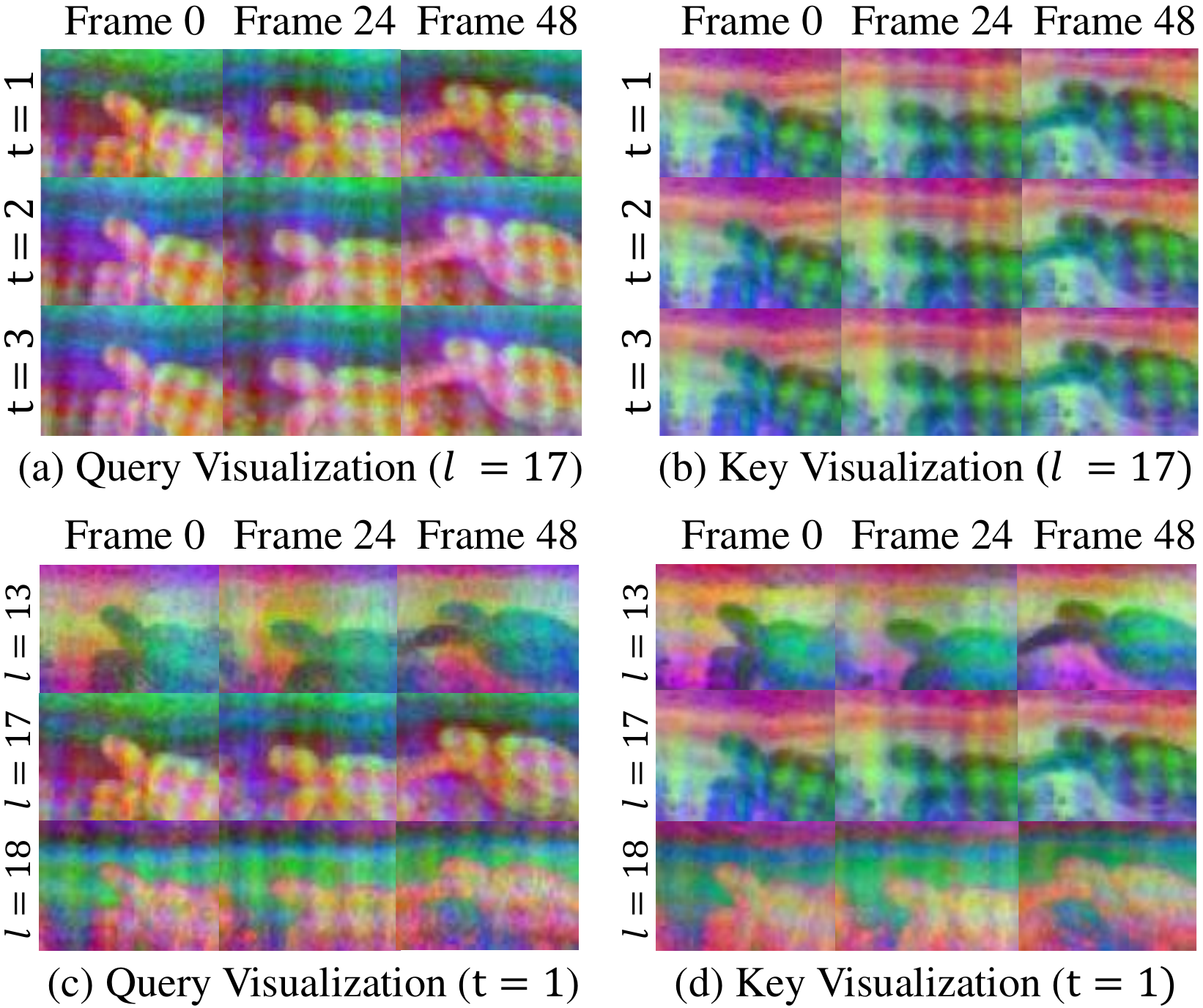}
    \captionsetup{width=1\linewidth}
  \caption{\textbf{PCA visualizations of query-key features across timesteps and layers.} }
  \vspace{-15pt}
    \label{fig:multi_feature_pca}
\end{wrapfigure}

%% file: figure/temporal_compression.tex
\begin{figure*}[t!]
    \centering
\includegraphics[width=1\linewidth]{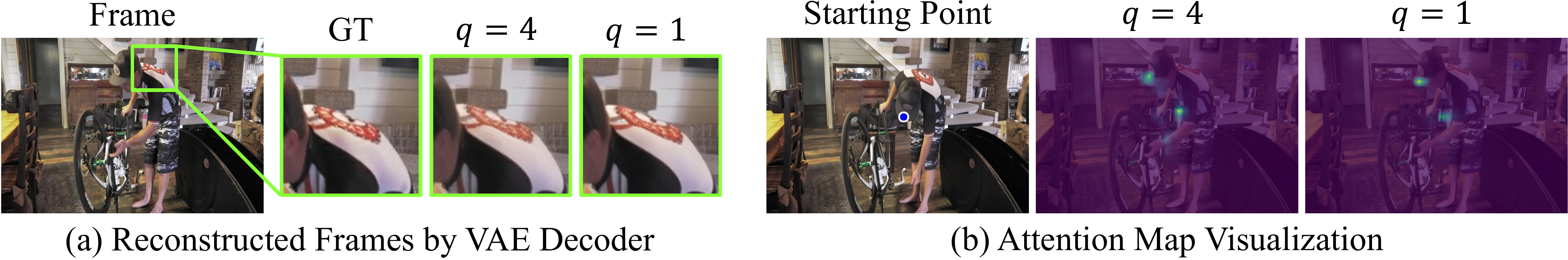}
    \vspace{-15pt}
    \caption{\textbf{Analysis of temporal compression.} (a) Reconstructed frames from the VAE decoder using different temporal compression ratios $q$. (b) Visualization of attention maps under different temporal compression ratios.}
    \label{fig:temporal_compression}
    \figureaftercaption
\end{figure*}

%% file: figure/supp_chunk_len.tex
\begin{figure}[htbp]
    \begin{minipage}[t]{0.35\textwidth}
        \centering
        \vspace{-59pt}
        \resizebox{\linewidth}{!}{
            \begin{tabular}{c|cccc}
                \toprule
                \#Frames & $<\delta^{0}$ & $<\delta^{2}$ & $<\delta^{4}$ & $<\delta^{x}_\text{avg}$ \\
                \midrule
                2 &  4.4 & 43.9 & 77.1 & 41.7 \\
                5 & 4.5 & 46.3 & 82.2 & 44.3 \\
                8 & 4.7 & 47.9 & 83.8 & 45.5 \\
                11 & 4.8 & 48.5 & 84.3 & 46.1 \\
                13 & \textbf{4.8} & \textbf{49.2} & \textbf{84.3} & \textbf{46.3} \\
                \bottomrule
            \end{tabular}
        }
        \caption{\textbf{Ablation study} on the number of frames per chunk.}
        \label{fig:chunk_len_table}
    \end{minipage}
    \begin{minipage}[t]{0.645\textwidth}
        \includegraphics[width=\linewidth]{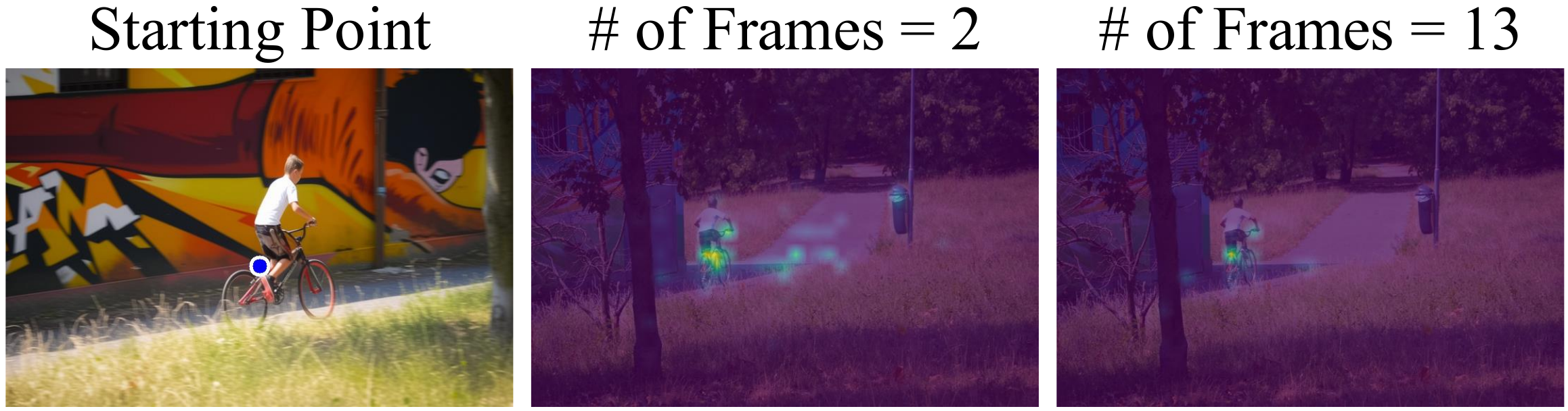}
        \caption{\textbf{Attention map for different number of frames.}}
        \label{fig:chunk_len}
    \end{minipage}
\hfill
\end{figure}

%% file: table/supp_motion_guidance.tex
\begin{wraptable}{r}{0.55\textwidth}
\centering
\vspace{-10pt}
\resizebox{1\linewidth}{!}{
\begin{tabular}{l|cccc}  
    \toprule
    \textbf{Method} & \shortstack{Subject\\Consistency} & \shortstack{Background\\Consistency} & \shortstack{Dynamic\\Degree} & \shortstack{Imaging\\Quality} \\
    \midrule
    CogVideoX-5B & 0.9158&0.9590&0.6667&0.5531 \\
    CogVideoX-5B + \textbf{CAG $(l=30,31,32)$} & 0.9147 & 0.9580 & \textbf{0.7059} & 0.5683\\
     CogVideoX-5B + \textbf{CAG $(l=13,17,18)$} &\textbf{0.9283}&\textbf{0.9644}&{0.6863}& \textbf{0.6051} \\
    \bottomrule
\end{tabular}
}
\vspace{-5pt}
\caption{{\textbf{Quantitative comparison of CAG applied at different layers.}}}
\vspace{-10pt}
\label{tab:supp_cag}
\end{wraptable}

%% file: figure/foreground_prompt.tex
\begin{figure*}
    \centering
\includegraphics[width=1.0\linewidth]{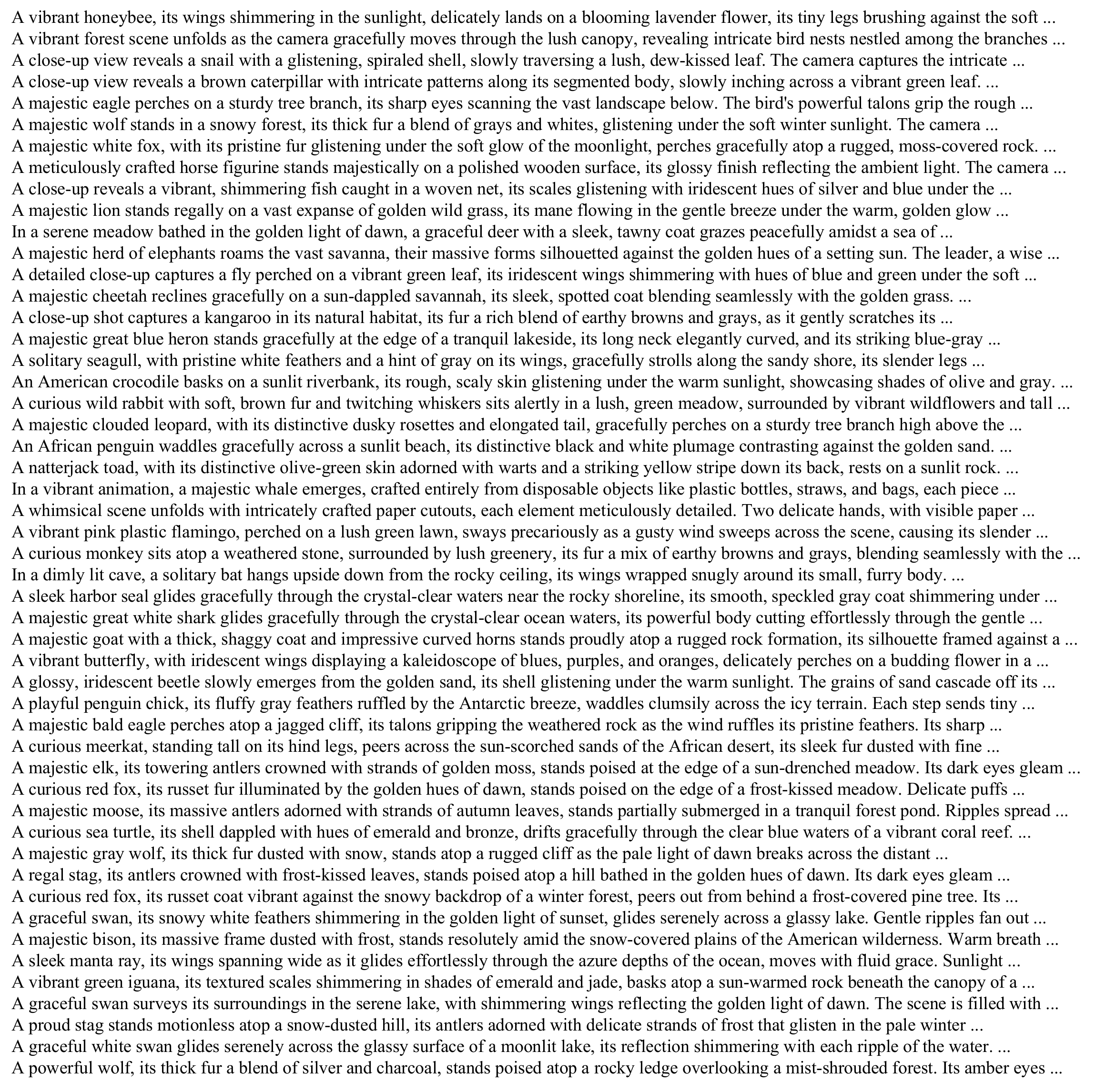}
    \caption{\textbf{Evaluation prompts for the object dataset.} Our high-quality text prompts are curated from existing benchmarks and generated by a large language model, followed by human annotation to ensure motion consistency and video fidelity in the generated videos.}
    \label{fig:fg_prompt}
    \figureaftercaption
\end{figure*}

%% file: figure/background_prompt.tex
\begin{figure*}
    \centering
\includegraphics[width=1.0\linewidth]{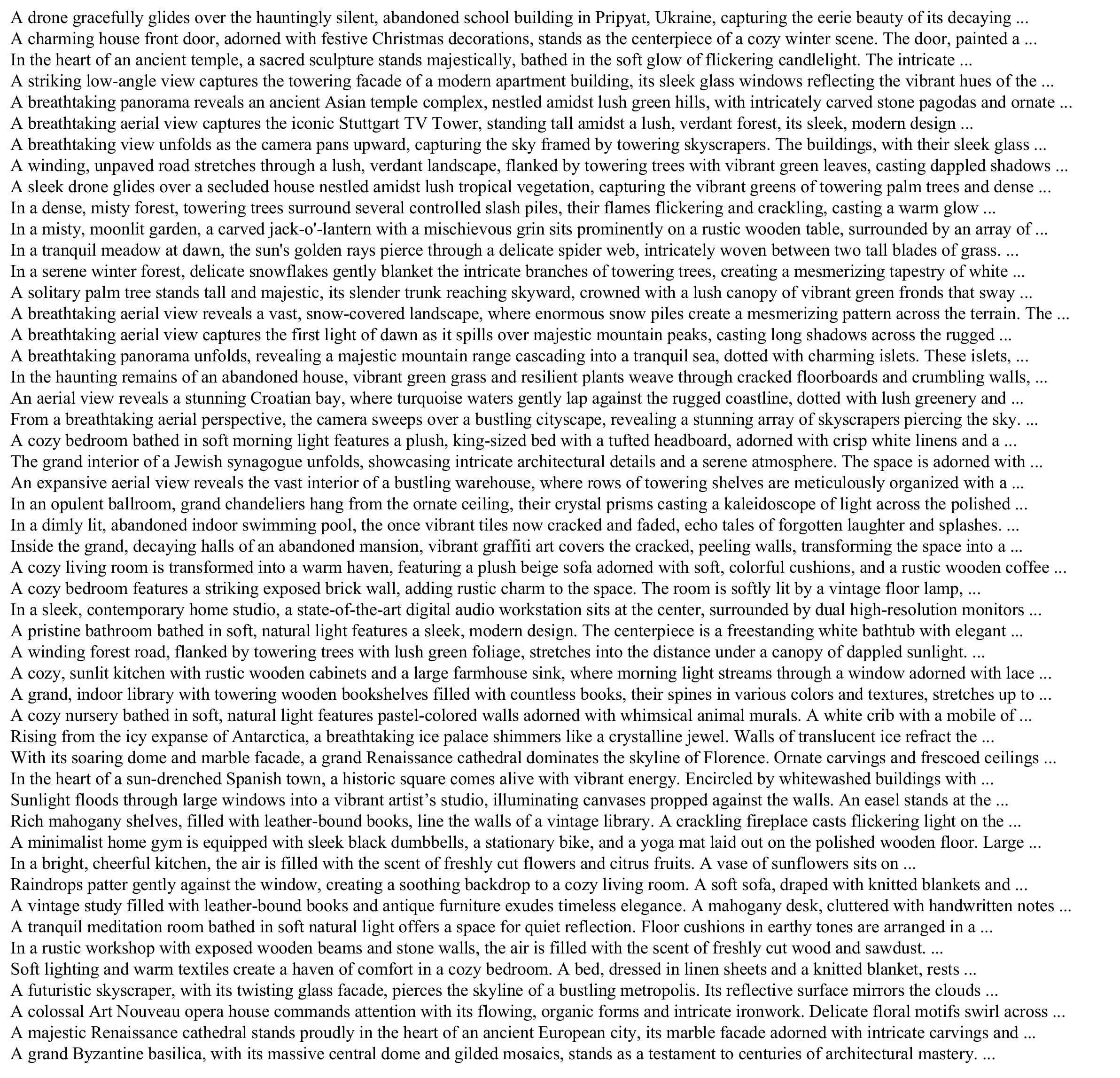}
    \caption{\textbf{Evaluation prompts for the scene dataset.} Our high-quality text prompts are curated from existing benchmarks and generated by a large language model, followed by human annotation to ensure motion consistency and video fidelity in the generated videos.}
    \label{fig:bg_prompt}
    \figureaftercaption
\end{figure*}

%% file: figure/additional_dataset_ex.tex
\begin{figure*}
    \centering
\includegraphics[width=1.0\linewidth]{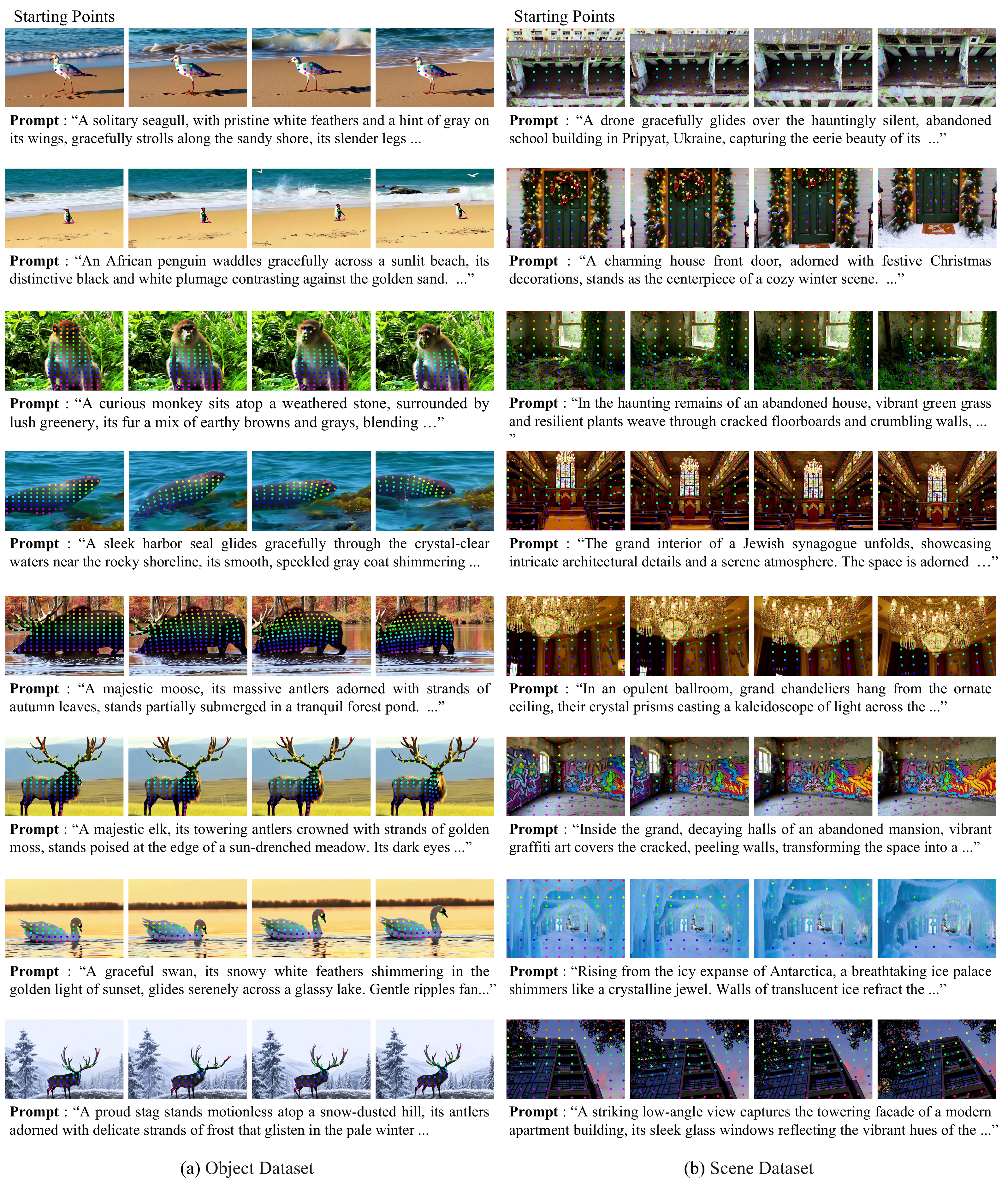}
    \caption{\textbf{Additional examples of our curated dataset.} (a) An object dataset for dynamic object-centric videos and (b) a scene dataset for static scenes with camera motion. The dataset includes predefined starting points in the first frame and their pseudo ground-truth trajectories, obtained using an off-the-shelf tracking method.}
    \label{fig:add_dataset_ex}
    \figureaftercaption
\end{figure*}

%% file: figure/user_study.tex
\begin{figure*}
    \centering
\includegraphics[width=1.0\linewidth]{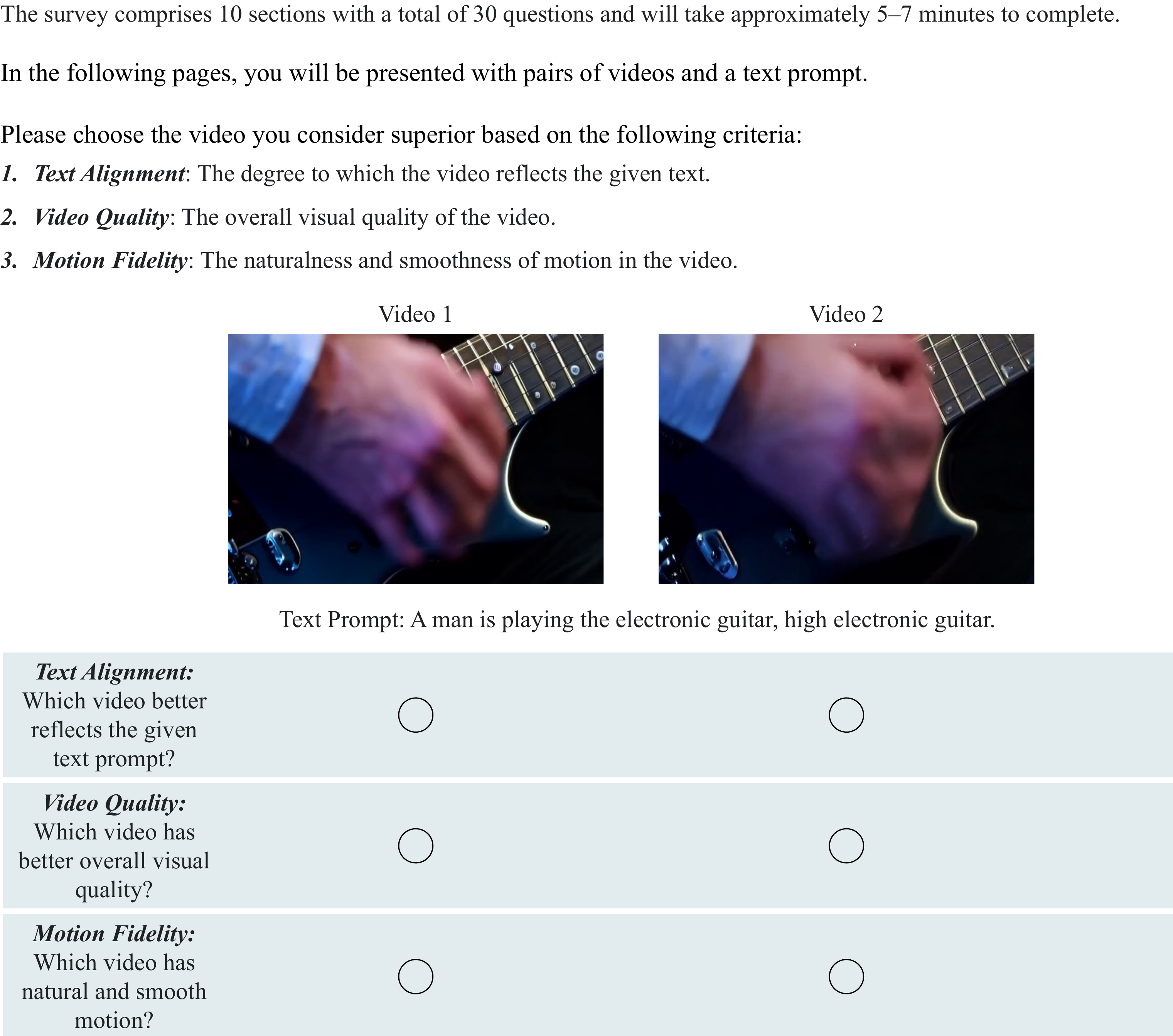}
    \caption{\textbf{An example of human evaluation.} Participants are presented with a pair of videos and a text prompt and are instructed to evaluate text alignment, video quality, and motion fidelity.}
    \label{fig:user_study}
    \figureaftercaption
\end{figure*}
\clearpage

%% file: figure/supp_i2i_attn_layers.tex
\begin{figure*}[t!]
    \centering
\includegraphics[width=\linewidth]{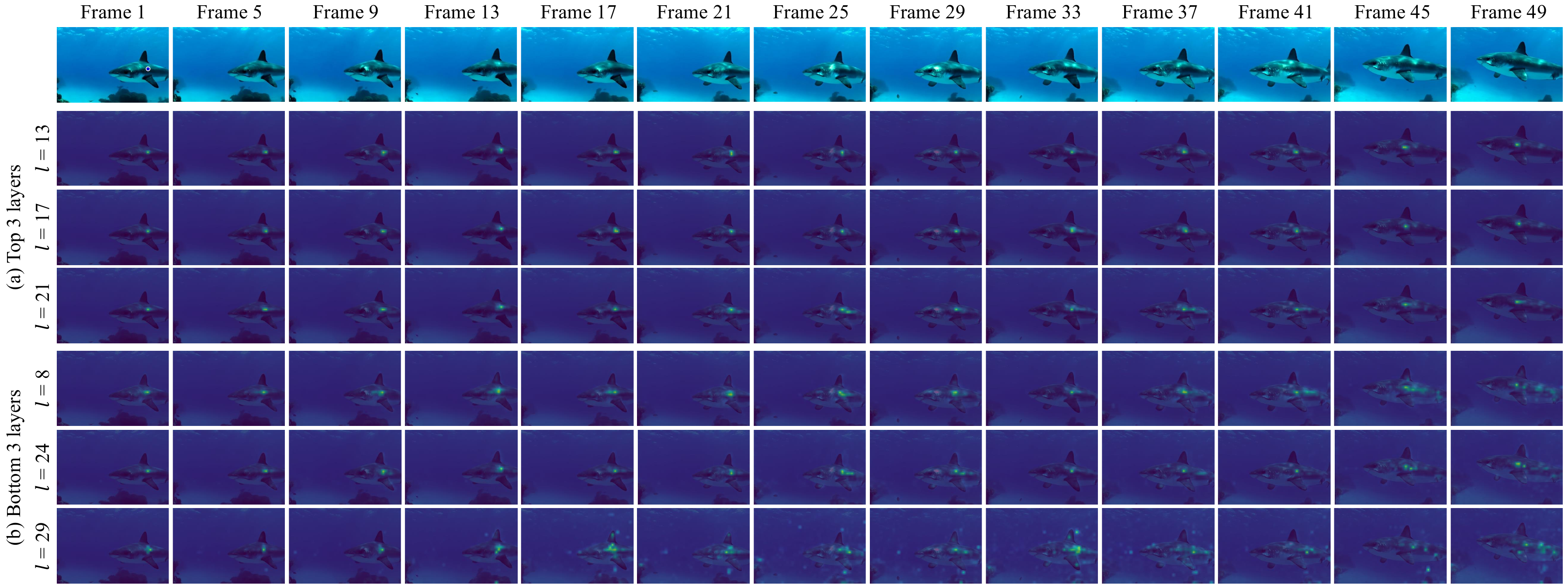}
    \vspace{-15pt}
    \caption{\textbf{Cross-frame attention visualization across layers at timestep $t=1$.} (a) Top-$3$ layers ($l=13,17,21$) exhibit sharp and precise localization. (b) Bottom-$3$ layers ($l=8,24,29$) display diffuse and scattered attention.}
    \label{fig:i2i_attn_layers}
    \figureaftercaption
\end{figure*}

%% file: figure/supp_i2i_attn_timesteps.tex
\begin{figure*}[t!]
    \centering
\includegraphics[width=\linewidth]{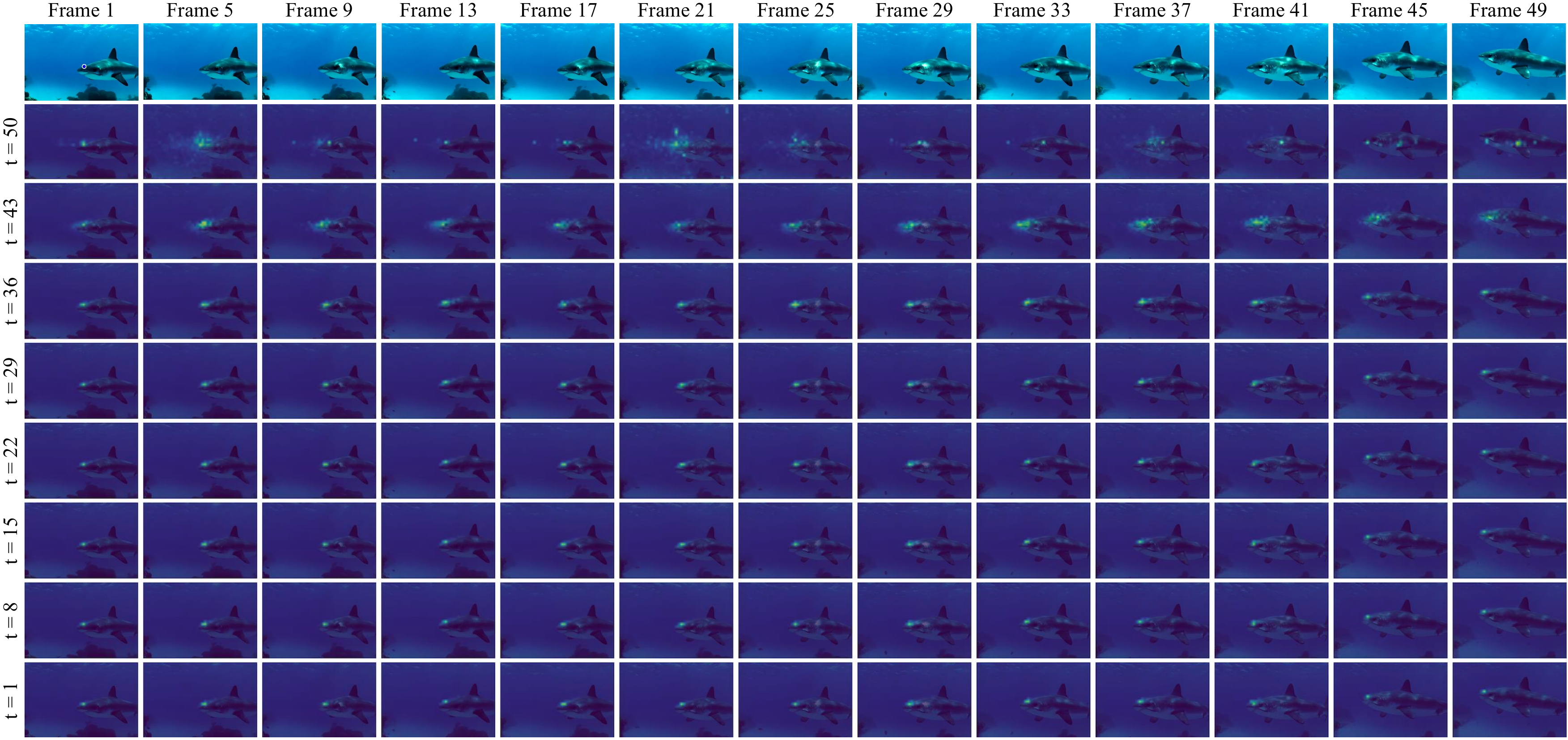}
    \vspace{-15pt}
    \caption{\textbf{Cross-frame attention visualization across denoising timesteps at layer $l=17$.} Attention progressively sharpens and localizes throughout denoising ($t=50$ to $1$).}
    \label{fig:i2i_attn_timesteps}
    \figureaftercaption
\end{figure*}

%% file: figure/supp_t2i_attn_timesteps_noun.tex
\begin{figure*}[!htbp]
    \centering
\includegraphics[width=\linewidth]{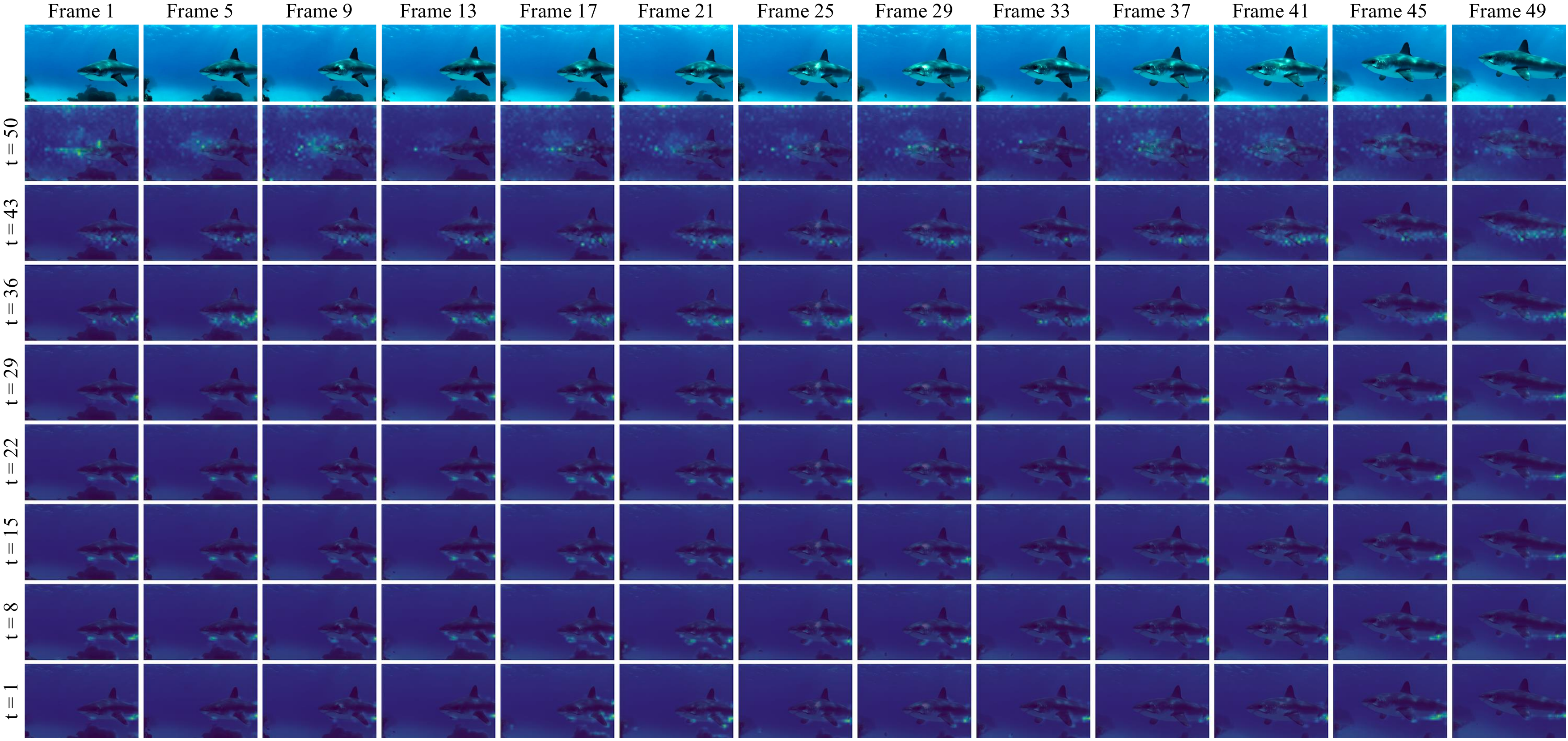}
    \vspace{-15pt}
    \caption{\textbf{Text-to-frame attention visualization for the word “shark” across denoising timesteps at layer $l=17$.} Attention evolves from coarse to fine throughout denoising ($t=50$ to $1$), indicating that the text prompt primarily guides the global semantic layout in early timesteps.}
    \label{fig:t2i_attn_noun}
    \figureaftercaption
\end{figure*}

%% file: figure/sup_qual.tex
\begin{figure*}
    \centering
\includegraphics[width=1.0\linewidth]{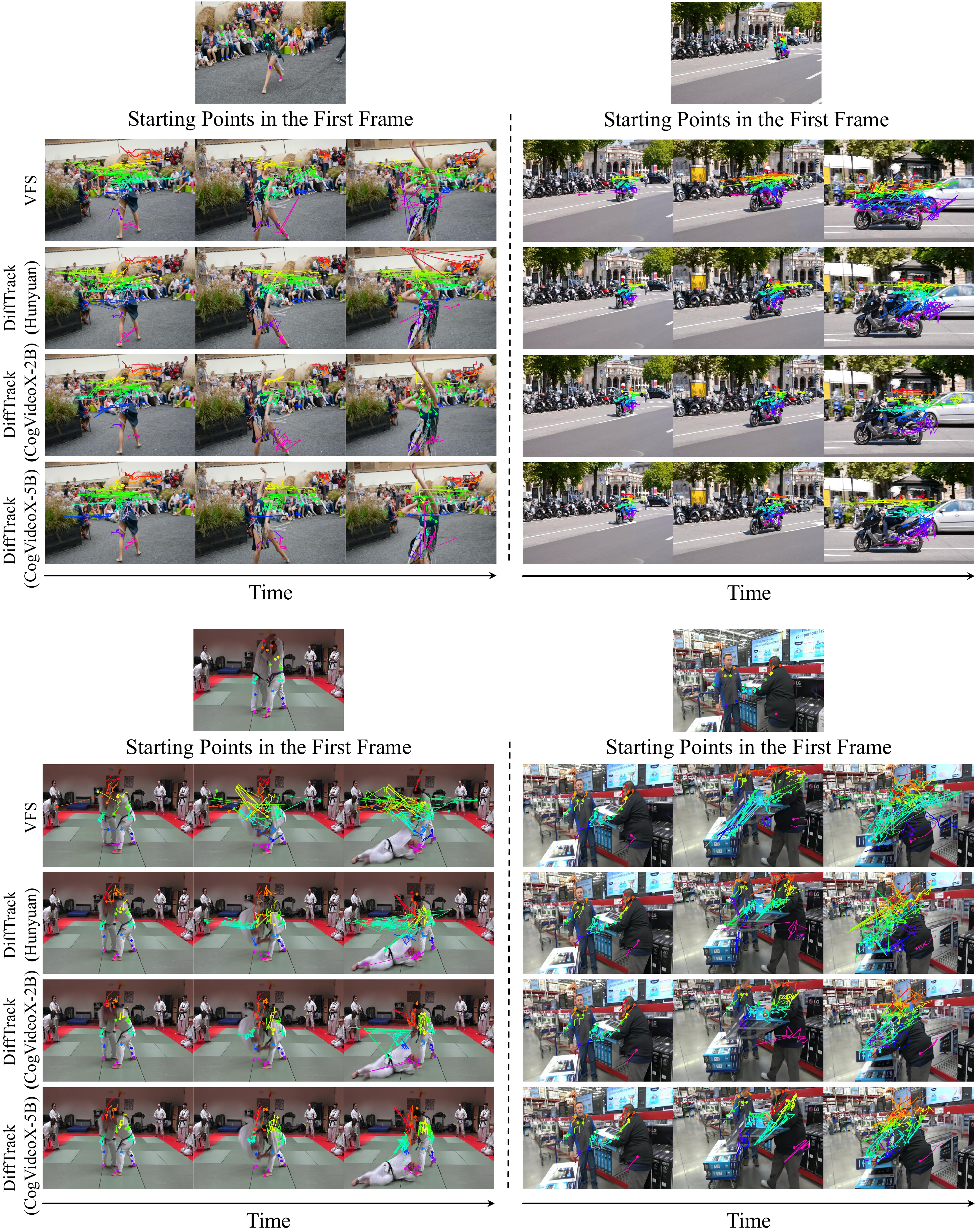}
    \caption{\textbf{Additional qualitative comparison for zero-shot point tracking.}
Our method produces smoother and more accurate trajectories compared to VFS~\cite{xu2021rethinking}, which struggle with temporal dynamics and often yield inconsistent tracks.}

    \label{fig:supp_qual}
    \figureaftercaption
\end{figure*}

%% file: figure/sup_qual_guidance_5b.tex
\begin{figure*}
    \centering
\includegraphics[width=1.0\linewidth]{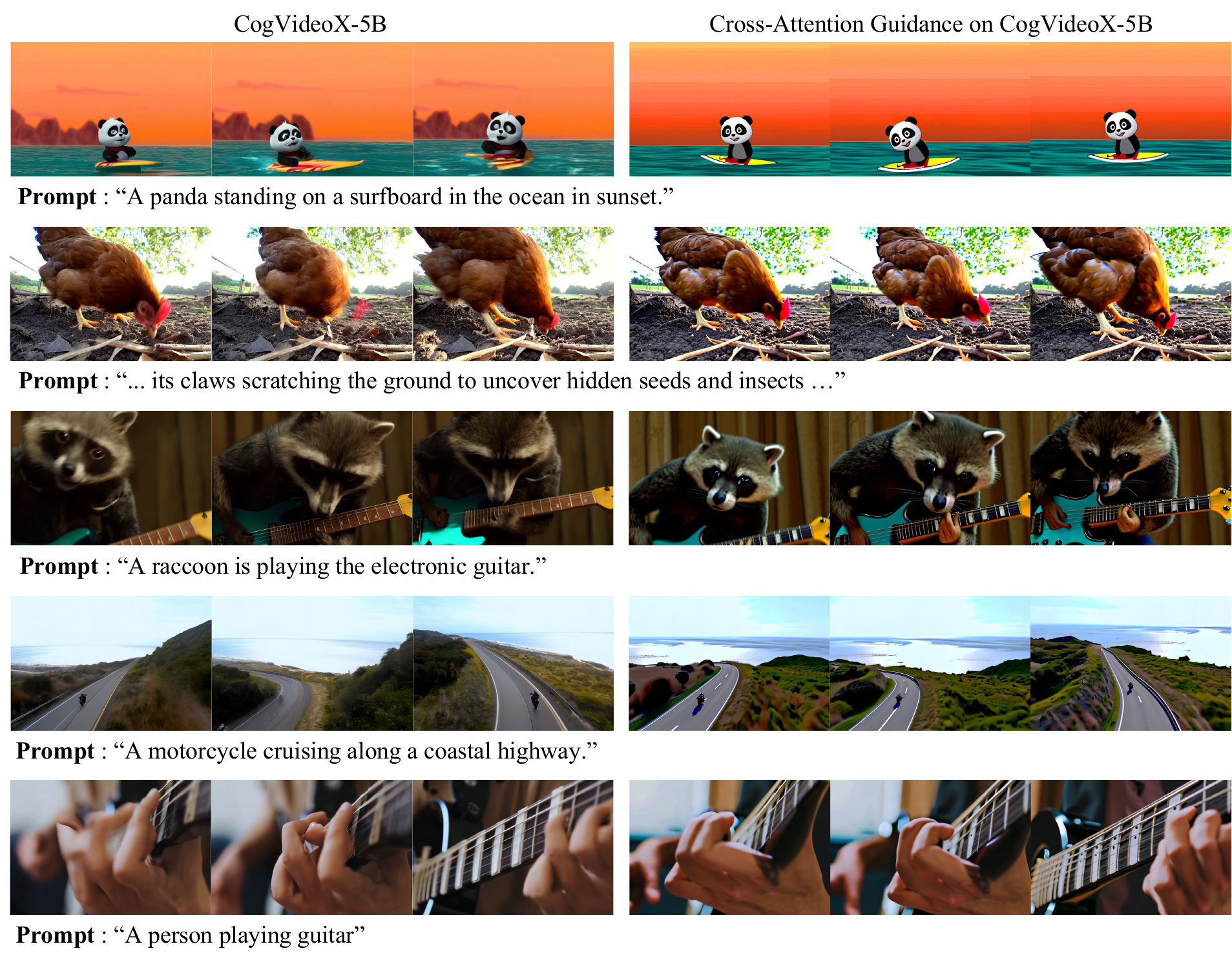}
    \caption{\textbf{Additional qualitative results on motion-enhanced generation with CogVideoX-5B.} Our sampling method, CAG, produces videos with improved motion consistency.}
    \label{fig:supp_qual_guidance_5b}
    \figureaftercaption
\end{figure*}

%% file: neurips_2025.bbl
\begin{thebibliography}{10}

\bibitem{achiam2023gpt}
Josh Achiam, Steven Adler, Sandhini Agarwal, Lama Ahmad, Ilge Akkaya, Florencia~Leoni Aleman, Diogo Almeida, Janko Altenschmidt, Sam Altman, Shyamal Anadkat, et~al.
\newblock {GPT}-4 technical report.
\newblock {\em arXiv preprint arXiv:2303.08774}, 2023.

\bibitem{ahn2024self}
Donghoon Ahn, Hyoungwon Cho, Jaewon Min, Wooseok Jang, Jungwoo Kim, SeonHwa Kim, Hyun~Hee Park, Kyong~Hwan Jin, and Seungryong Kim.
\newblock Self-rectifying diffusion sampling with perturbed-attention guidance.
\newblock In {\em ECCV}, pages 1--17, 2024.

\bibitem{an2024cross}
Honggyu An, Jinhyeon Kim, Seonghoon Park, Jaewoo Jung, Jisang Han, Sunghwan Hong, and Seungryong Kim.
\newblock Cross-view completion models are zero-shot correspondence estimators.
\newblock {\em arXiv preprint arXiv:2412.09072}, 2024.

\bibitem{an2023latent}
Jie An, Songyang Zhang, Harry Yang, Sonal Gupta, Jia-Bin Huang, Jiebo Luo, and Xi~Yin.
\newblock Latent-shift: Latent diffusion with temporal shift for efficient text-to-video generation.
\newblock {\em arXiv preprint arXiv:2304.08477}, 2023.

\bibitem{aydemir2024can}
G{\"o}rkay Aydemir, Weidi Xie, and Fatma G{\"u}ney.
\newblock Can visual foundation models achieve long-term point tracking?
\newblock {\em arXiv preprint arXiv:2408.13575}, 2024.

\bibitem{blattmann2023stable}
Andreas Blattmann, Tim Dockhorn, Sumith Kulal, Daniel Mendelevitch, Maciej Kilian, Dominik Lorenz, Yam Levi, Zion English, Vikram Voleti, Adam Letts, et~al.
\newblock Stable video diffusion: Scaling latent video diffusion models to large datasets.
\newblock {\em arXiv preprint arXiv:2311.15127}, 2023.

\bibitem{blattmann2023align}
Andreas Blattmann, Robin Rombach, Huan Ling, Tim Dockhorn, Seung~Wook Kim, Sanja Fidler, and Karsten Kreis.
\newblock Align your latents: High-resolution video synthesis with latent diffusion models.
\newblock In {\em CVPR}, pages 22563--22575, 2023.

\bibitem{cai2024ditctrl}
Minghong Cai, Xiaodong Cun, Xiaoyu Li, Wenze Liu, Zhaoyang Zhang, Yong Zhang, Ying Shan, and Xiangyu Yue.
\newblock {DiTCtrl}: Exploring attention control in multi-modal diffusion transformer for tuning-free multi-prompt longer video generation.
\newblock {\em arXiv preprint arXiv:2412.18597}, 2024.

\bibitem{cai2024can}
Ruojin Cai, Jason~Y. Zhang, Philipp Henzler, Zhengqi Li, Noah Snavely, and Ricardo Martin-Brualla.
\newblock Can generative video models help pose estimation?
\newblock {\em arXiv preprint arXiv:2412.16155}, 2024.

\bibitem{caron2021emerging}
Mathilde Caron, Hugo Touvron, Ishan Misra, Herv{\'e} J{\'e}gou, Julien Mairal, Piotr Bojanowski, and Armand Joulin.
\newblock Emerging properties in self-supervised vision transformers.
\newblock In {\em ICCV}, pages 9650--9660, 2021.

\bibitem{chefer2025videojam}
Hila Chefer, Uriel Singer, Amit Zohar, Yuval Kirstain, Adam Polyak, Yaniv Taigman, Lior Wolf, and Shelly Sheynin.
\newblock {VideoJAM}: Joint appearance-motion representations for enhanced motion generation in video models.
\newblock {\em arXiv preprint arXiv:2502.02492}, 2025.

\bibitem{chen2023videocrafter1}
Haoxin Chen, Menghan Xia, Yingqing He, Yong Zhang, Xiaodong Cun, Shaoshu Yang, Jinbo Xing, Yaofang Liu, Qifeng Chen, Xintao Wang, et~al.
\newblock Videocrafter1: Open diffusion models for high-quality video generation.
\newblock {\em arXiv preprint arXiv:2310.19512}, 2023.

\bibitem{cho2021cats}
Seokju Cho, Sunghwan Hong, Sangryul Jeon, Yunsung Lee, Kwanghoon Sohn, and Seungryong Kim.
\newblock {CAT}s: Cost aggregation transformers for visual correspondence.
\newblock {\em NeurIPS}, 34:9011--9023, 2021.

\bibitem{cho2022cats++}
Seokju Cho, Sunghwan Hong, and Seungryong Kim.
\newblock {CATs++}: Boosting cost aggregation with convolutions and transformers.
\newblock {\em IEEE TPAMI}, 45(6):7174--7194, 2022.

\bibitem{cho2025seurat}
Seokju Cho, Jiahui Huang, Seungryong Kim, and Joon-Young Lee.
\newblock Seurat: From moving points to depth.
\newblock {\em arXiv preprint arXiv:2504.14687}, 2025.

\bibitem{cho2024local}
Seokju Cho, Jiahui Huang, Jisu Nam, Honggyu An, Seungryong Kim, and Joon-Young Lee.
\newblock Local all-pair correspondence for point tracking.
\newblock In {\em ECCV}, pages 306--325, 2024.

\bibitem{darcet2023vision}
Timoth{\'e}e Darcet, Maxime Oquab, Julien Mairal, and Piotr Bojanowski.
\newblock Vision transformers need registers.
\newblock {\em arXiv preprint arXiv:2309.16588}, 2023.

\bibitem{doersch2022tap}
Carl Doersch, Ankush Gupta, Larisa Markeeva, Adria Recasens, Lucas Smaira, Yusuf Aytar, Joao Carreira, Andrew Zisserman, and Yi~Yang.
\newblock {TAP-Vid}: A benchmark for tracking any point in a video.
\newblock {\em NeurIPS}, 35:13610--13626, 2022.

\bibitem{doersch2023tapir}
Carl Doersch, Yi~Yang, Mel Vecerik, Dilara Gokay, Ankush Gupta, Yusuf Aytar, Joao Carreira, and Andrew Zisserman.
\newblock {TAPIR}: Tracking any point with per-frame initialization and temporal refinement.
\newblock In {\em ICCV}, pages 10061--10072, 2023.

\bibitem{esser2024scaling}
Patrick Esser, Sumith Kulal, Andreas Blattmann, Rahim Entezari, Jonas M{\"u}ller, Harry Saini, Yam Levi, Dominik Lorenz, Axel Sauer, Frederic Boesel, et~al.
\newblock Scaling rectified flow transformers for high-resolution image synthesis.
\newblock In {\em ICML}, 2024.

\bibitem{fang2020perceptual}
Yuming Fang, Hanwei Zhu, Yan Zeng, Kede Ma, and Zhou Wang.
\newblock Perceptual quality assessment of smartphone photography.
\newblock In {\em CVPR}, pages 3677--3686, 2020.

\bibitem{gao2024cat3d}
Ruiqi Gao, Aleksander Holynski, Philipp Henzler, Arthur Brussee, Ricardo Martin-Brualla, Pratul Srinivasan, Jonathan~T. Barron, and Ben Poole.
\newblock {CAT3D}: Create anything in 3{D} with multi-view diffusion models.
\newblock {\em arXiv preprint arXiv:2405.10314}, 2024.

\bibitem{geng2024motion}
Daniel Geng, Charles Herrmann, Junhwa Hur, Forrester Cole, Serena Zhang, Tobias Pfaff, Tatiana Lopez-Guevara, Carl Doersch, Yusuf Aytar, Michael Rubinstein, et~al.
\newblock Motion prompting: Controlling video generation with motion trajectories.
\newblock {\em arXiv preprint arXiv:2412.02700}, 2024.

\bibitem{genmo2023mochi}
Genmo.
\newblock Mochi 1: A new {SOTA} in open-source video generation models, 2023.

\bibitem{guo2024sparsectrl}
Yuwei Guo, Ceyuan Yang, Anyi Rao, Maneesh Agrawala, Dahua Lin, and Bo~Dai.
\newblock {SparseCtrl}: Adding sparse controls to text-to-video diffusion models.
\newblock In {\em ECCV}, pages 330--348, 2024.

\bibitem{guo2023animatediff}
Yuwei Guo, Ceyuan Yang, Anyi Rao, Zhengyang Liang, Yaohui Wang, Yu~Qiao, Maneesh Agrawala, Dahua Lin, and Bo~Dai.
\newblock {AnimateDiff}: Animate your personalized text-to-image diffusion models without specific tuning.
\newblock {\em arXiv preprint arXiv:2307.04725}, 2023.

\bibitem{hacohen2024ltx}
Yoav HaCohen, Nisan Chiprut, Benny Brazowski, Daniel Shalem, Dudu Moshe, Eitan Richardson, Eran Levin, Guy Shiran, Nir Zabari, Ori Gordon, et~al.
\newblock {LTX-Video}: Realtime video latent diffusion.
\newblock {\em arXiv preprint arXiv:2501.00103}, 2024.

\bibitem{harley2022particle}
Adam~W. Harley, Zhaoyuan Fang, and Katerina Fragkiadaki.
\newblock Particle video revisited: Tracking through occlusions using point trajectories.
\newblock In {\em ECCV}, pages 59--75, 2022.

\bibitem{he2022latent}
Yingqing He, Tianyu Yang, Yong Zhang, Ying Shan, and Qifeng Chen.
\newblock Latent video diffusion models for high-fidelity long video generation.
\newblock {\em arXiv preprint arXiv:2211.13221}, 2022.

\bibitem{hedlin2023unsupervised}
Eric Hedlin, Gopal Sharma, Shweta Mahajan, Hossam Isack, Abhishek Kar, Andrea Tagliasacchi, and Kwang~Moo Yi.
\newblock Unsupervised semantic correspondence using stable diffusion.
\newblock {\em NeurIPS}, 36:8266--8279, 2023.

\bibitem{ho2020denoising}
Jonathan Ho, Ajay Jain, and Pieter Abbeel.
\newblock Denoising diffusion probabilistic models.
\newblock {\em NeurIPS}, 33:6840--6851, 2020.

\bibitem{hong2022integrative}
Sunghwan Hong, Seokju Cho, Seungryong Kim, and Stephen Lin.
\newblock Integrative feature and cost aggregation with transformers for dense correspondence.
\newblock {\em arXiv preprint arXiv:2209.08742}, 2022.

\bibitem{hong2022cost}
Sunghwan Hong, Seokju Cho, Jisu Nam, Stephen Lin, and Seungryong Kim.
\newblock Cost aggregation with 4{D} convolutional swin transformer for few-shot segmentation.
\newblock In {\em ECCV}, pages 108--126. Springer, 2022.

\bibitem{hong2024unifying}
Sunghwan Hong, Jaewoo Jung, Heeseong Shin, Jiaolong Yang, Seungryong Kim, and Chong Luo.
\newblock Unifying correspondence pose and nerf for generalized pose-free novel view synthesis.
\newblock In {\em CVPR}, pages 20196--20206, 2024.

\bibitem{hong2021deep}
Sunghwan Hong and Seungryong Kim.
\newblock Deep matching prior: Test-time optimization for dense correspondence.
\newblock In {\em ICCV}, pages 9907--9917, 2021.

\bibitem{hong2022neural}
Sunghwan Hong, Jisu Nam, Seokju Cho, Susung Hong, Sangryul Jeon, Dongbo Min, and Seungryong Kim.
\newblock Neural matching fields: Implicit representation of matching fields for visual correspondence.
\newblock {\em NeurIPS}, 35:13512--13526, 2022.

\bibitem{huang2024vbench}
Ziqi Huang, Yinan He, Jiashuo Yu, Fan Zhang, Chenyang Si, Yuming Jiang, Yuanhan Zhang, Tianxing Wu, Qingyang Jin, Nattapol Chanpaisit, et~al.
\newblock {VBench}: Comprehensive benchmark suite for video generative models.
\newblock In {\em CVPR}, pages 21807--21818, 2024.

\bibitem{jabri2020space}
Allan Jabri, Andrew Owens, and Alexei Efros.
\newblock Space-time correspondence as a contrastive random walk.
\newblock {\em NeurIPS}, 33:19545--19560, 2020.

\bibitem{jeong2024track4gen}
Hyeonho Jeong, Chun-Hao~Paul Huang, Jong~Chul Ye, Niloy Mitra, and Duygu Ceylan.
\newblock {Track4Gen}: Teaching video diffusion models to track points improves video generation.
\newblock {\em arXiv preprint arXiv:2412.06016}, 2024.

\bibitem{jin2025appearance}
Siyoon Jin, Jisu Nam, Jiyoung Kim, Dahyun Chung, Yeong-Seok Kim, Joonhyung Park, Heonjeong Chu, and Seungryong Kim.
\newblock Appearance matching adapter for exemplar-based semantic image synthesis.
\newblock {\em arXiv preprint arXiv:2412.03150}, 2024.

\bibitem{karaev2024cotracker3}
Nikita Karaev, Iurii Makarov, Jianyuan Wang, Natalia Neverova, Andrea Vedaldi, and Christian Rupprecht.
\newblock {CoTracker3}: Simpler and better point tracking by pseudo-labelling real videos.
\newblock {\em arXiv preprint arXiv:2410.11831}, 2024.

\bibitem{karaev2024cotracker}
Nikita Karaev, Ignacio Rocco, Benjamin Graham, Natalia Neverova, Andrea Vedaldi, and Christian Rupprecht.
\newblock {CoTracker}: It is better to track together.
\newblock In {\em ECCV}, pages 18--35, 2024.

\bibitem{ke2021musiq}
Junjie Ke, Qifei Wang, Yilin Wang, Peyman Milanfar, and Feng Yang.
\newblock {MUSIQ}: Multi-scale image quality transformer.
\newblock In {\em ICCV}, pages 5148--5157, 2021.

\bibitem{kim2025exploring}
In{\`e}s~Hyeonsu Kim, Seokju Cho, Jiahui Huang, Jung Yi, Joon-Young Lee, and Seungryong Kim.
\newblock Exploring temporally-aware features for point tracking.
\newblock {\em arXiv preprint arXiv:2501.12218}, 2025.

\bibitem{kim2025moditalker}
Seyeon Kim, Siyoon Jin, Jihye Park, Kihong Kim, Jiyoung Kim, Jisu Nam, and Seungryong Kim.
\newblock {MoDiTalker}: Motion-disentangled diffusion model for high-fidelity talking head generation.
\newblock In {\em AAAI}, pages 4302--4310, 2025.

\bibitem{kirillov2023segment}
Alexander Kirillov, Eric Mintun, Nikhila Ravi, Hanzi Mao, Chloe Rolland, Laura Gustafson, Tete Xiao, Spencer Whitehead, Alexander~C. Berg, Wan-Yen Lo, et~al.
\newblock Segment anything.
\newblock In {\em ICCV}, pages 4015--4026, 2023.

\bibitem{kong2024hunyuanvideo}
Weijie Kong, Qi~Tian, Zijian Zhang, Rox Min, Zuozhuo Dai, Jin Zhou, Jiangfeng Xiong, Xin Li, Bo~Wu, Jianwei Zhang, et~al.
\newblock {HunyuanVideo}: A systematic framework for large video generative models.
\newblock {\em arXiv preprint arXiv:2412.03603}, 2024.

\bibitem{kling2024}
Kuaishou.
\newblock Kling: Video generation by kuaishou, 2024.
\newblock Accessed: 2025-03.

\bibitem{li2022efficient}
Muyang Li, Ji~Lin, Chenlin Meng, Stefano Ermon, Song Han, and Jun-Yan Zhu.
\newblock Efficient spatially sparse inference for conditional gans and diffusion models.
\newblock {\em NeurIPS}, 35:28858--28873, 2022.

\bibitem{li2023spatial}
Rui Li and Dong Liu.
\newblock Spatial-then-temporal self-supervised learning for video correspondence.
\newblock In {\em CVPR}, pages 2279--2288, 2023.

\bibitem{liu2024reconx}
Fangfu Liu, Wenqiang Sun, Hanyang Wang, Yikai Wang, Haowen Sun, Junliang Ye, Jun Zhang, and Yueqi Duan.
\newblock {ReconX}: Reconstruct any scene from sparse views with video diffusion model.
\newblock {\em arXiv preprint arXiv:2408.16767}, 2024.

\bibitem{liu2024sora}
Yixin Liu, Kai Zhang, Yuan Li, Zhiling Yan, Chujie Gao, Ruoxi Chen, Zhengqing Yuan, Yue Huang, Hanchi Sun, Jianfeng Gao, et~al.
\newblock Sora: A review on background, technology, limitations, and opportunities of large vision models.
\newblock {\em arXiv preprint arXiv:2402.17177}, 2024.

\bibitem{meng2025not}
Benyuan Meng, Qianqian Xu, Zitai Wang, Xiaochun Cao, and Qingming Huang.
\newblock Not all diffusion model activations have been evaluated as discriminative features.
\newblock {\em NeurIPS}, 37:55141--55177, 2025.

\bibitem{nam2024dreammatcher}
Jisu Nam, Heesu Kim, DongJae Lee, Siyoon Jin, Seungryong Kim, and Seunggyu Chang.
\newblock {DreamMatcher}: appearance matching self-attention for semantically-consistent text-to-image personalization.
\newblock In {\em CVPR}, pages 8100--8110, 2024.

\bibitem{nam2023diffusion}
Jisu Nam, Gyuseong Lee, Sunwoo Kim, Hyeonsu Kim, Hyoungwon Cho, Seyeon Kim, and Seungryong Kim.
\newblock Diffusion model for dense matching.
\newblock {\em arXiv preprint arXiv:2305.19094}, 2023.

\bibitem{nam2025visual}
Jisu Nam, Soowon Son, Zhan Xu, Jing Shi, Difan Liu, Feng Liu, Aashish Misraa, Seungryong Kim, and Yang Zhou.
\newblock Visual {P}ersona: Foundation model for full-body human customization.
\newblock {\em arXiv preprint arXiv:2503.15406}, 2025.

\bibitem{oquab2023dinov2}
Maxime Oquab, Timoth{\'e}e Darcet, Th{\'e}o Moutakanni, Huy Vo, Marc Szafraniec, Vasil Khalidov, Pierre Fernandez, Daniel Haziza, Francisco Massa, Alaaeldin El-Nouby, et~al.
\newblock {DINOv2}: Learning robust visual features without supervision.
\newblock {\em arXiv preprint arXiv:2304.07193}, 2023.

\bibitem{peebles2023scalable}
William Peebles and Saining Xie.
\newblock Scalable diffusion models with transformers.
\newblock In {\em ICCV}, pages 4195--4205, 2023.

\bibitem{podell2023sdxl}
Dustin Podell, Zion English, Kyle Lacey, Andreas Blattmann, Tim Dockhorn, Jonas M{\"u}ller, Joe Penna, and Robin Rombach.
\newblock {SDXL}: Improving latent diffusion models for high-resolution image synthesis.
\newblock {\em arXiv preprint arXiv:2307.01952}, 2023.

\bibitem{polyak2025moviegencastmedia}
Adam Polyak, Amit Zohar, Andrew Brown, Andros Tjandra, Animesh Sinha, Ann Lee, Apoorv Vyas, Bowen Shi, Chih-Yao Ma, Ching-Yao Chuang, et~al.
\newblock Movie {Gen}: A cast of media foundation models, 2025.

\bibitem{pont20172017}
Jordi Pont-Tuset, Federico Perazzi, Sergi Caelles, Pablo Arbel{\'a}ez, Alex Sorkine-Hornung, and Luc Van~Gool.
\newblock The 2017 {DAVIS} challenge on video object segmentation.
\newblock {\em arXiv preprint arXiv:1704.00675}, 2017.

\bibitem{qian2023semantics}
Rui Qian, Shuangrui Ding, Xian Liu, and Dahua Lin.
\newblock Semantics meets temporal correspondence: Self-supervised object-centric learning in videos.
\newblock In {\em ICCV}, pages 16675--16687, 2023.

\bibitem{radford2021learning}
Alec Radford, Jong~Wook Kim, Chris Hallacy, Aditya Ramesh, Gabriel Goh, Sandhini Agarwal, Girish Sastry, Amanda Askell, Pamela Mishkin, Jack Clark, et~al.
\newblock Learning transferable visual models from natural language supervision.
\newblock In {\em ICML}, pages 8748--8763, 2021.

\bibitem{raffel2020exploring}
Colin Raffel, Noam Shazeer, Adam Roberts, Katherine Lee, Sharan Narang, Michael Matena, Yanqi Zhou, Wei Li, and Peter~J. Liu.
\newblock Exploring the limits of transfer learning with a unified text-to-text transformer.
\newblock {\em JMLR}, 21(140):1--67, 2020.

\bibitem{rombach2022high}
Robin Rombach, Andreas Blattmann, Dominik Lorenz, Patrick Esser, and Bj{\"o}rn Ommer.
\newblock High-resolution image synthesis with latent diffusion models.
\newblock In {\em CVPR}, pages 10684--10695, 2022.

\bibitem{runway2024gen3}
Runway.
\newblock Introducing {Gen}-3 {Alpha}, 2024.
\newblock Accessed: 2025-03.

\bibitem{singer2022make}
Uriel Singer, Adam Polyak, Thomas Hayes, Xi~Yin, Jie An, Songyang Zhang, Qiyuan Hu, Harry Yang, Oron Ashual, Oran Gafni, et~al.
\newblock {Make-A-Video}: Text-to-video generation without text-video data.
\newblock {\em arXiv preprint arXiv:2209.14792}, 2022.

\bibitem{song2020denoising}
Jiaming Song, Chenlin Meng, and Stefano Ermon.
\newblock Denoising diffusion implicit models.
\newblock {\em arXiv preprint arXiv:2010.02502}, 2020.

\bibitem{su2024roformer}
Jianlin Su, Murtadha Ahmed, Yu~Lu, Shengfeng Pan, Wen Bo, and Yunfeng Liu.
\newblock Roformer: Enhanced transformer with rotary position embedding.
\newblock {\em Neurocomputing}, 568:127063, 2024.

\bibitem{sun2024dimensionx}
Wenqiang Sun, Shuo Chen, Fangfu Liu, Zilong Chen, Yueqi Duan, Jun Zhang, and Yikai Wang.
\newblock Dimension{X}: Create any 3{D} and 4{D} scenes from a single image with controllable video diffusion.
\newblock {\em arXiv preprint arXiv:2411.04928}, 2024.

\bibitem{tang2023emergent}
Luming Tang, Menglin Jia, Qianqian Wang, Cheng~Perng Phoo, and Bharath Hariharan.
\newblock Emergent correspondence from image diffusion.
\newblock {\em NeurIPS}, 36:1363--1389, 2023.

\bibitem{teed2020raft}
Zachary Teed and Jia Deng.
\newblock {RAFT}: Recurrent all-pairs field transforms for optical flow.
\newblock In {\em ECCV}, pages 402--419, 2020.

\bibitem{truong2020glu}
Prune Truong, Martin Danelljan, and Radu Timofte.
\newblock {GLU-Net}: Global-local universal network for dense flow and correspondences.
\newblock In {\em CVPR}, pages 6258--6268, 2020.

\bibitem{truong2021learning}
Prune Truong, Martin Danelljan, Luc Van~Gool, and Radu Timofte.
\newblock Learning accurate dense correspondences and when to trust them.
\newblock In {\em CVPR}, pages 5714--5724, 2021.

\bibitem{vaswani2017attention}
Ashish Vaswani, Noam Shazeer, Niki Parmar, Jakob Uszkoreit, Llion Jones, Aidan~N. Gomez, {\L}ukasz Kaiser, and Illia Polosukhin.
\newblock Attention is all you need.
\newblock {\em NeurIPS}, 30, 2017.

\bibitem{wu2024cat4d}
Rundi Wu, Ruiqi Gao, Ben Poole, Alex Trevithick, Changxi Zheng, Jonathan~T. Barron, and Aleksander Holynski.
\newblock {CAT4D}: Create anything in 4{D} with multi-view video diffusion models.
\newblock {\em arXiv preprint arXiv:2411.18613}, 2024.

\bibitem{xiao2024video}
Zeqi Xiao, Yifan Zhou, Shuai Yang, and Xingang Pan.
\newblock Video diffusion models are training-free motion interpreter and controller.
\newblock {\em arXiv preprint arXiv:2405.14864}, 2024.

\bibitem{xing2024dynamicrafter}
Jinbo Xing, Menghan Xia, Yong Zhang, Haoxin Chen, Wangbo Yu, Hanyuan Liu, Gongye Liu, Xintao Wang, Ying Shan, and Tien-Tsin Wong.
\newblock {DynamiCrafter}: Animating open-domain images with video diffusion priors.
\newblock In {\em ECCV}, pages 399--417, 2024.

\bibitem{xu2021rethinking}
Jiarui Xu and Xiaolong Wang.
\newblock Rethinking self-supervised correspondence learning: A video frame-level similarity perspective.
\newblock In {\em ICCV}, pages 10075--10085, 2021.

\bibitem{yang2024cogvideox}
Zhuoyi Yang, Jiayan Teng, Wendi Zheng, Ming Ding, Shiyu Huang, Jiazheng Xu, Yuanming Yang, Wenyi Hong, Xiaohan Zhang, Guanyu Feng, et~al.
\newblock {CogVideoX}: Text-to-video diffusion models with an expert transformer.
\newblock {\em arXiv preprint arXiv:2408.06072}, 2024.

\bibitem{zhang2023tale}
Junyi Zhang, Charles Herrmann, Junhwa Hur, Luisa Polania~Cabrera, Varun Jampani, Deqing Sun, and Ming-Hsuan Yang.
\newblock A tale of two features: Stable diffusion complements dino for zero-shot semantic correspondence.
\newblock {\em NeurIPS}, 36:45533--45547, 2023.

\bibitem{zhang2024world}
Qihang Zhang, Shuangfei Zhai, Miguel~Angel Bautista, Kevin Miao, Alexander Toshev, Joshua Susskind, and Jiatao Gu.
\newblock World-consistent video diffusion with explicit 3{D} modeling.
\newblock {\em arXiv preprint arXiv:2412.01821}, 2024.

\bibitem{zheng2024open}
Zangwei Zheng, Xiangyu Peng, Tianji Yang, Chenhui Shen, Shenggui Li, Hongxin Liu, Yukun Zhou, Tianyi Li, and Yang You.
\newblock {Open-Sora}: Democratizing efficient video production for all.
\newblock {\em arXiv preprint arXiv:2412.20404}, 2024.

\end{thebibliography}
